# Operational early warning of thunderstorm-driven power outages from open data: a two-stage machine learning approach

Iryna Stanishevska[1] and Seth Guikema[1,2]

**Abstract**

Thunderstorm-driven power outages are difficult to predict because most storms do not cause damage, convective processes occur rapidly and chaotically, and the available public data are noisy and incomplete. Severe convective storms now account for a large and rising share of U.S. weather losses, yet thunderstorm-induced outages remain understudied. We develop a 48-hour early-warning model for summer thunderstorm-related outages in Michigan using only open-source outage (EAGLE-I) and weather (METAR) data. Relative to prior work, we (i) rely solely on public data, (ii) preserve convective extremes from a sparse station network via parameter-specific kriging and causal spatiotemporal features, and (iii) use a multi-level LSTM-based architecture evaluated on event-centric peak metrics. The pipeline builds rolling and k-NN inverse-distance aggregates to capture moisture advection, wind shifts, and pressure drops. A two-stage design uses a logistic gate followed by a long short-term memory (LSTM) regressor to filter routine periods and limit noise exposure. Evaluation focuses on state-level peaks of at least 50,000 customers without power, using hits, misses, false alarms, and peak-conditional MASE (cMASE) within 48-hour windows, with uncertainty quantified by block bootstrapping. On the test sample, the Two-Stage model detects more peaks with only one additional false alarm and reduces cMASE near peaks, providing event-focused early warnings without the utility-specific data.

## 1. Introduction

Reliable energy supply is essential for social and economic stability, and large power outages threaten public health and safety. In the United States, the frequency of weather-driven outages has increased in recent years (Mukherjee et al., 2018; Kenward and Raja, 2014). Hurricane-related risks are well studied, but severe convective storms have become a major and growing driver of U.S. outages. NOAA's Billion-Dollar Disasters record shows rising frequency and CPI-adjusted costs of "severe storm" events in recent years, often comprising the largest share of billion-dollar disasters (NOAA NCEI, 2025). Yet thunderstorm-induced outages remain under-examined (Mukherjee, 2018). Modeling is challenging because many storms cause no damage, onset is rapid and localized, and impacts depend on local factors and vulnerabilities (e.g., vegetation, soil type, and infrastructure conditions) (Haseltine, 2017). Publicly available data are also inconsistent and incomplete (Swanson et al., 2022), which adds epistemic uncertainty.

[1] Department of Safety, Economics, and Planning, University of Stavanger, Stavanger, Norway. Corresponding author: stanishevska.iryna@gmail.com alternative: i.stanishevska@stud.uis.no
[2] Civil and Environmental Engineering, University of Michigan, Ann Arbor, MI. sguikema@umich.edu

This paper addresses this gap with a 48-hour early warning model for thunderstorm-driven outages built solely on open data (e.g., ORNL EAGLE-I; IEM METAR) and evaluated with event-centric metrics. The open data aspect is a critical advance. Previous thunderstorm outage modeling has used privately held utility data (e.g., Kabir et al. 2019 and Shashaani et al. 2018). These models have strong predictive accuracy but are usable only within the utility for which they were developed. This limits their broader use for storm preparation by both public and private sector organizations. We show in this paper that an accurate thunderstorm outage prediction model can be developed based on only publicly accessible data.

The study focuses on Michigan, as a case study, but the method can be used elsewhere as well. Michigan consistently exceeds the U.S. average reliability indices (Casey et al., 2020; Climate Central, 2022). The analysis is limited to summer periods 2021-2022 because this was the latest period with publicly available data.

Relative to previous work modeling thunderstorm-induced power outages (e.g., Kabir et al., 2019 Shashanni at el. 2018), our approach makes several contributions. First, we develop a model using only publicly available, open-source data for both the weather driver and the response variable for training. Second, the data we use consists of only weather station data. We preserve convective extremes from sparse METAR data via parameter-specific kriging with hourly variograms and targeted over-drafting to retain peaks and build strictly causal spatiotemporal features. Focusing on station data rather than gridded forecast data decreases the burden of outage forecasting in practice because station data are easier to process than gridded forecast data once the kriging model has been trained. Third, while we use a multi-stage approach as in Guikema and Quiring (2012), Shashanni et al. (2018), and McRoberts (2019) we use a more advanced LSTM neural network for our second stage predictor, increasing the predictive accuracy of the model. We also focus on accuracy assessment on the time periods with peaks in outages through a cluster-based error metric. We train the model on the 2021 data and test it with the 2022 data. Our results show that the two-stage approach detects more peak outage times at a 48-hour lead time and more accurately predicts outages during these peaks relative to a more traditional single-stage LSTM model.

## 2. Background

### 2.1. Infrastructure and outage trends

Michigan shows comparatively high weather-related outage frequency, with 157 significant weather-related outage events over 2000–2023, second only to Texas (Climate Central, 2022; DOE OE-417, ORML, 2024). At the same time, Michigan's electric infrastructure has been graded "D" ("At Risk"), reflecting aging grids and under investment (ASCE, 2023; MPSC, 2023). Consistent with this, that SAIDI (System Average Interruption Duration Index) for Michigan exceeds U.S. averages even without Major Event Days (IEEE, EIA, Table 11.4). Uneven population distribution across the state, with urbanized areas and industries primarily concentrated in the south of Michigan, is reflected in the spatial heterogeneity of outage risks. The condition of distribution lines varies across the state, with more frequent failures in Detroit and socially vulnerable regions (MPSC; MPSC, 2025). This further contributes to spatial heterogeneity of outage risks in the region, together with vulnerability factors beyond thunderstorm exposure.

Within this context, our goal is to develop a model to predict (a) when a major thunderstorm-induced power outage event will occur in Michigan and (b) the magnitude of this event in terms of the number of customers without power. The goal is to make this prediction with a 48-hour lead time based on only publicly accessible data.

### 2.2. Meteorological rationale for explanatory features

Summer conditions in Michigan are characterized by convective storms with rapid, localized onset. There are about 30–40 thunderstorm days per year in Michigan. This thunderstorm activity occurs mostly during summer months (WeatherSpark, n.d.), and the number of storm-induced outages is continuously growing, likely caused by climate warming (NOAA NCEI, 2025).

Summer convective environments in Michigan are formed by warm air masses, frontal tendencies with cooler air flows, and vertical wind shear, causing high convective available potential energy (CAPE) (Changnon and Kunkel, 2006). Therefore, this study uses data on near-surface air temperature, dew point, pressure, wind speed and direction, relative humidity, and precipitation. The higher dew point and rising instability are expected to increase outage risk, while pressure falls and shear gusts signal storm formation and damaging winds (Panthou et al., 2014). However, the fragmented, uneven dynamics of thunderstorm cell formation, characterized by complex interconnections and temporal dependencies and sharp gradients (e.g., the most intense precipitation typically occurs just to the north of the area with the highest dew point), are equally significant. Furthermore, these processes include an inherent stochastic (aleatoric) component, as does the chaotic nature of the atmosphere.

## 3. Related works

### 3.1. Modeling in outage risk management

The earliest approaches used generalized linear models (GLMs) to estimate outage risks (Liu et al., 2005) but did not show high accuracy in spatial distribution. Liu et al. (2008) addressed the spatial correlation of outages through spatial generalized linear mixed models (GLMMs), but they did not observe a substantial improvement in accuracy. With decision-tree-based methods (Breiman, 2001), nonparametric models (CART, BART, and GAM) captured nonlinear interactions (Guikema et al., 2010). Random Forests were successfully applied to hurricane outage risk using only open-source data (Nateghi et al., 2014) and to identify key risk factors (Guikema et al., 2014). Quiring and Guikema (2012) developed a two-stage approach for outage modeling in which a first-stage classifier first predicted if an outage would occur, and then a second-stage regression model predicted the number of outages for the predicted outage events. A number of papers built on this approach for outage prediction, including models for thunderstorm outage prediction (Kabir et al. 2019) and all-weather outage prediction (McRoberts et al. 2018). Shashanni et al. (2018) extended this approach to a three-stage model where the second stage predicts which of a set of clusters of outage events a given predicted outage event will be in, and then a separate regression model is used to predict outage totals for each cluster. Kabir et al. (2019) further extended this line of work by developing a Bayesian ensemble prediction approach for the second stage with the weights assigned to each ensemble member dynamically updated based on model error in the previous epoch. In parallel with this, deep learning algorithms have also been applied to power outage forecasting to try to better capture complex temporal and spatial patterns (LeCun et al., 2015; Kor et al., 2020; Haseltine, 2017). LSTM-based models were used to predict outages at the customer level (Abaas et al., 2022), and CNN+LSTM forecasted outage-prone areas (Huang et al., 2024).

## 3.2. Distinctive aspects in outage modeling

Most studies treat individual geographic units (e.g., utility service areas or counties) in isolation, even though power grid damages caused by external conditions are typically not independent, and outage dynamics in neighboring areas are often correlated. However, models tied to a single utility and local conditions can poorly generalize to other regions (Arora and Ceferino, 2023). Local conditions (tree-trimming, infrastructure age, soil, vegetation) can strongly shape outage risk (Guikema et al., 2006; Wanik et al., 2017), but data on these factors are often unavailable. Moreover, outage historical records are frequently highly imbalanced/zero-inflated, with rare large-scale peaks. Multi-stage setups mitigate this by filtering zero-inflated samples at the first stage and then modeling magnitude (McRoberts et al., 2018; Shashaani et al., 2018). All of the above issues are even more severe for thunderstorms. Unlike hurricanes, thunderstorms exhibit a weak and noisy mapping from occurrence to actual damage, while impacts are highly sensitive to local conditions. The onset is rapid and localized, while the impact depends strongly on local conditions (He et al., 2016).

## 4. Methods

### 4.1. Data overview

One of our key goals in this paper is to use only publicly accessible data. This eases the ability of the resulting model and modeling approach to be used by public officials and others without access to utility-specific data that only utilities have access to. In particular, a model based on public data would be much more useful for public officials to use in planning power outage events.

#### 4.1.1. EAGLE-I outage data

EAGLE-I (Environment for the Analysis of Geo-Located Energy Information) is a platform developed by the Oak Ridge National Laboratory (ORNL) and commissioned by the U.S. Department of Energy's Office of Cybersecurity, Energy Security, and Emergency Response (OCESER). This project provides centralized monitoring of electricity outages for more than 92% of electricity consumers in the USA and Territories (Chinthavali et al., 2023).

The dataset contains historical outage records for an 8-year period (2014–2022) with a 15-minute time interval, which allows observing outage dynamics with high temporal detail. It should be noted that the coverage of utilities has changed over time, with better coverage in the later years of the data set than in the earlier years. All values are aggregated by county level, accounting for an approximate number of "customers" in each county. Michigan records were downloaded from the official DOI repository via ORNL Constellation (10.13139/ORNLNCCS/1975203). The dataset includes the state, county FIPS code, county name, timestamp, and the count of "customers" without power, where a "customer" can refer to a household, a business, or an individual meter. Coverage includes all 83 counties.

Key limitations of the dataset are partial utility coverage, especially in rural areas, heterogeneous utility feeds (points, ZIP codes, or polygons crossing several counties) normalized to counties, and scraping/timeout gaps during major events (Brelsford et al., 2024). Zero records do not always mean the actual absence of outage records, because data can be missed or recorded incorrectly due to connection failures or other technical difficulties (Brelsford et al., 2024). These factors introduce epistemic uncertainty with geo-referencing noise and temporal discrepancies. Therefore, this study uses EAGLE-I as an operational indicator of outage magnitude and applies consistency checks with OE417 reports (Section 4.2).

#### 4.1.2. OE-417 reports

The DOE OE417 is a mandatory incident report that applies to large disruptions, defined as either affecting more than 50,000 customers or resulting in a load loss of 300 MW or more for over 15 minutes (U.S. Department of Energy, n.d.). Records contain the start and end timestamps of the outage, the affected area, the number of affected customers, and the reason for the incident. Reports with Michigan incidents for summer 2021–2022 were extracted from the OE-417 annual summaries (Office of Cybersecurity, Energy Security, and Emergency Response [OCESER], 2023). Event lists for summer 2021-2022 are provided in Appendix A, Tables A.1-A.2.

#### 4.1.3. METAR data

For short-horizon convective outage prediction, this study uses raw airport station observations (ASOS METAR) from the Environmental Mesonet (IEM). Compared to reanalysis or aggregated feeds, station data provide better initial accuracy (Rzeszutek et al., 2017), more accurate temporal detail (Åström and Wittenmark, 2013; Kim et al., 2019) and finer temporal resolution compared to simplistic interpolation (Hofstra et al., 2010) that is critical for short-term storms. We use ten METAR variables from the IEM archive: air temperature (tmpf, °F), dew point temperature (dwpf, °F), relative humidity (relh, %), wind direction (drct, °), wind speed (sknt, kt), precipitation (p01i, in), altimeter (alti, inHg), sea-level pressure (mslp, mb/hPa), wind gust (gust, kt), and present weather codes (wxcode). Because airport stations are unevenly distributed across the state (Appendix Fig. A.1), we interpolate weather parameters to county centroids as described in Section 5.1.4.

### 4.2. Data preprocessing

Our goal is to transform raw EAGLE-I and METAR feeds into a county-hour dataset aligned at observation time to feed both feature construction and label formation. The sections below discuss the process of the outage data first, followed by the processing of the weather data.

#### 4.2.1. EAGLE-I data cleaning and alignment

We work at the county-hour level across all 83 Michigan counties. Raw EAGLE-I data feeds provide county totals at 15-minute resolution. In 2021 the share of missing data is 3.2% and in 2022 it is 2.37%. Missing values within time windows of up to 4 consecutive hours were filled using linear interpolation. This assumes that outage records do not significantly change during this interval. Longer gaps were removed rather than assuming no change over these longer periods.

For hourly aggregation, we select within each hour the 15-min timestamp at which the statewide total is maximal ('max-concurrency') and take all county values at that timestamp. This preserves temporal concurrency of outages across counties, captures short events relevant to decision thresholds defined at the state level (e.g., 50,000 customers), and avoids inflation from summing per-county within-hour maxima. Using end-to-end hour bins shows similar aggregate errors but misses brief threshold crossings. State-level hourly series after cleaning are shown in Appendix B, Fig. B.1-B.2.

The +48 h lead for target formation is applied per county. Because some county series contain short pre-event gaps, small residual asynchrony can remain after the shift. As a result, the +48 h aligned series may show slight timing attenuation and lower apparent peaks relative to $y_t$. This reflects the operational scoring grid (state-level hourly series after cleaning shown on Figs. B3-B4).

#### 4.2.2. METAR data cleaning and alignment

Raw METAR observations from 75 Michigan airport stations (10 meteorological parameters from the IEM archive; see Section 4.3) were resampled hourly with parameter-specific aggregation. Summary statistics were checked before/after resampling (Table 1) to verify that the initial data distribution and outliers were preserved for further analysis. Different methods were used to handle missing data during the aggregation to avoid artificial smoothing or the distortion of values, and details are provided in Table 1. Full before-and-after aggregation summaries are in Tables B.1-B.2.

**Table 1** METAR variables, hourly aggregation, and missing data handling prior to spatial interpolation.

| No. | **Variable** | **Description** | **Units** | **Hourly aggregation** | **Missing-data handling** |
|---|---|---|---|---|---|
| 1 | tmpf | Air Temperature | °F | mean | forward-fill ⩽ 2 h |
| 2 | dwpf | Dew Point Temperature | °F | mean | forward-fill ⩽ 2 h |
| 3 | relh | Relative Humidity | % | mean | forward-fill ⩽ 2 h |
| 4 | drct | Wind Direction | ° | direction at timestamp of within-hour max wind speed | forward-fill ⩽ 2 h |
| 5 | sknt | Wind Speed | kt | max | forward-fill ⩽ 2 h |
| 6 | p01i | precipitation | in | max over available values | if all sub-hourly values are missing, 0 (assumed no precip). |
| 7 | alti | Altimeter Setting | inHg | mean | forward-fill ⩽ 2 h |
| 8 | mslp | Sea-level pressure | mb | mean | forward-fill ⩽ 2 h |
| 9 | gust | Wind gust | kt | max | no forward-fill; gusts kept missing |
| 10 | wxcode | Weather codes: TS (thunderstorm); SQ (squalls); HR (heavy rain) with boolean indicators | N/A | boolean flags (TS/SQ/HR) | absent code marked as False (no forward-fill) |

Note: Hourly resampling and missing-data handling for METAR variables used in this study before spatial interpolation

We convert process wind reports (direction in degrees from true north; speed in knots) to horizontal components. Within each hour, we pair the maximum wind speed with the direction at the timestamp of that maximum. Speed is converted to m/s ($1\ kt\ =\ 0.514444\ m/s$). Let $V$ be speed (m/s) and $\theta$ the direction in radians measured clockwise from north. Following the meteorological convention (Grange, 2014), the eastward and northward components are:

$$\theta = drct \cdot \frac{\pi}{180}, \qquad u = -V sin\,(\theta), \quad v = -V cos\,(\theta)$$

As a result, the final dataset includes wind components (*u* and *v*) that correspond to the hourly maximum wind speed, which can be used for further interpolation and analysis. For diagnostics, we reconstruct:

$$V = \sqrt{u^2 + v^2}\,,\ \widehat{drct} = (atan2(-u,\ -v)\ \frac{\pi}{180})\ mod\ 360.$$

### 4.2.3. Meteorological Interpolation

An uneven station network (Fig. A.1) creates “blind” areas and epistemic uncertainty. We therefore map station observations to county centroids with parameter-specific methods.

*Kriging of continuous fields (OK/UK)*

For air temperature, dew point, pressure/altimeter, and relative humidity, we fit hourly variograms Following Burrough and McDonnell (1998) and krige to county centroids within a fixed search radius ($r_{max}$) based on the parameter’s characteristics. Stations beyond $r_{max}$ are excluded to avoid spurious long-range correlation. The kriging predictor given by (Burrough and McDonnell, 1998) is:

$$Z^*\,(x_0) = \sum_{i=1}^{n} \lambda_i \;\; z(x_i)$$

where $\lambda_i$ from the variogram and covariance. Universal Kriging with linear drift is applied at a noticeable linear trend (e.g., from the Great Lakes to the south for air temperature); otherwise, ordinary kriging is used to preserve local gradients and peaks. We krige station observations to county centroids within a fixed search radius. We evaluated out-of-sample performance by holding out each of three representative stations (one isolated, one median-density, and one densely surrounded). Kriging-variance diagnostics are provided in Appendixes C1-C4.

*Kriging of wind direction*

We apply ordinary kriging to the *u* and *v* components using hourly variograms and a fixed search radius. Direction is reconstructed only for diagnostics from interpolated *u* and *v*. Distributional before/after summaries and kriging diagnostics are provided in Appendix C.7. Figure 1 contrasts station winds with kriged county-centroid winds at 19:00 UTC, August 2021.

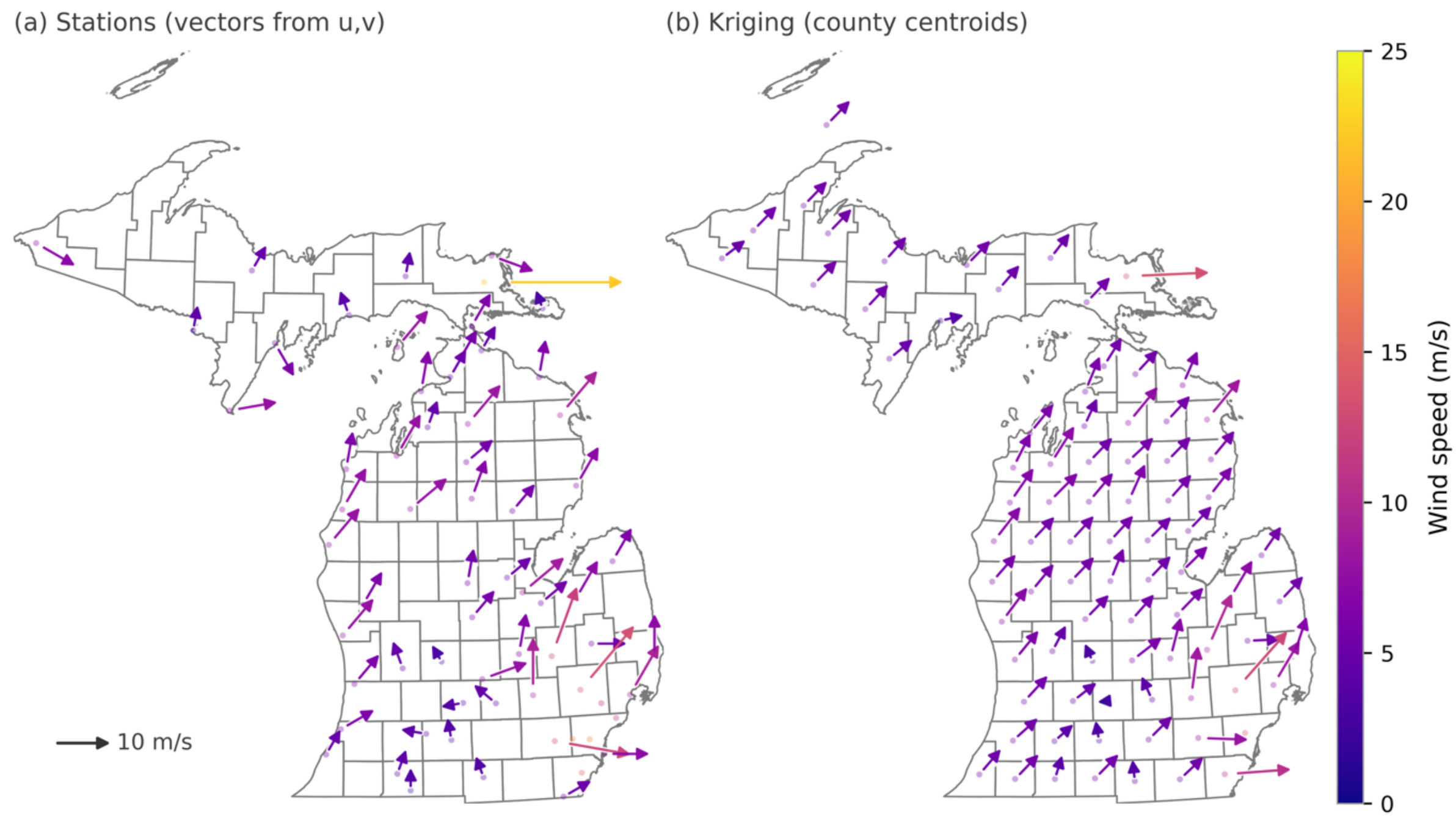

**Figure 1**. Wind vectors (m/s) at 11 Aug 2021 19:00 UTC. Left: station observations (vectors reconstructed from hourly u, v; dots mark station locations). Right: kriged vectors at county centroids. Arrow length is scaled by a common factor: color encodes wind speed (shared color bar), A 10 m $s^{-1}$ reference vector is shown in panel. County boundaries are shown in gray.

*"Overdrafting" of extremes.*

To prevent smoothing out physically important peaks (e.g., local gusts, dew points), after kriging we overwrite the kriged value at a county centroid with the nearest observed extreme if a station is within a variable-specific radius. We apply this to dew points and gusts. Exact radii and quantiles are listed in Table 2 (dew point: $r = 200$ km, ≥ 90th and ≤ 10th percentile; wind speed: $r = 100$ km, ≥ 90th percentile). This preserves sharp convective signatures without unrealistic spatial blurring (Table 2). Figure 2 shows an example of overdrafting at the dew point kriging field. Before/after summaries after overdrafting are shown in Appendix C.5 (dew point: Tables C.5.1-C.5.2; Figs. C.5.2-C.5.4) and Appendix C.6 (gusts: Tables C.6.1-C.6.2; Figs. C.6.2-C.6.4).

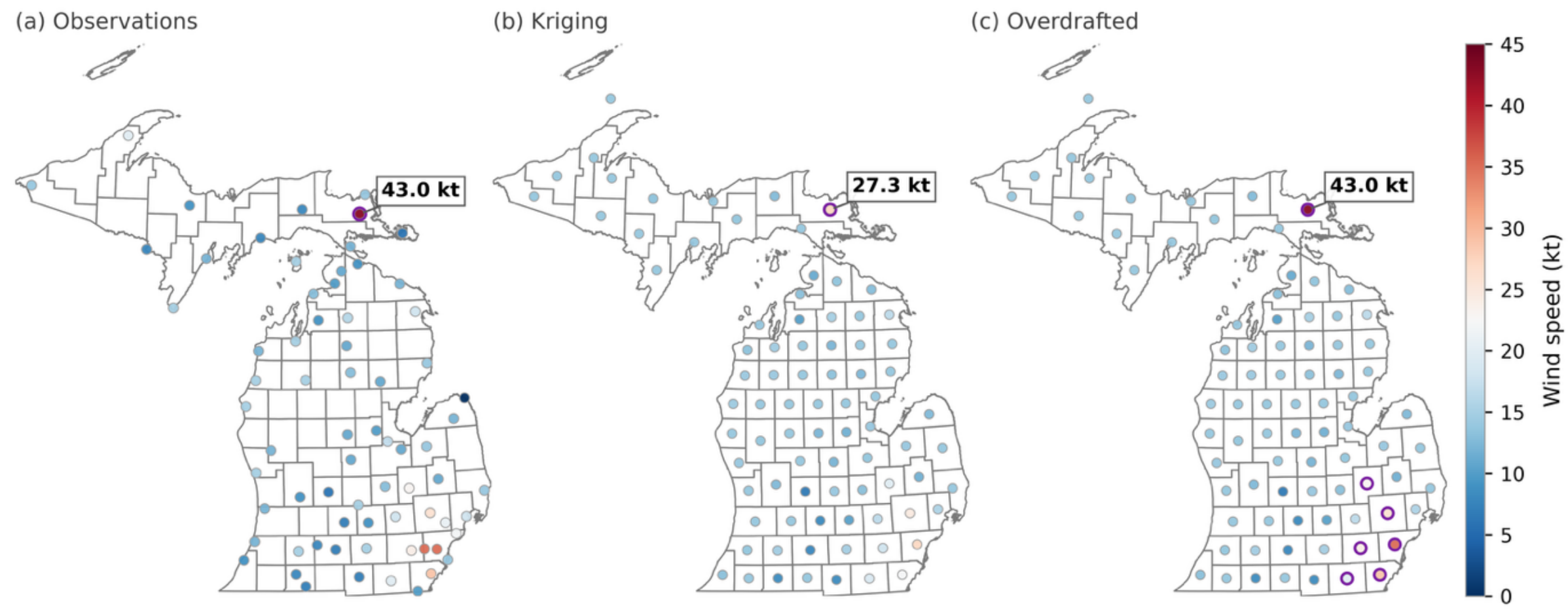

**Figure 2.** Wind speed (kt) at 2021-08-11 over Michigan. (a) METAR station observations; (b) kriged estimates at county centroids; (c) overdrafted field that restores physically plausible station extremes. Outlined circles mark centroids where values changed after overdrafting, and the boxed label indicates the maximum wind speed of the overdrafted value. All panels share the same map extent and color scale.

*Local humidity contrast*

In addition to the relative humidity kriging field, we compute the gradient of the nearest neighbors as a marker of local contrast processes essential for convection:

$$g_i = \frac{|z_i - z_j|}{d_{ij}}, \quad j = \arg\min_{j \neq i} d_{ij}, \quad d_{min} = d_{ij}$$

where $z_i$ is the relative humidity value on the station $i$, $z_j$ is the value at the nearest station, $d_{ij}$ is the distance between the stations $i$ and $j$ (KD-tree; limited to 100 km radius; no neighbors = 0) and $d_{min} = d_{ij}$ is the minimum distance. Before/after distributions of humidity gradient are provided in Table C.4.6.

*Discrete codes and localized weather phenomena*

Continuous kriging is not applied to discrete flags with weather codes (TS/SQ/HR) and such parameters as gusts and precipitation due to the narrow physical footprint. We instead apply a polygon join based on a predefined parameter-specific radius (Table 2).

**Table 2**. Interpolation settings by variable: kriging type and variogram parameters; overdrafting applied where noted

| Meteorological variable | Interpolation method | Variogram model & key parameters |
|---|---|---|
| Air Temperature (tmpf) | Universal Kriging | Spherical; max lag=250 km;<br>linear regional drift |
| Altimeter Setting (alti) | Universal Kriging | Spherical; max lag=250 km;<br>linear regional drift |
| Sea Level Pressure (mslp) | Universal Kriging | Spherical; max lag=250 km;<br>binning: Sturges; linear regional drift |
| Wind Direction (drct) | Ordinary Kriging | Spherical; max lag=180 km; 7 lags |
| Relative Humidity (relh) | Ordinary Kriging<br>with gradient computation | Spherical; max lag=100 km |
| Dew Point (dwpf) | Universal Kriging<br>with overdrafting of outliers<br>(max radius 250 km)[1] | Spherical; max lag=250 km;<br>linear regional drift |
| Wind Speed (sknt) | Ordinary Kriging<br>with overdrafting of outliers<br>(max radius 100 km)[2] | Spherical; max lag=100 km<br>15 lags, binning: fd |
| Wind Gust (gust) | Point-in-polygon to county | N/A |
| Precipitation (p01i) | Point-in-polygon to county | N/A |
| Thunderstorm Flag (ts_flag) | Point-in-polygon to county | N/A |
| Heavy Rain Flag (hr_flag) | Point-in-polygon to county | N/A |
| Squall Flag (sq_flag) | Point-in-polygon to county | N/A |

Notes: Units: °F, kt, in, inHg, mb.IEM = Iowa Environmental Mesonet.

### 4.3. Data Analysis

#### 4.3.1. Distribution Shift 2021-2022

Between 2021 and 2022, medium/long-lag autocorrelation during calm periods declined (Tables D.1-D.2). Across counties, the median $\Delta ACF_{6\text{-}24h}$ was approximately -0.12 (e.g., Δ at 6 h = -0.20, 12 h = -0.16), suggesting changes in background dynamics (operations, infrastructure, emergency response). Within peak windows changes are small ($\Delta ACF_{6\text{-}24h}$ = -0.02), with a slight increase at 1 h (+0.05), indicating that extreme events have comparable temporal dependencies (Appendix D, Tables D.3-D.4). County-level heterogeneity is substantially larger during calm periods (IQR $ACF_{6\text{-}24h}$= 0.32) than in peaks (IQR $ACF_{6\text{-}24h}$ = 0.11).

#### 4.3.2. Moran's I Index

Global Moran's *I* (Anselin, 2003) for large-scale outages shows significant positive spatial clustering in both years (for 2021: $I$ = 0.400, p-value < 0.001, $z$ = 7,368; 2022: $I$ = 0.311, p-value < 0.001, $z$ = 6,365). During major events (2-h windows around peaks ≥ 50,000 customers), clustering remains positive and often significant, with $I \approx$ 0.04-0.36 (Appendix D, Table D.5). Thus, the degree of clustering may vary, indicating storm-to-storm variations may occur in the composition of counties involved in high peaks.

### 4.4. Response variable and holdout splits of the data

The modeling unit is *county* x *hour* across all 83 Michigan counties. The train data is for June-August 2021 and the test data is for June-August 2022. All features variables used are from only periods prior to the period for which the prediction is to be made $t_0$ (e.g,., $t \leq t_0$). All preprocessing (scaling, imputation, PCA, threshold) is fit on the training split only and applied to validation/test to avoid look-ahead bias. ). All preprocessing (scaling, imputation, PCA, threshold) is fit on the training split only and applied to validation/test to avoid look-ahead bias.

*Targets*

<u>Stage-1 (classification):</u> Let $t_0$ denote the observation hour. The binary label at $t_0$ is 1 (positive) if the county's outage at $t_0 + 48\ h$ is at or above its county-specific 90th percentile of outages (min-max normalization by county mitigates population-size bias), otherwise it is 0. A logistic regression predicts whether such an anomaly will occur at $t_0 + 48\ h$.The goal is a high recall near peak hours while limiting pass-through to Stage 2, particularly of events that have few outages.

<u>Stage-2 (regression):</u> the target is the county-level outage magnitude at *t+48h* in the *log* scale. This reduces distribution skew, handles zero values and down-weights the influence of rare extreme spikes. For county *i* we define:

$$z_{i,t_0} = log(1 + y_{i,t_0+48})$$

where $y_{i,t}$ is the number of disconnected customers in county $i$ per hour $t$. . Stage-2 predictions are inverse-transformed to original scale and summed across counties to form the state-level series used for evaluation:

$$\hat{S}_{t+48} = \sum_{i=1}^{83} \hat{y}_{i,t+48}$$

where $\hat{y}_{i,t+48}$ is the predicted number of disconnected customers in county $i$ at the time $t + 48h$ $\hat{S}_{t+48}$is the corresponding state-level prediction, the sum runs over all 83 Michigan counties and $\hat{y}_{i,t+48} S_t = \Sigma_i\, y_{i,t}$denotes the county-level prediction after Stage-1 filtering.

## 4.5. Class Imbalance Handling

To address class imbalance, we implement event-window undersampling. For each county, the series is partitioned into non-overlapping 48-hour windows. We retain only windows that contain ≥ 3 anomalies and ≥ 1 normal outage to remove isolated spikes and long quiet periods while preserving temporal contiguity around major events. Each retained window thus contains both classes within the same spatiotemporal context. Residual imbalance is addressed with class weights in Stage-1.

Our training data (Jun-Aug 2021) had *N* = 68,755 records with 4,626 positives (6.73%). Our test data (Jun-Aug 2022) and *N* = 62,936 records with 3,152 positives (5.01%). For Stage-1, we also train on an undersampled 2021 subset used only for the classifier (*N* = 9,918; positives: 3,375; 34%; Figure I.1).

All reported classification metrics use the full 2022 test. The Stage-2 regressor is evaluated on the subset of county-hours forwarded by Stage-1 on the test set (pass-through = 47.25%). Event-level metrics are computed on the state aggregate (sum over counties per hour) using hits, misses, and false alarms as described in Section 5.4. Regression errors are reported in the original customer scale (after inverse log transform). All predictions are scored after the +48 h alignment and state-level aggregation (see Section 5.4).

## 4.6. Feature Engineering

We construct features in six groups: temporal lags, rolling statistics, spatial aggregates, county-centroid coordinates, demographics, and calendar. Temporal features use only past observations at lags of 6, 12, 24, and 48 h relative to $t_0$, avoiding leakage across the 24-48 h horizon.

To capture spatial context, we use inverse-distance weighting (IDW) (Shepard, 1968; Bartier and Keller, 1996). The value at location $x$ is a normalized weighted average of nearby observations $z_i$ with distance-decay weights. Let $N_{(x)}$ denotes the neighborhood (k-nearest neighbors) $d(\cdot,\cdot)$ is the distance (the Euclidean in a projected CRS). If $d_{(x_i x_j)} = 0$ we set $z_{(x)} = z_i$. We use $\alpha = 2$ .

IDW is applied to selected raw variables and their temporal aggregates (lags/rolling stats). Nearby counties receive higher weights, improving sensitivity to local anomalies (e.g., high dew point, strong gusts, thunderstorm flags). Population density (ACS 2021-2022; density = population / area km$^2$), county coordinates, and calendar encodings complement meteorological signals. All features are computed causally with respect to the forecast time $t_0$ $(t \leq t_0)$. Preprocessing and hyperparameters (e.g., scaling, imputation, thresholds) are fitted on training folds only and applied to validation/test. Engineered feature groups and design rationale are summarized in Table 3. The full feature set *(N=87)* is provided in Appendix E, Table E.1.

**Table 3.** Engineered feature groups used and designed rationale.
Operations are causal (trailing windows) relative to $t_0$. The windows are in hours. All features are standardized on the training split.

| Parameter | Feature | Window(s), h | Rationale |
|---|---|---|---|
| air temp, dew point | lags | 6, 12, 24, 48 | fronts/shifts; moisture influx |
| wind_u, wind_v | lags | 6, 12, 24, 48 | sudden shifts; squall line |
| precipitation | rolling sum | 6, 12, 24, 48 | accumulated rain; soil saturation |
| relative humidity | rolling mean | 6, 12, 24, 48 | humidity build-up; short-term change |
| relative humidity gradient | rolling max | 6, 12, 24, 48 | local short-term change |
| MSLP/altimeter | rolling mean | 6, 12, 24, 48 | persistent pressure fall |
| gust/wind speed | rolling max | 6, 12, 24, 48 | extremes |
| thunderstorm/heavy rain/squall flags | rolling sum/max | 6, 12, 24, 48 | repeated storms |
| IDW aggregates | IDW(p=2) | - | nearby anomalies inform local risk |
| Population density | static | - | denser areas contribute more to large-scale peaks |
| Calendar (DOW) | categorical (0-6) | - | weekly patterns |
| Geographic proxy (county centroid lon, lat) | static | - | coarse infrastructural context |

*Notes*: RH = relative humidity; MSLP = mean sea-level pressure; IDW = inverse distance weighting; DOW = day-of-week. Distances are computed in projected CRS. Distances are computed in projected CBS.

On the 2021 training split, we compute Pearson correlations with the t+48 h targets (binary anomaly flag and log1p(sum_48)). Engineered features exhibit consistently higher absolute correlations than raw meteorological variables, indicating a stronger concentration of predictive signals (Table 4). Full lists are in Appendix I (Tables E.2-E.3)

**Table 4.** Top 15 features by Pearson correlation with the t+48 h targets on the 2021 training split. Left: anomaly flag. Right: log1p(sum_48).

| Rank | Feature | Corr (flag +48 h) | Rank | Feature | Corr (sum +48 h) |
|---|---|---|---|---|---|
| 1 | IDW_drct_v_lag_6h | 0.171 | 1 | population_density | 0.274 |
| 2 | IDW_dwpf_lag_12h | 0.164 | 2 | sq_flag_rolling_sum_48h | 0.231 |
| 3 | dwpf_lag_6h | 0.163 | 3 | sq_flag_rolling_sum_24h | 0.193 |
| 4 | dwpf_lag_12h | 0.160 | 4 | ts_flag_rolling_sum_48h | 0.168 |
| 5 | IDW_drct_v | 0.160 | 5 | sknt_rolling_max_48h | 0.162 |
| 6 | dwpf | 0.158 | 6 | p01i_rolling_sum_48h | 0.161 |
| 7 | drct_v_lag_12h | 0.157 | 7 | gust_rolling_max_48h | 0.149 |
| 8 | drct_v_lag_24h | 0.156 | 8 | ts_flag_rolling_sum_24h | 0.148 |
| 9 | IDW_gust_rolling_max_24h | 0.154 | 9 | sknt_rolling_max_24h | 0.144 |
| 10 | IDW_dwpf | 0.153 | 10 | sq_flag_rolling_sum_12h | 0.141 |
| 11 | drct_v | 0.150 | 11 | p01i_rolling_sum_24h | 0.139 |
| 12 | drct_v_lag_6h | 0.148 | 12 | gust_rolling_max_24h | 0.134 |
| 13 | sq_flag_rolling_sum_48h | 0.142 | 13 | IDW_gust_rolling_max_24h | 0.130 |
| 14 | IDW_sknt_rolling_max_48h | 0.133 | 14 | sknt_rolling_max_12h | 0.125 |
| 15 | sknt_rolling_max_24h | 0.131 | 15 | gust_rolling_max_12h | 0.123 |

## 4.7. Model

### 4.7.1. Stage-1

A logistic regression classifies whether a county-hour at $t_0$ contains signals of an anomaly label at $t_0 + 48\ h$. The objective is high recall near peak hours while limiting pass-through to Stage-2. We perform L1-based feature selection (8 predictors retained; Appendix I, Table I.1) and fit the final model with L2 under a class imbalance (undersampling + class weights). Hyperparameters are tuned via 3-fold time-series cross validation (Appendix I, Table I.2.). The operating threshold $\tau = 0.70$ is fixed on 2021 validation to maintain recall on peaks while reducing pass-through to Stage-2.

### 4.7.2. Stage 2

We fit a single-step LSTM (16 units) followed by a Dense (1) layer, using the same feature set to predict *log1p(sum_48)* as defined in Section 4.4. Inputs are the Stage-1 filtered samples formatted as tensors of shape (batch, 1, n_features), i.e., a single step per window. Thus, we avoid padding/masking, required by irregular gaps introduced at Stage-1. We optimize MSE in the log space with Adam optimizer and early stopping on validation loss. Predictions are inverted back to the original customer scale via exmp1(·). Hyperparameters, learning curves, and sensitivity are in Appendix I (Methods Note I.6). The 35 features used by the Stage-2 LSTM are listed in Appendix I, Table 1.7.

### 4.7.3. Baseline

For comparison to the two-stage model, we also train a one-step LSTM with the same architecture, features, target, preprocessing, and training setup as Stage-2, but without the Stage-1 filter (i.e., on the full 2021). Full details are provided in Appendix I (Methods Note I.6).

## 4.8. Feature Selection and PCA

Starting from 88 min-max-scaled variables, we removed near-constant predictors (IQR<1%) and pruning multicollinearity by dropping one of any pair with Spearman $|\rho|>0.90$ and iteratively removing variables with Variance Inflation Factor (VIF) >10. We then used 5-fold forward-changing time-series cross validation on Stage-1 positive windows, keeping a feature only if it reduced RMSE in logged scale for the +48 h target. Per-fold results are reported in Appendix Table I.7. All preprocessing and selection were refit within training folds to avoid leakage. Thus, we received a stable 35 feature-set presented in Appendix I, Table I.8.

PCA is computed on the training split for diagnostics only. Models are trained on the selected original features (no PCA projection). In the full 88-feature space, the first four PCs explain about 38% of variance. After feature selection, they explain about 48%, indicating that redundant features were removed. Loadings align with meteorological dynamics relevant to convective patterns: PC1 loads on dew point and pressure (moisture advection under falling pressure), PC2 on wind and gust proxies (background shear), PC3 on lagged moisture and directional shear, and PC4 on temperature-humidity contrast with gust/precipitation signals, consistent with convective development. Detailed loadings are variance plots in Appendix I (Tables I.8-I.9).

## 4.9. Evaluation Protocol

### 4.9.1. Task setup and alignment

As the objective is early warning of rare state-scale peaks (≥ 50,000 customers), we form hourly state-level series by summing county-level values per hour. All timestamps are UTC. Predictions issued at time *t* are pre-aligned to a *+48 h* and compared against the state total *at t+48 h* derived from EAGLE-I. County-level predictions at *t+48 h* are summed to the state level for scoring. Based on this, we compare the Two-Stage model against the One-Step LSTM baseline. No additional baselines are fitted.

#### 4.9.2. Stage-1 (classification) metrics

Given severe class imbalance, we report precision-recall (PR) curves, AUCPR (with the prevalence baseline), ROC-AUC, and Precision/Recall/F1 at the operation threshold $\tau = 0.70$ fixed on 2021 validation. We also report the pass-through rate (the fraction of county hours flagged positive and forwarded to Stage-2.

#### 4.9.3. Stage-2 (regression) metrics

On the state aggregate, we compute RMSE, MAE and MASE (after the inverse *log1p*).

### 4.9.4. Peak-conditional MASE (cMASE)

To focus on large state-level events, we evaluate MASE only on hours in near peaks. $P = \{t\colon y_t \geq 50{,}000\}$
be the set of peak hour, and for a tolerance $\Delta \in 6{,}12{,}24{,}36{,}48$ define the evaluation subset $S_\Delta = \left\{t\colon \min_{\rho \in \rho} |t-\rho| \leq \Delta\right\}$, where $|t-\rho|$ is in hours. Given the state series $y_t$ and its prediction $\hat{y}_t$ with seasonal lag h = 24 h, the cMASE at window Δ is:

$$P = \{t\colon y_t \geq 50{,}000\}$$

$$cMASE(\Delta) = \frac{\frac{1}{|S_\Delta|}\sum_{t \in S_\Delta} |y_t - \hat{y}_t|}{\frac{1}{T-h}\sum_{t=h+1}^{T} |y_t - y_{t-h}|}, \quad h = 24$$

Here *T* is the number of test hours. The denominator (seasonal-naïve scale) is computed once over the full test series to keep scales comparable across Δ and reporting scopes.

### 4.9.5. Event-level detection

We score detection of state-level events (≥ 50,000 customers) with the same procedure applied to the observed series $y_t$ and to the +48 h-aligned prediction $\hat{y}_t$ (all timestamps in UTC. That is, we use a cluster-based peak detector applied identically to the observed series $y_t$ and the +48 h-aligned prediction series $\hat{y}_t$ (all timestamps in UTC) The process is as follows:

1. Smooth the time series of outages with centered moving average, $k = 5$ h (partial edge windows, no padding);
2. Contiguous time segments are marked where the smoothed series of outages exceeds 50,000;
3. Time segments with the gap ≤ 24 h are merged;
4. For each merged segment, we define the event time as the arg max of the unsmoothed series within the segment. The event magnitude is the original value at the time.

Predicted events are extracted from $\hat{y}_t$ identically. Predicted and reference events are matched one-to-one by nearest time within a window $\pm w$ h, where $w \in \{6, 12, 24, 36, 48$ (greedy nearest-neighbour assignment is used). We report hits (matched predictions), misses (unmatched references), and false alarms (unmatched predictions) and derive precision, recall, and F1.

#### 4.9.6. Uncertainty estimation via bootstrap

We use a moving-block bootstrap with 168-hour blocks (≈ 1 week) and $B = 500 - 1000$ replicates. For each replicate we sample contiguous blocks with replacement to the test length, concatenate, and truncate. For every resample we: (i) re-detect and one-to-one match events using the same procedure (centered moving average *k=5* h with boundary shrinkage, gap-merge ≤ 24 h) within matching windows $\omega \in \{6,\ 12,\ 24,\ 36,\ 48\}$ hours; and (ii) recompute cMASE on peak-neighborhood sets $S_\Delta$. The seasonal-naive denominator (*h=24*) is computed once on the full test and reused across replicates to keep scales comparable. We report medians and 95% percentile intervals [2.5%, 97.5%] across replicates.

### 4.9.7. OE-417 matching

We use federal OE-417 reports as an external credibility check for EAGLE-I state-level peaks (≥50,000 customers). For each OE-417 event window, we take the within-window maximum of the EAGLE-I state total and take the within-window maximum. When multiple OE-417 entries refer to one multi-day event, we use the sum across days. The signed percent difference is:

$$\Delta\% = \frac{max(EAGLE - I) - max(OE - 417)}{max(OE - 417)} \ 100\%$$

If OE-417 magnitudes are missing (e.g., "loss of monitoring/control"), Δ% is reported as NA and the event is flagged as low-confidence. We assign qualitative match confidence as follows: high if peak times align within $\pm 6$ h and $|\Delta\%| \leq 15\%$; medium if within $\pm 12$ h and $|\Delta\%| \leq 40\%$ or if event windows are incomplete; low for "loss of monitoring/control", missing magnitudes, or poorly specified multi-day windows. Models are trained/evaluated on EAGLE-I data. OE-417 is used only to contextualize label uncertainty.

## 5. Results

### 5.2. Stage-1

The operating threshold $\tau = 0.70$ was fixed on the 2021 validation set to balance recall and pass-through (the share of county-hours forwarded to Stage-2), and the same threshold is used on Test-2022.

On Test-2022 (prevalence 5.01%): AUCPR = 0.059 (vs. the prevalence baseline 0.050); ROC-AUC = 0.560; At the operating threshold of 0.70 Precision, Recall, and F1 are 6.02%, 56.79%, and 10.88% respectively and the pass-through rate is 47.25%. *On Train-2021* (prevalence 6.73%)*,* AUCPR is 0.176, and ROC-AUC = 0.736. At the same threshold, Precision, Recall, and F1 are 8.80%, 87.29%, and 15.99%, pass-through rate is 66.73% and ROC-AUC = 0.736 (Table 5). PR curves are shown in Figure 3 (train) and Figure 4 (test). The threshold sweep on the original (unsampled distribution) is presented in Appendix I, Table I.3. Confusion matrices are shown in Tables I.4-5.

**Table 5.** Stage-1 performance at operating threshold τ =0.70 (fixed on 2021 validation). Metrics are evaluated on the full, unsampled distribution.

| Metric | Train (2021) | Test (2022) |
|---|---|---|
| N | 68,755 | 62,936 |
| Positives (prevalence) | 4,626 (6.7%) | 3,152 (5.0%) |
| AUCPR | 0.176 | 0.059 |
| Baseline (prevalence) | 6.7% | 5.0% |
| ROC-AUC | 0.736 | 0.560 |
| Precision | 8.8% | 6.0% |
| Recall | 87.3% | 56.8% |
| F1 | 15.9% | 10.9% |

| Pass-through to Stage-2 | 66.7% | 47.3% |
|---|---|---|

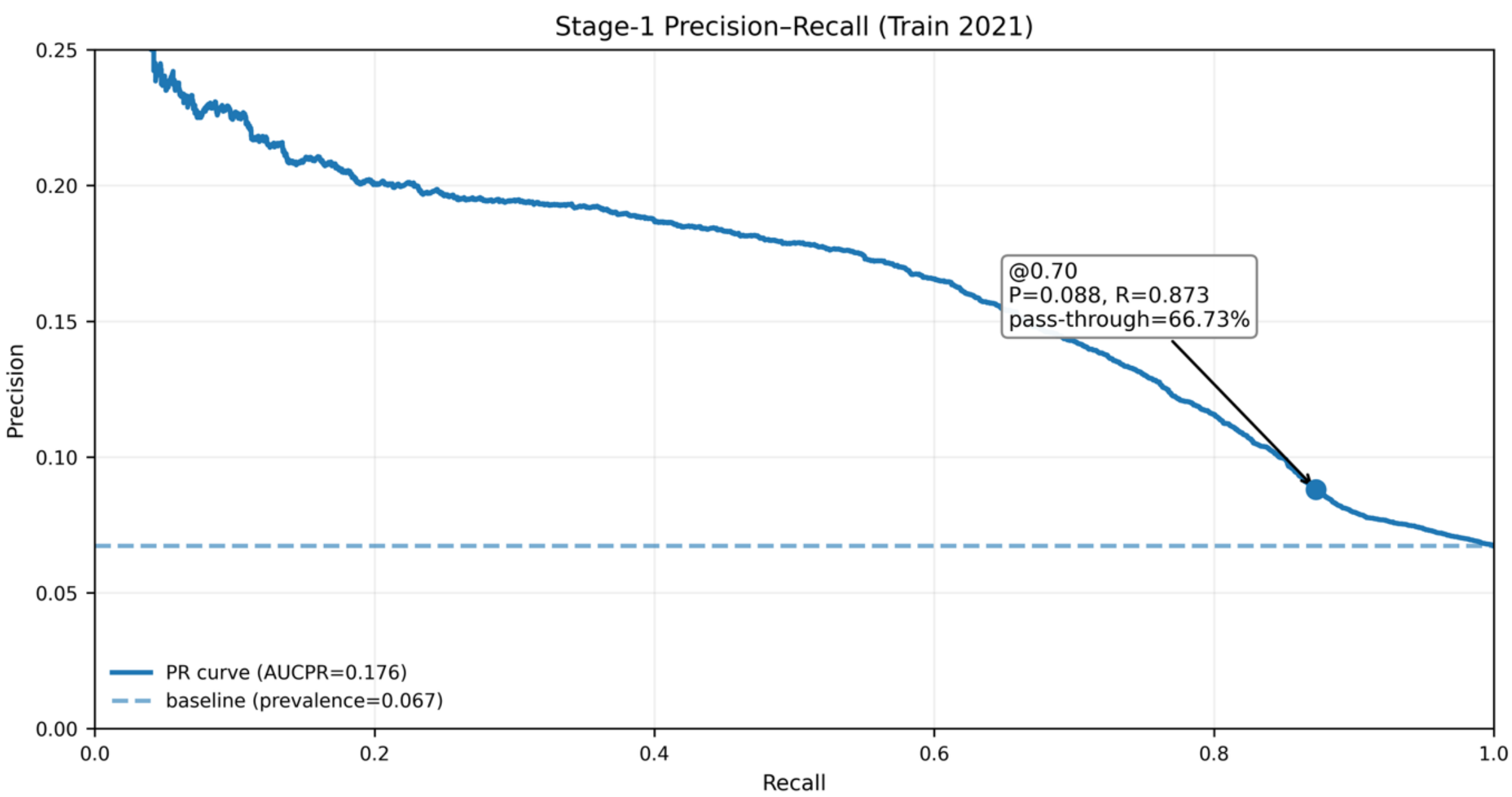

**Figure 3**. Stage-1 Precision-Recall curves Train-2021

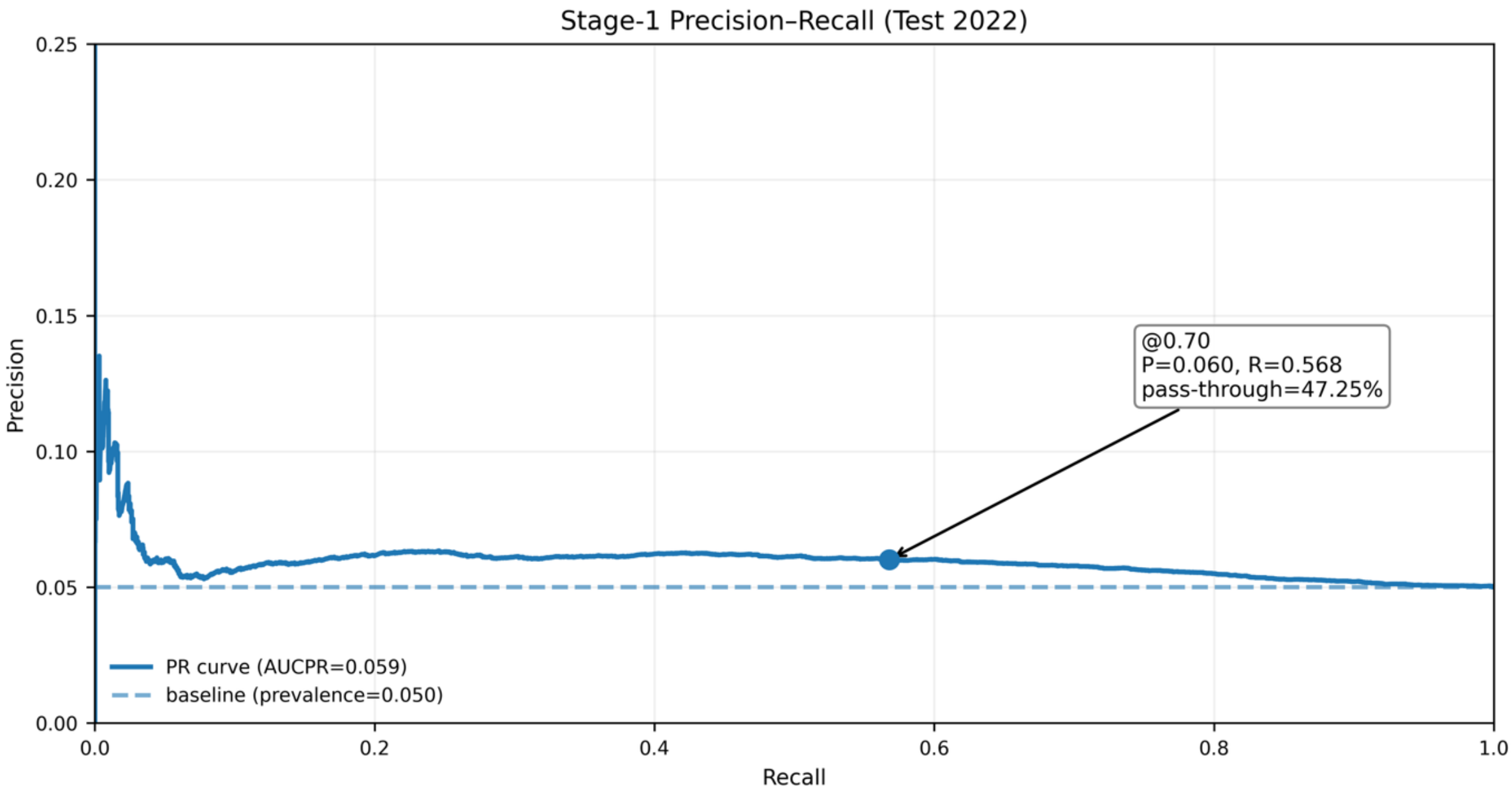

**Figure 4.** Stage-1 Precision-Recall curves Test-2022

## 5.2. Stage-2 Overall errors

Two-Stage: RMSE = 4,609, MAE = 606, $R^2$ = 0.133;
One-Step LSTM (baseline): RMSE = 4,110, MAE = 372, $R^2$ = 0.107

Overall, errors are of the same order. The one-stage baseline attains lower RMSE/MAE, while the two-stage yields a slightly higher $R^2$ (Appendix I, Tables I.11-I.14). Metrics are computed on the state aggregate in the original customer scale (after inverse log1p). Figure 5 shows 48-hour-ahead state-level outages vs. actuals (Test-2022). Dashed lines mark model peak times, and a horizontal dashed line shows the 50,000-customer threshold.

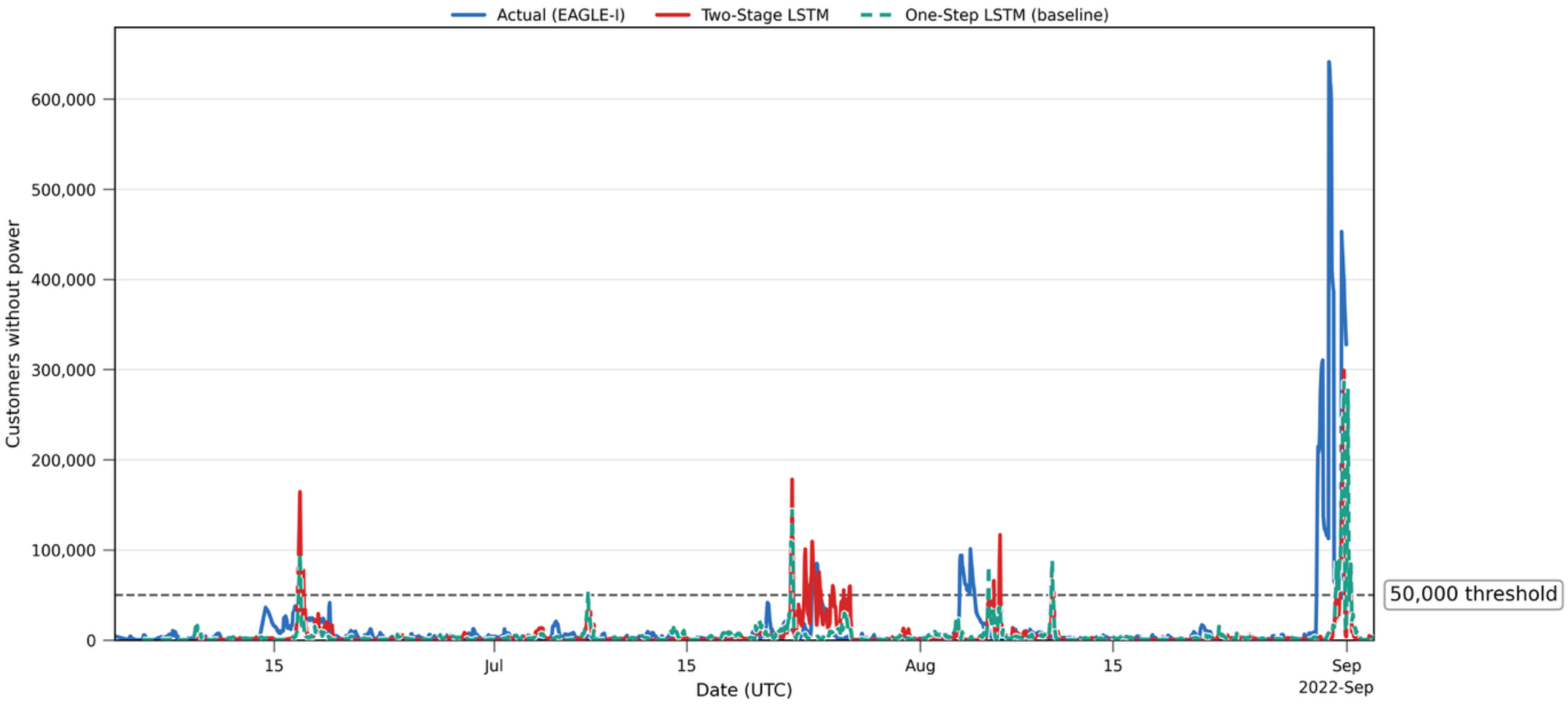

**Figure 5**. State-level outages vs. 48 h-ahead predictions on the 2022 test set (UTC). Blue - actual (EAGLE-I), red - Two-Stage LSTM, and green - One-Step LSTM baseline. Series are aligned at t+48 h. The gray dashed line indicates the 50,000 threshold.

## 5.3. Peak-conditional accuracy (cMASE)

We report cMASE on two scopes defined in Section 4.8: All-hours (full hourly grid) and available-hours (timestamps where Stage-2 predictions are preset). On the 2022 test sample, Stage-2 coverage is 65.7%. Results on the two scopes are nearly identical, indicating that coverage gaps do not drive the observed differences.

<u>All-hours:</u> The two-stage model is slightly better near peaks and slightly worse at broader windows. Table 6 gives the errors rates at different lead times.

**Table 6.** cMASE on All-hours (Test-2022), lower is better; Δ% = 100 × (1 - Two-Stage/Baseline)

| Δ window | Two-Stage | Baseline (One-Step) | Δ% (Two-Stage vs. Baseline) |
|---|---|---|---|
| ± 0 h | 12.39 | 12.69 | 2.4% |
| ± 6 h | 9.04 | 9.33 | +3.1% |
| ± 12 h | 7.33 | 7.49 | +2.1% |
| ± 24 h | 5.45 | 5.38 | -1.3% |
| ± 36 h | 4.49 | 4.38 | -2.5% |
| ± 48 h | 3.91 | 3.76 | -4.0% |

<u>Available-hours (65.7% coverage):</u> The same pattern holds, with slightly larger gains close to the peak as shown in Tabe 7.

**Table 7.** cMASE on Available-hours (65.7% coverage); Δ% = 100 × (1 - Two-Stage/Baseline)

| Δ window | Two-Stage | Baseline (One-Step) | Δ% (Two-Stage vs. Baseline) |
|---|---|---|---|
| ± 0 h | 8.93 | 9.15 | +2.5% |
| ± 6 h | 6.35 | 6.57 | +3.3% |
| ± 12 h | 5.06 | 5.18 | +2.3% |
| ± 24 h | 3.69 | 3.65 | -1.1% |
| ± 36 h | 3.12 | 3.04 | -2.6% |
| ± 48 h | 2.73 | 2.64 | -3.4% |

A moving-block bootstrap (168 h blocks, *B = 500*) confirms these patterns. Uncertainty bands are shown in Figure 6. Median cMASE shows small improvements at ± 6 -12 h (≈ +9-13%) and small degradations at ± 36-48 h (≈ -8-13%), with wide 95% percentile intervals [2.5; 97.5] reflecting the small number of peak hours. The same pattern holds on the Available-hours scope (Appendix. Fig. Detailed bootstrap summaries are provided in Appendix I (Tables I.14-I.15).

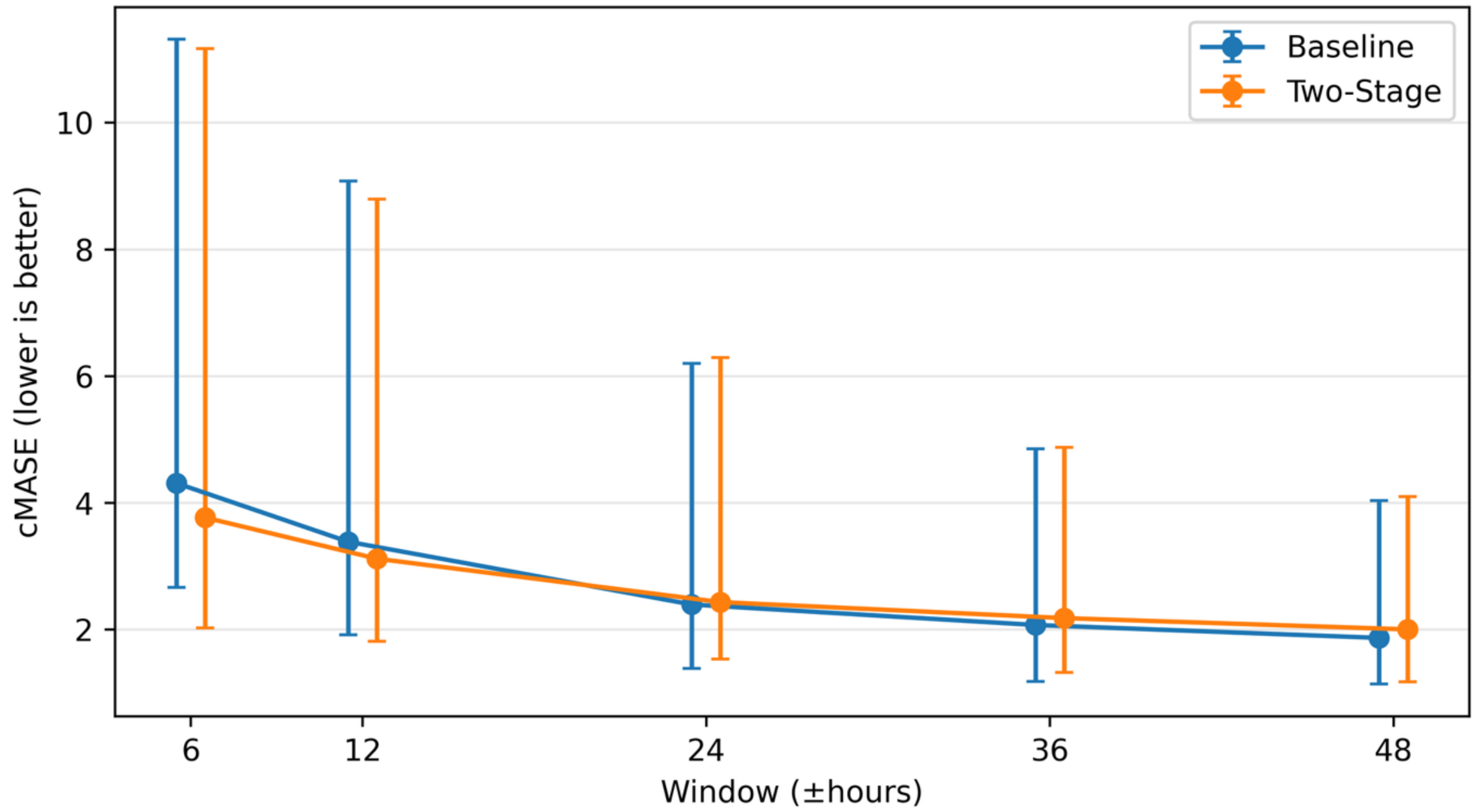

**Figure 6.** Peak-conditional cMASE on All-hours (Test 2022). Points bootstrap medians (moving-block, 168 h, B=500). 95% percentile intervals [2.5, 97.5]. Lower is better.

## 5.4. Event-level detection

We evaluate peak detection with the cluster-based peak detector (Section 5.4) and one-to-one matching within a tolerance window. On the 2022 test data, there are four reference peaks (Tables 8-9).

The summary of the deterministic results by windows is:
± 6 -12 h: Two-Stage detects 1/4 peaks (R=25%) vs. 0/4 for the baseline; precision is 20% (FA=4).

± 36 h: Two-Stage F1=44.44% vs. baseline 28.57%.
± 48 h: Two-Stage 3/4 hits (P=60%, R=75%, F1=66.67%) vs baseline 2/4 (P=66.67%, R=50%, F1=57.14%).

**Table 8.** Event-level detection for Two-Stage LSTM (counts and P/R/F1 by window).

| Δ h | Reference peaks | Predicted peaks | Hits | Miss | FA | P (%) | R (%) | F1 (%) |
|---|---|---|---|---|---|---|---|---|
| ±6 h | 4 | 5 | 0 | 3 | 4 | 20.00 | 25.00 | 22.22 |
| ±12 h | 4 | 5 | 1 | 3 | 4 | 20.00 | 25.00 | 22.22 |
| ±24 h | 4 | 5 | 1 | 3 | 4 | 20.00 | 25.00 | 22.22 |
| ±36 h | 4 | 5 | 2 | 2 | 3 | 40.00 | 50.00 | 44.44 |
| ±48 h | 4 | 5 | 3 | 1 | 2 | 60.00 | 75.00 | 66.67 |

**Table 9.** Event-level detection for One-Step LSTM (baseline), same format.

| Δ h | Reference peaks | Predicted peaks | Hits | Miss | FA | P (%) | R (%) | F1 (%) |
|---|---|---|---|---|---|---|---|---|
| ±6 h | 4 | 3 | 0 | 4 | 3 | 0.00 | 0.00 | - |
| ±12 h | 4 | 3 | 0 | 4 | 3 | 0.00 | 0.00 | - |
| ±24 h | 4 | 3 | 0 | 4 | 3 | 0.00 | 0.00 | - |
| ±36 h | 4 | 3 | 1 | 3 | 2 | 33.33 | 25.00 | 28.57 |
| ±48 h | 4 | 3 | 2 | 2 | 1 | 66.67 | 50.00 | 57.14 |

Bootstrap results (168 h blocks, $B=500$) are consistent with the deterministic view. At ±48 h, median recall is higher for the two-stage model (All-hours: 33% vs. 25%; Available-hours: 50% vs. 33%), but 95% intervals are wide (0-100%) because there are only four reference events. Full bootstrap results are available in Appendix I (Tables I.15.-I.16)

## 5.5. OE-417 Alignment

At peak hours, EO-417 alignment is incomplete and sometimes inconsistent (multi-day windows, missing magnitudes, “loss of monitoring/control”. Where magnitudes are available, Δ% is typically within ±10-30% for well-specified events, while several multi-day episodes show larger discrepancies (EAGLE-I lower than OE-417), consistent with reporting and coverage differences. As model targets come from EAGLE-I, these misalignments contribute to epistemic uncertainty. Nevertheless, the near-peak cMASE improvements remain on matching events (Appendix B, Tables B.3-B.4).

## 5.6. Limitations

This study has several limitations that affect performance and generalizability.

The training dataset covers only three months of hourly data, which restricts the model capacity. EDA indicates clear train-test differences and event variability. Adding more years would expose the model to a broader range of atmospheric conditions and likely improve reliability. A larger sample would also justify the implementation of more sophisticated algorithms.

The model is trained on Michigan data and reflects local infrastructural, demographic, and geographical context. Summer convection dominates the studied period, and the results do not directly generalize to other seasons. Application to other regions requires adaptation to local conditions.

The 50,000 customer outage peak definition and 90th-percentile anomaly label were chosen to match current early-warning objectives. Different operational tolerances for false alarms vs. misses will require resetting these thresholds and the Stage-1 decision point.

EAGLE-I reports can diverge from OE-417 (including "loss monitoring/control"), METAR coverage is sparse, and hourly aggregation smooths sharp gradients. These factors introduce epistemic (data) and aleatoric (storm randomness) uncertainty and helps explain moderate $R^2$ despite reasonable RMSE and near-peak cMASE gains.

Finally, the peak detector uses fixed parameters, with a moving-average window that merges over 24 hours and matching windows of ± {6, 12, 24, 36, 48} hours. Sensitivity exists and is reported separately.

## 5.7. Practical Implementation

Despite these constraints, the proposed summer thunderstorm model for Michigan can provide actionable early-warning signals using publicly available, aggregated EAGLE-I outage data (lacking fine spatial/temporal accuracy), METAR weather observations, and public demographic covariates. We deliberately exclude outage counts from Stage-2 (regression) because such signals may be delayed or missing during major failures. Stage-1 still uses their lags as anomaly indicators. This trade-off prioritizes inputs that remain reliable at critical periods.

On our reference setup (single NVIDIA Tesla T4, this dataset), training completes in approximately 15 minutes, and prediction for the test horizon is approximately 10 seconds, which is suitable for near-real-time dashboards. The approach requires neither satellite imagery nor heavy HPC resources and runs on standard infrastructure. It is compatible with preventive planning and risk screening for state and local agencies, grid operators, logistics, and insurance stakeholders.

For production, we recommend periodic (seasonal) retraining with drift monitoring (AUCPR, cMASE, coverage drift), region-specific retraining of Stage-1 thresholds, and optional integration of additional weather/infrastructure data where available. Reported latencies and accuracies will vary with data latency/quality, horizon length, and hardware.

## 6. Discussion

### 6.1. Effect of Stage-1 threshold on cascade performance

Strong class imbalance increases estimation variance because the minority class has few representative examples. The two-stage design is intended to reduce this epistemic uncertainty by filtering the majority class and passing a more balanced subset to the regressor. This strategy works only if the classifier keeps recall on peak precursors while reducing pass-through.

On the 2022 test set, the confusion matrix shows true positives of 1,790 and false negatives of 1,362 (Appendix I, Table I.5), indicating limited generalization (2021-2022 distribution shift or overfit). Even so, Stage-2 shows modestly lower cMASE within ± 0-12 h windows around peaks, while being slightly worse at broader windows. This is consistent with the regressor benefiting from reduced noise exposure around large events.

The operating threshold (0.70) was fixed on 2021 validation to balance recall vs. pass-through (PR trade-off (Table 5). The Stage-1 label intentionally uses a county-level 90th percentile cutoff to be inclusive. Given the complex spatiotemporal dynamics of state-level events, a single county-level criterion risks excluding true precursors. With limited covariates (no infrastructure features), an L1-regularized logistic classifier is a pragmatic gate/risk screen. Hence, we retain an inclusive threshold to keep potential contributors, rather than narrowing only state-level peaks. We recommend reporting the pass-through rate and recall at the operating point alongside AUCPR for transparency and a brief threshold sensitivity (e.g., ±0.05) to show stability.

### 6.2. Epistemic and Aleatoric Uncertainty

Convective storms introduce inherent aleatoric uncertainty (intrinsic variability of mesoscale phenomena and their spatial footprint). Epistemic uncertainty arises from fragmented data: sparse METAR coverage, hourly aggregation and spatial interpolation smooths sharp gradients that can obscure sharp gradients, occasional EAGLE-I inconsistencies with OE-417, missing values and monitoring losses during large-scale events, and a zero-inflated target. These factors constrain predictability and help explain moderate $R^2$ despite reasonable RMSE with improvements of cMASE gains near peaks. Therefore, we focus on event-centric metrics (hits/misses/false alarms) and cMASE within Δ± windows.

To mitigate these factors, we (i) validate EAGLE-I outages against OE-417; (ii) fit hourly variograms with finite search radii and preserved physically meaningful extremes via overdrafting; (iii) strictly causal temporal/spatial aggregates (lags/rolling stats; k-NN IDW). These steps increased linear signal and improved peak-window accuracy but do not eliminate uncertainty from sparse observations and reporting noise. For deployment, we recommend seasonal retraining with drift monitoring (AUCPR/cMASE and covariate drift) and modest regional recalibration of the Stage-1 threshold.

To reflect serial dependence and few reference peaks, we report medians and 95% percentile intervals [2.5%, 97.5%] from a moving-block bootstrap (weekly 16-h blocks: events re-detected per resample; cMASE denominator fixed from the full test). Intervals are wide for event metrics, consistent with the small number of matched peaks, while the near-peak cMASE pattern is stable (Section 5.3).

### 6.3. Feature contributions

#### 6.3.1. Stage-1

The L1-regularized logistic regression retained eight predictors (Table 17) out of the initial list of 98 features (full list in Table H.1), with the dominating variables being dew point lags (local and IDW) and a 6-hour temperature gradient with wind shifts. These features align with convective physics (moisture advection under falling pressure and wind shear), highlighting that the classifier elevates timestamps with precursors to thunderstorm-driven damage rather than routine outages.

#### 6.3.2. Stage-2 (LSTM regressor)

SHAP on 3,000 rows (2022 test set) shows that static proxies (population density, north-south coordinate, day-of-week) provide context, while meteorological predictors vary with the weather (Appendix J, Tables J.1-J.2). In the two-stage model, weather features dominate global importance (e.g., IDW dew point lags, gust/wind maxima), while in the one-step LSTM baseline, demographic and geographic proxies rank higher. This shift is consistent with Stage-1 filtering, which aims to cut out routine periods. On true peaks (e.g., 24 July 2022), both architectures emphasize gusts/wind indicators, moisture, and pressure signals. On 10 August 2022, both raised a false alarm, but with contrasting SHAP profiles: the two-stage model is driven by moisture advection (IDW dew-point lags), while the one-step model is dominated by geographical proxy (latitude) and short-horizon gust aggregates. We view this as consistent with data/label ambiguity (sparse station coverage, aggregation) and the selection effect of the Stage-1. We avoid causal claims. Together, static proxy features help reduce epistemic uncertainty from spatial gaps and infrastructure heterogeneity, while spatiotemporal aggregates of meteorological variables capture aleatoric variability of convective storms.

## 7. Conclusions

We developed and evaluated a two-stage early-warning pipeline for 48-hour-ahead prediction of large, state-level outages in Michigan's summer season. Stage-1 is a lightweight logistic screen that limits the flow of routine hours, and Stage-2 is an LSTM regressor trained on the filtered subset. All scoring is performed on the +48 h aligned, state-aggregated series and focused on behavior near large peaks (≥ 50,000 customers).

On the 2022 test, overall errors are of the same order across models: the one-step baseline attains lower RMSE/MAE, while the two-stage model shows similar $R^2$. Peak-conditional accuracy (cMASE at the 50,000 threshold) shows modest but consistent gains for the Two-Stage near peaks (deterministic 2-3% cMASE at ±0-12 h) and small degradations at broader windows (≈ -3-4% at ±36-48 h). Results are nearly identical on All-hours and Available-hours (Stage-2 time coverage 65.7%), indicating that coverage gaps do not drive the difference.

Event-centric detection shows that Two-Stage improves hit/miss balance at wider tolerances (e.g., F1 = 66.7% vs. 57.1% at ±48 h, hits 3/4 vs. 2/4). Thus, the two-stage design more reliably identifies the timing of major state-level events, even though the baseline is slightly better on aggregate error metrics.

Uncertainty analysis with hourly block bootstrapping (168 h blocks, $B = 500$) confirms the near-peak pattern with median cMASE improvements at ± 6–12 h (≈ +9–13%) and small losses at ± 36–48 h (≈ - 8–13%). Percentile intervals are wide due to the small number of reference events (N=4), so estimates should be interpreted with caution. Nevertheless, the evidence suggests that Two-Stage provides a tangible advantage around peak hours.

Operationally, the approach is practical and low-cost. It relies on open meteorological and demographic data, trains in minutes on a single GPU, and runs predictions in seconds, suitable for dashboards. For deployment, we recommend seasonal training, threshold tuning to local risk tolerance, and drift monitoring (AUCPR, cMASE, and covariate drift). Given occasional mismatches between EAGLE-I and OE-417 reports, routine event-level reconciliation and clear communication of uncertainty are advisable.

Limitations include the coverage of one state, the summer period, and a short training history (3 months), which constrained statistical power and external generalization. Extending the training set across multiple years and adding key local-infrastructure covariates may further improve model accuracy and robustness.

In summary, a feature-centric, two-stage design improves near-peak detection for thunderstorm-driven outages under open-data constraints and provides a practical starting point for open-data early warning, which is suitable for dashboards.

## Acknowledgement

This work is based on the author's MSc thesis at the University of Stavanger, supervised by Prof. Seth D. Guikema. We thank the U.S. Department of Energy's EAGLE-I program for outage data and the Iowa Environmental Mesonet (IEM) for providing access to NOAA/NWS METAR observations. Any results and opinions in this paper are the responsibility of the authors, not these data providers.

## Author contributions (CRediT)

**Iryna Stanishevska:** Conceptualization, Data curation, Formal analysis, Investigation, Methodology, Software, Validation, Visualization, Writing - original draft, Writing – review & editing.
**Seth D. Guikema**: Supervision, Writing - review & editing.

## Data and Code Availability

The EAGLE-I outage data are publicly available from Oak Ridge National Laboratory: umbrella DOI https://doi.ccs.ornl.gov/ui/doi/436 and 2014 - 2022 DOI: https://doi.ccs.ornl.gov/ui/doi/435. METAR observations were retrieved from the Iowa Environmental Mesonet (IEM) ASOS-AWOS METAR archive (Iowa Environmental Mesonet, n.d.).

A minimal, reproducible code is available on GitHub and archived on Zenodo. Concept DOI (latest): https://doi.org/10.5281/zenodo.17303651. This study uses release v1.0.2.
The repository generates a synthetic hourly dataset and reproduces the main comparison figure (Actual vs. Two-Stage vs. One-Step baseline) without access to private data or model weights. Code is licensed under Apache License 2.0; the synthetic dataset in data/ is licensed under CC BY 4.0.
GitHub: https://github.com/IrynaStanishevska/peak-outage-forecasting-/

Release: https://github.com/IrynaStanishevska/peak-outage-forecasting-/releases/tag/v1.0.2

## Competing interests

The author declares no competing interests.

## Declaration of generative AI and AI-assisted technologies in the manuscript preparation process

During the preparation of this work, the author used ChartGPT (OpenAI) for grammar editing and formatting suggestions. All suggested changes were manually reviewed and accepted if relevant. ChartGPT and Colab autocomplete were used for code troubleshooting, such as identifying dependency conflicts, interpreting error messages. After using these tools, the author reviewed, rewrote and verified all code, cross-checked suggestions against official documentation, and takes full responsibility for the content of the published article. No data analysis, model training, results, or figures were generated by AI. All methods, code, and conclusions are the author's.

# Appendixes

## Appendix A. Data Overview and Detailed Tables

**Table A. 1.** Reported electric emergency incidents in Michigan, Summer 2021 (June–August).

Source: DOE OE-417 Annual Summaries, OCESER (2023).

| Event Began | Restoration | Number of Customers Affected | Alert Criteria | Event Type | Demand Loss |
|---|---|---|---|---|---|
| 2021-06-21 05:14:00 | 2021-06-21 19:00:00 | 151,852 | Loss of electric service to more than 50,000 customers for 1 h or more. | Severe Weather | Unknown |
| 2021-06-29 16:00:00 | Not reported | 53,000 | Loss of electric service to more than 50,000 customers for 1 hour or more. | Severe Weather | Unknown |
| 2021-07-07 15:30:00 | Not reported | 90,000 | Loss of electric service to more than 50,000 customers for 1 hour or more. | Severe Weather | Unknown |
| 2021-07-24 20:30:00 | Not reported | 225,949 | Loss of electric service to more than 50,000 customers for 1 hour or more. | Severe Weather | Unknown |
| 2021-08-10 22:30:00 | 2021-08-13 16:38:00 | 372,600 | Loss of electric service to more than 50,000 customers for 1 hour or more. | Severe Weather | Unknown |
| 2021-08-11 15:35:00 | Not reported | 700,000 | Loss of electric service to more than 50,000 customers for 1 hour or more. | Severe Weather | Unknown |
| 2021-08-24 21:37:00 | Not reported | 65,000 | Loss of electric service to more than 50,000 customers for 1 hour or more. | Severe Weather | Unknown |
| 2021-08-24 17:00:00 | 2021-08-26 14:07:00 | 84,987 | Loss of electric service to more than 50,000 customers for 1 hour or more. | Severe Weather | Unknown |

**Table A. 2.** Reported electric emergency incidents in Michigan, summer 2022 (June–August).

Source: DOE OE-417 Annual Summaries, OCESER (2023).

| Event Began | Restoration | Number of Customers Affected | Alert Criteria | Event Type | Demand Loss |
|---|---|---|---|---|---|
| 2022-06-13 23:54:00 | 2022-06-14 00:45:00 | 0 | Complete loss of monitoring or control capability at its staffed Bulk Electric System control center for 30 continuous minutes or more. | System Operations | Unknown |
| 2022-06-15 10:45:00 | 2022-06-15 13:45:00 | 0 | Complete loss of monitoring or control capability at its staffed Bulk Electric System control center for 30 continuous minutes or more. | System Operations | Unknown |
| 2022-06-27 17:07:00 | 2022-06-28 01:42:00 | 0 | Complete loss of monitoring or control capability at its staffed Bulk Electric System control center for 30 continuous minutes or more. | System Operations | Unknown |
| 2022-06-30 19:58:00 | 2022-06-30 20:50:00 | 0 | Complete loss of monitoring or control capability at its staffed Bulk Electric System control center for 30 continuous minutes or more. | System Operations | Unknown |
| 2022-07-23 20:45:00 | 2022-07-24 11:30:00 | 93,750 | Loss of electric service to more than 50,000 customers for 1 hour or more. | Severe Weather | Unknown |
| 2022-08-03 19:00:00 | 2022-08-05 08:53:00 | 71,000 | Loss of electric service to more than 50,000 customers for 1 hour or more. | Severe Weather | Unknown |
| 2022-08-03 17:00:00 | 2022-08-03 20:30:00 | 91,264 | Loss of electric service to more than 50,000 customers for 1 hour or more. | Severe Weather | Unknown |
| 2022-08-29 15:00:00 | 2022-09-01 09:00:00 | 197,740 | Loss of electric service to more than 50,000 customers for 1 hour or more. | Severe Weather | Unknown |
| 2022-08-29 17:50:00 | NaT | Unknown | Loss of electric service to more than 50,000 customers for 1 hour or more | Severe Weather | |

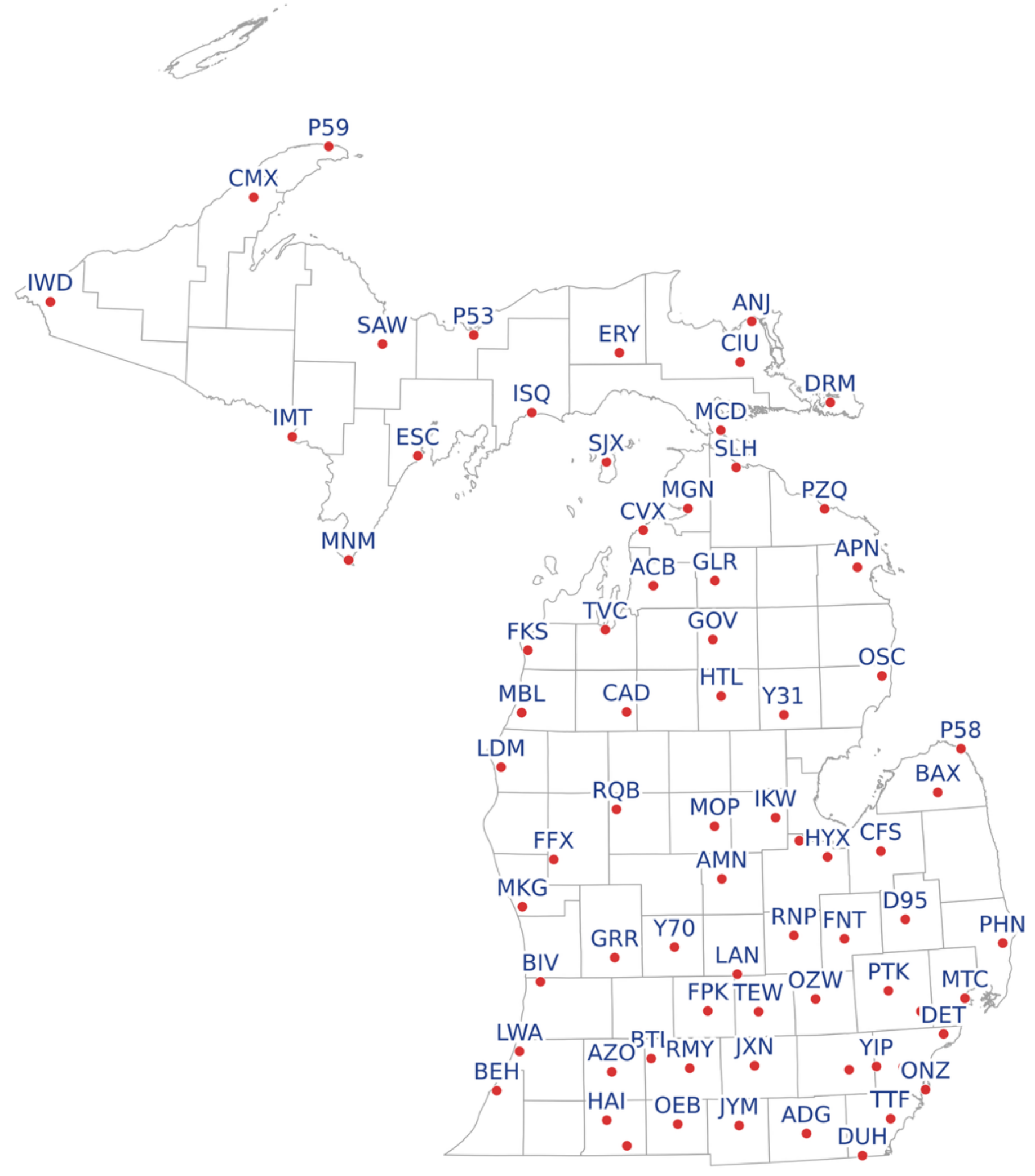

**Figure A 1.** Spatial distribution of airport (METAR) stations.

Points mark station locations over county boundaries; labels show a subset of station identifiers.

## Appendix B. Data Preprocessing

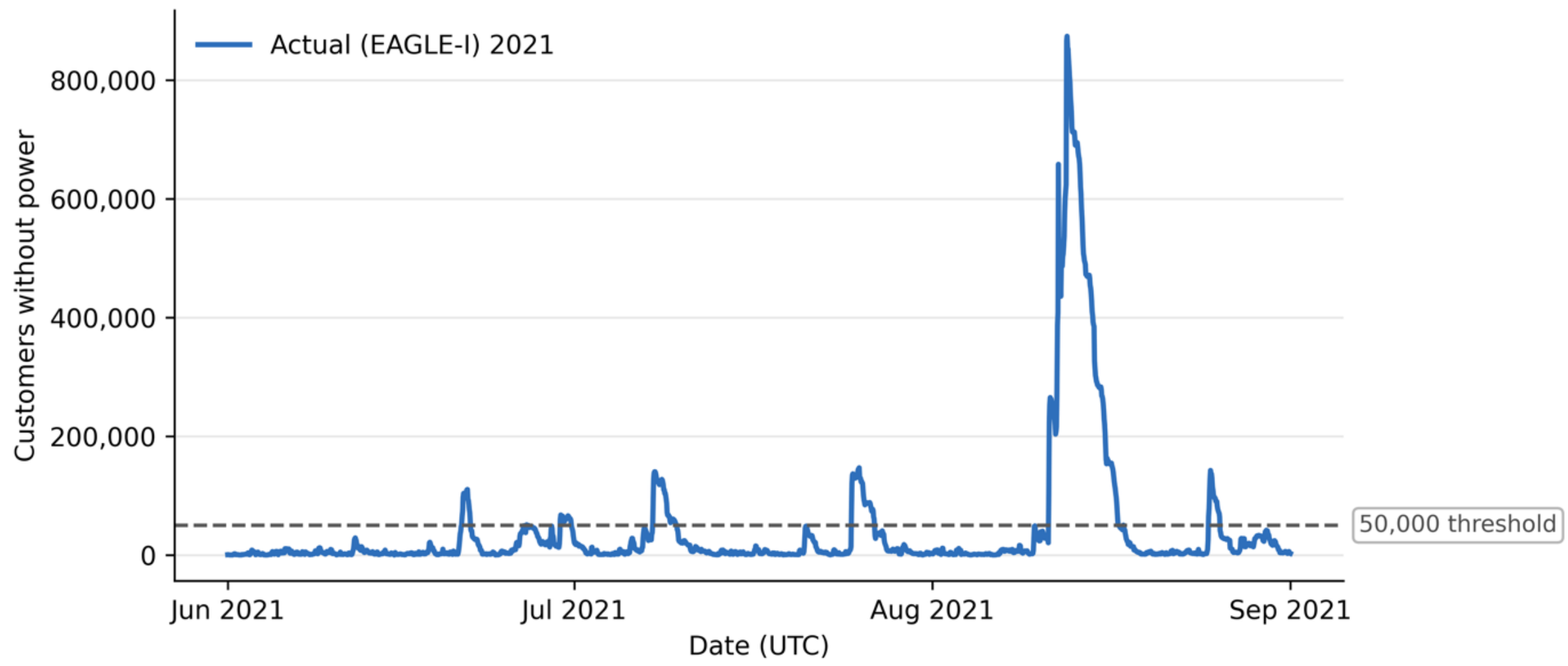

**Figure B 1.** Michigan state-level outages (EAGLE-I), Summer 2021, hourly after cleaning (UTC).

Y-axis: customers without power; X-axis: date-time. A dashed horizontal line marks the 50K threshold.

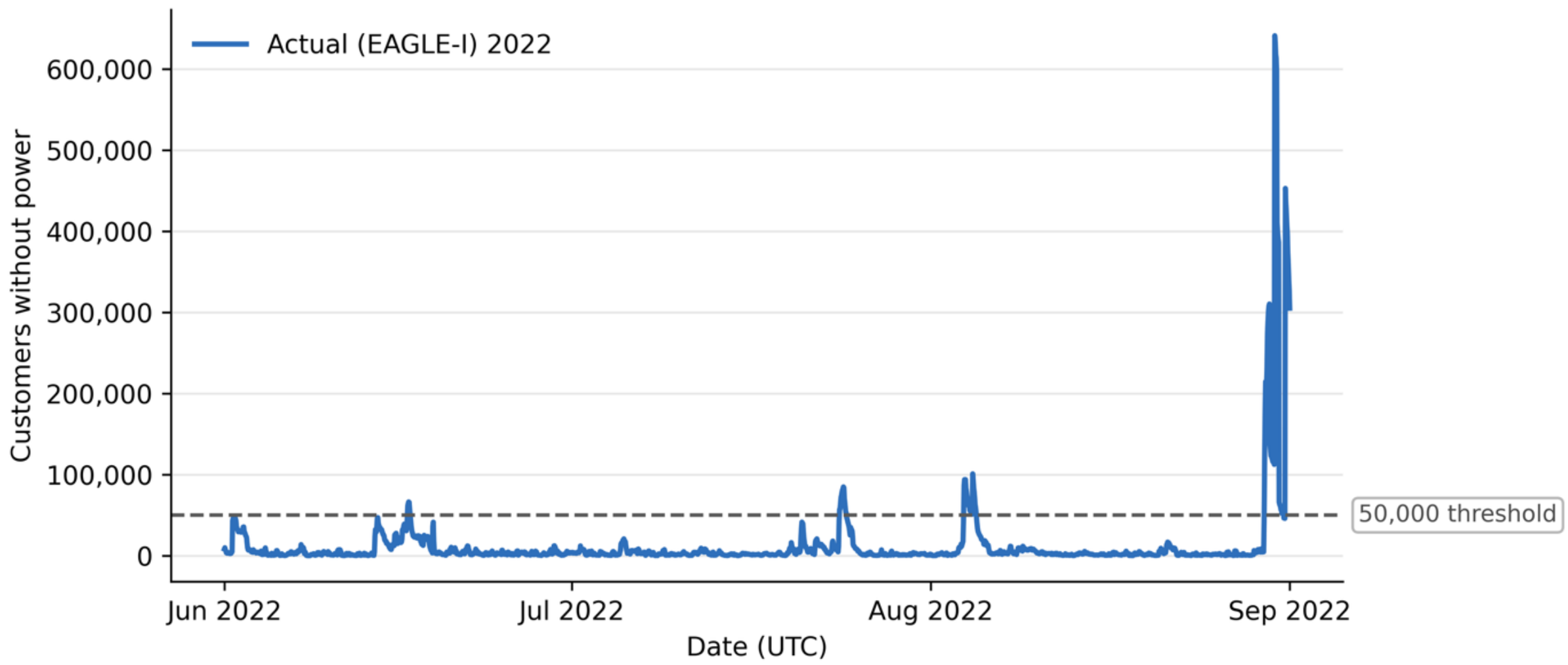

**Figure B 2.** Michigan state-level outages (EAGLE-I), Summer 2022, hourly after cleaning (UTC).

Y-axis: customers without power; X-axis: date-time. A dashed horizontal line marks the 50K threshold.

**Table B 1.** Raw Meteorological Data Summary Statistics

| Column | Count | Mean | SD | Min | 25% | 50% | 75% | Max |
|---|---|---|---|---|---|---|---|---|
| *tmpf* | 719,318 | 68.93 | 9.27 | 28.40 | 62.80 | 69.30 | 75.20 | 98.00 |
| *relh* | 719,123 | 74.02 | 19.40 | 0.93 | 59.29 | 77.34 | 91.09 | 100.00 |
| *dwpf* | 719,123 | 59.12 | 8.27 | -51.20 | 53.80 | 60.00 | 65.50 | 86.50 |
| *p01i* | 766,013 | 0.01 | 0.05 | 0.00 | 0.00 | 0.00 | 0.00 | 1.20 |
| *alti* | 1,987,574 | 29.98 | 0.13 | 19.65 | 29.99 | 29.99 | 30.08 | 30.36 |
| *mslp* | 151,277 | 1014.85 | 4.70 | 992.10 | 1011.80 | 1015.10 | 1018.10 | 1028.20 |
| *gust* | 155,899 | 18.55 | 4.19 | 1.00 | 16.00 | 18.00 | 21.00 | 123.00 |
| *sknt* | 2,152,068 | 5.23 | 4.01 | 0.00 | 2.00 | 5.00 | 8.00 | 43.00 |

**Notes:** Units: °F (tmpf, dwpf), % (relh), kt (sknt, gust), in (p0li), inHg (alti, ), hPa (mslp). IEM = Iowa Environmental Mesonet.

**Table B 2**. Aggregated METAR Data Summary Statistics

| Column | Count | Mean | SD | Min | 25% | 50% | 75% | Max |
|---|---|---|---|---|---|---|---|---|
| *tmpf* | 331,180 | 69.27 | 9.29 | 28.40 | 63.00 | 69.80 | 75.90 | 98.00 |
| *relh* | 328,008 | 72.30 | 19.19 | 0.93 | 57.66 | 74.79 | 89.22 | 100.00 |
| *dwpf* | 328,008 | 58.76 | 8.36 | -51.20 | 53.40 | 59.17 | 65.00 | 85.35 |
| *p01i* | 331,180 | 0.02 | 0.23 | 0.00 | 0.00 | 0.00 | 0.00 | 1.50 |
| *alti* | 328,758 | 29.98 | 0.13 | 24.65 | 29.89 | 29.99 | 30.08 | 30.36 |
| *mslp* | 148,131 | 1,014.87 | 4.70 | 992.10 | 1,011.90 | 1,015.10 | 1,018.20 | 1,028.20 |
| *gust* | 61,220 | 18.59 | 4.40 | 1.00 | 15.00 | 18.00 | 21.00 | 123.00 |
| *sknt* | 327,590 | 6.39 | 4.25 | 0.00 | 4.00 | 6.00 | 9.00 | 43.00 |

**Notes:** Units: °F (tmpf, dwpf), % (relh), kt (sknt, gust), in (p0li), inHg (alti, ), hPa (mslp). IEM = Iowa Environmental Mesonet.

**Table B 3.** OE-417 vs. EAGLE-I alignment (2021, UTC).

<table>
<thead>
<tr><th>Event window</th><th>Duration (h)</th><th>OE-417 customers</th><th>EAGLE-I customers</th><th>Δ% vs. OE-417</th><th>Confidence</th><th>Notes</th></tr>
</thead>
<tbody>
<tr><td>2021-06-21 05:14 - 2021-06-21 19:00</td><td>13:46</td><td>151,852</td><td>106,203</td><td>- 30.1%</td><td>Medium</td><td>The window indicates several hours; EAGLE-I is understated</td></tr>
<tr><td>2021-06-29 16:00 - Unknown</td><td>-</td><td>53,000</td><td>66,072</td><td>+24.7%</td><td>Medium</td><td>Restoration time is unknown; The discrepancy is moderate; EAGLE-I probably include several utilities</td></tr>
<tr><td>2021-07-07 15:30- Unknown</td><td>-</td><td>90,000</td><td>140,223</td><td>+55.8%</td><td>Low</td><td>Large discrepancy with unclear window</td></tr>
<tr><td>2021-07-24 20:30 Unknown</td><td>-</td><td>225,949</td><td>146,536</td><td>-35.1%</td><td>Medium</td><td>EAGLE-I shows fewer outages; possible gaps and incomplete coverage</td></tr>
<tr><td>2021-08-10 22:30 - 2021-08-13 16:38</td><td>66:08</td><td>372,600</td><td rowspan="2">873,852</td><td rowspan="2">-18.5%</td><td rowspan="2">Low</td><td rowspan="2">A multi-day event; summing up multiple EO-417 reports might distort the maximum EAGLE-I shows fewer outages</td></tr>
<tr><td>2021-08-11 15:35 - Unknown</td><td>-</td><td>700,000</td></tr>
<tr><td>2021-08-24 21:37 - Unknown</td><td>-</td><td>65,000</td><td rowspan="2">142,482</td><td rowspan="2">-5%</td><td rowspan="2">Low</td><td rowspan="2">Low discrepancy, but there is a multi-day event without concurrent sum in EO-417 reporting</td></tr>
<tr><td>2021-08-24 17:00 - 2021-08-26 14:07</td><td>45:07</td><td>84,987</td></tr>
</tbody>
</table>

Notes: Δ%=100x(EAGLE-I-OE-417)/OE-417; NA if the OE-417 magnitude is missing. For multi-day OE-417 entries, we use the maximum within the reported window. “Loss of monitoring/control” entries are flagged as Low.

**Table B 4.** OE-417 vs. EAGLE-I alignment (2022, UTC).

| Event window | Duration (h) | OE-417 customers | EAGLE-I customers | Δ% vs. OE-417 | Confidence | Notes |
|---|---|---|---|---|---|---|
| 2022-06-13 23:54 - 2022-06-14 00:45 | 0:51 | 0 | 47,111 | - | Low | OE-417 indicates alert with the loss of monitoring; EAGLE-I shows peak |
| 2022-06-15 10:45 - 2022-06-15 13:45 | 3:00 | 0 | 65,862 | - | Low | OE-417 indicates alert with the loss of monitoring; EAGLE-I shows peak |
| 2022-06-27 17:07 - 2022-06-28 01:42 | 8:35 | 0 | 5,758 | - | Low | OE-417 indicates alert with the loss of monitoring |
| 2022-06-30 19:58 - 2022-06-30 20:50 | 0:52 | 0 | 3,960 | - | Low | OE-417 indicates alert with the loss of monitoring |
| 2022-07-23 20:45 - 2022-07-24 11:30 | 14:45 | 93,750 | 84,936 | -9.4% | High | Normal convergence; EAGLE-I shows fewer outages |
| 2022-08-03 19:00 - 2022-08-05 08:53 | 37:53 | 71,000 | 101,057 | -37.7% | Low | Significant discrepancy; a multi-day event without concurrent sum in EO-417 reporting |
| 2022-08-03 17:00 - 2022-08-03 20:30 | 3:30 | 91,264 | | | | |
| 2022-08-29 15:00 - 2022-09-01 09:00 | 66:00 | 197,740 | 638,567 | - | Low | Significant peak in EAGLE-I; a multi-day event without concurrent sum in EO-417 reporting; Unknown number of customers and time of restoration |
| 2022-08-29 17:50 - Unknown | - | Unknown | | | | |

Notes: Δ%=100x(EAGLE-I-OE-417)/OE-417; NA if the OE-417 magnitude is missing. For multi-day OE-417 entries, we use the maximum within the reported window. “Loss of monitoring/control” entries are flagged as Low.

## Appendix C. Detailed Kriging Results by Parameter

### C.1. Air Temperature

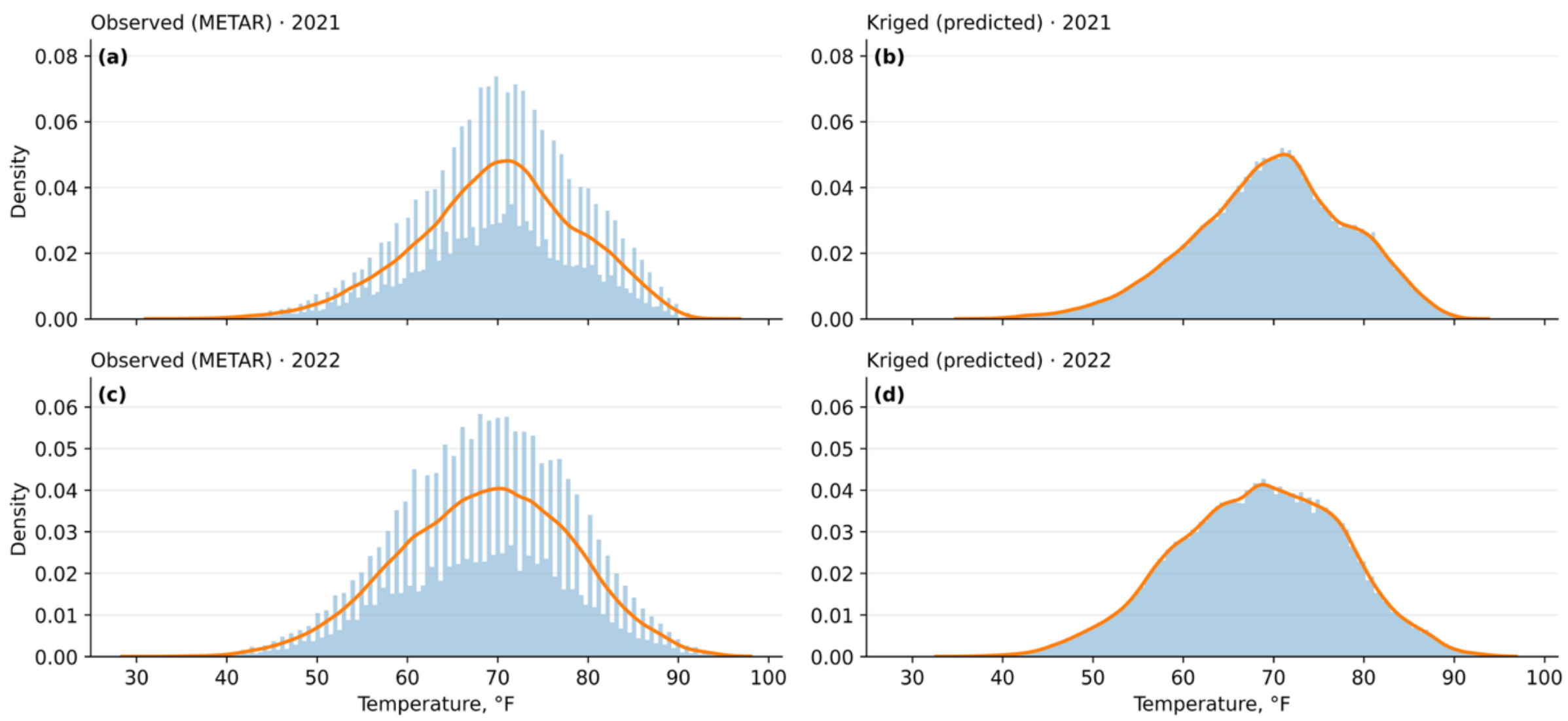

**Figure C.1. 1.** Air temperature (°F) in 2021-2022. Each panel shows a density histogram (Friedman-Diconis binning) with a Gaussian KDE overlay. (a) METAR observations, 2021; (b) Kriged to county centroids, 2021; (c) METAR observations, 2022; (d) Kriged to county centroids, 2022. The y-axis is probability density (area = 1).

**Table C.1. 1.** Descriptive Statistics (2021),
Air Temperature distribution before vs. after interpolation.

| Statistics | Initial Data | Interpolated Data |
|---|---|---|
| Mean | 69.96 | 69.61 |
| Standard Deviation | 8.95 | 8.67 |
| Minimum | 31.00 | 34.80 |
| 1st Quartile (25%) | 64.33 | 64.09 |
| Median (50%) | 70.23 | 70.03 |
| 3rd Quartile (75%) | 76.00 | 75.62 |
| Maximum | 96.8 | 93.77 |

**Table C.1. 2.** Descriptive statistics (2022),
Air Temperature distribution before vs. after interpolation.

| Statistics | Initial Data | Interpolated Data |
|---|---|---|
| Mean | 68.6 | 68.25 |
| Standard Deviation | 9.58 | 9.23 |
| Minimum | 24.4 | 32.61 |
| 1st Quartile (25%) | 62.00 | 61.85 |
| Median (50%) | 69.00 | 68.60 |
| 3rd Quartile (75%) | 75.37 | 74.99 |
| Maximum | 98.00 | 96.85 |

**Table C.1.3.** Descriptive statistics of kriging variance
(Air Temperature, in $(°F)^2$)

| Statistics | Kriging Variance (2021) | Kriging Variance (2022) |
|---|---|---|
| Mean | 4.28 | 4.53 |
| Standard Deviation | 3.39 | 3.87 |
| Minimum | 0.09 | 0.10 |
| 1st Quartile (25%) | 1.84 | 2.03 |
| Median (50%) | 3.41 | 3.76 |
| 3rd Quartile (75%) | 5.73 | 6.12 |
| Maximum | 31.84 | 30.59 |

**Table C.1.4.** Air Temperature (°F): Holdout
validation mean RMSE

| Station (type) | Mean ± SD | n dates |
|---|---|---|
| DTW (dense) | 2.06 ± 0.85 | 24 |
| JXN (median) | 2.67 ± 0.60 | 24 |
| IWD (isolated) | 7.18 ± 3.20 | 24 |

Note: Station columns indicate which station was
removed from the training set.

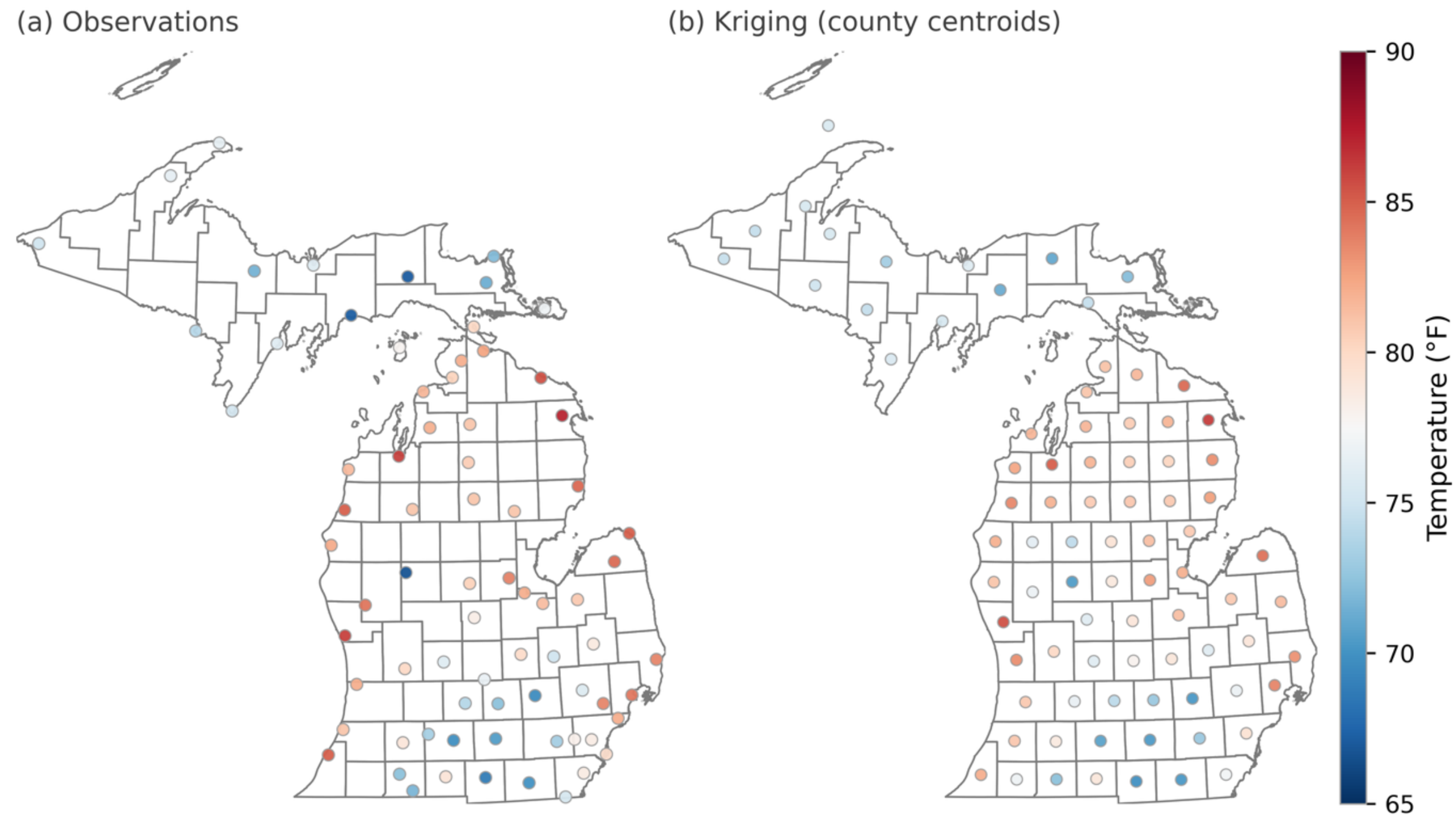

**Figure C.1.5**. Surface air temperature (°F) over Michigan at 11.08.2021 19:00 UTC. (a) METAR station observations; (b) kriging estimates at county centroids. County borders are shown in gray; the shared colorbar is on the right.

## C.2. Altimeter Pressure (alti)

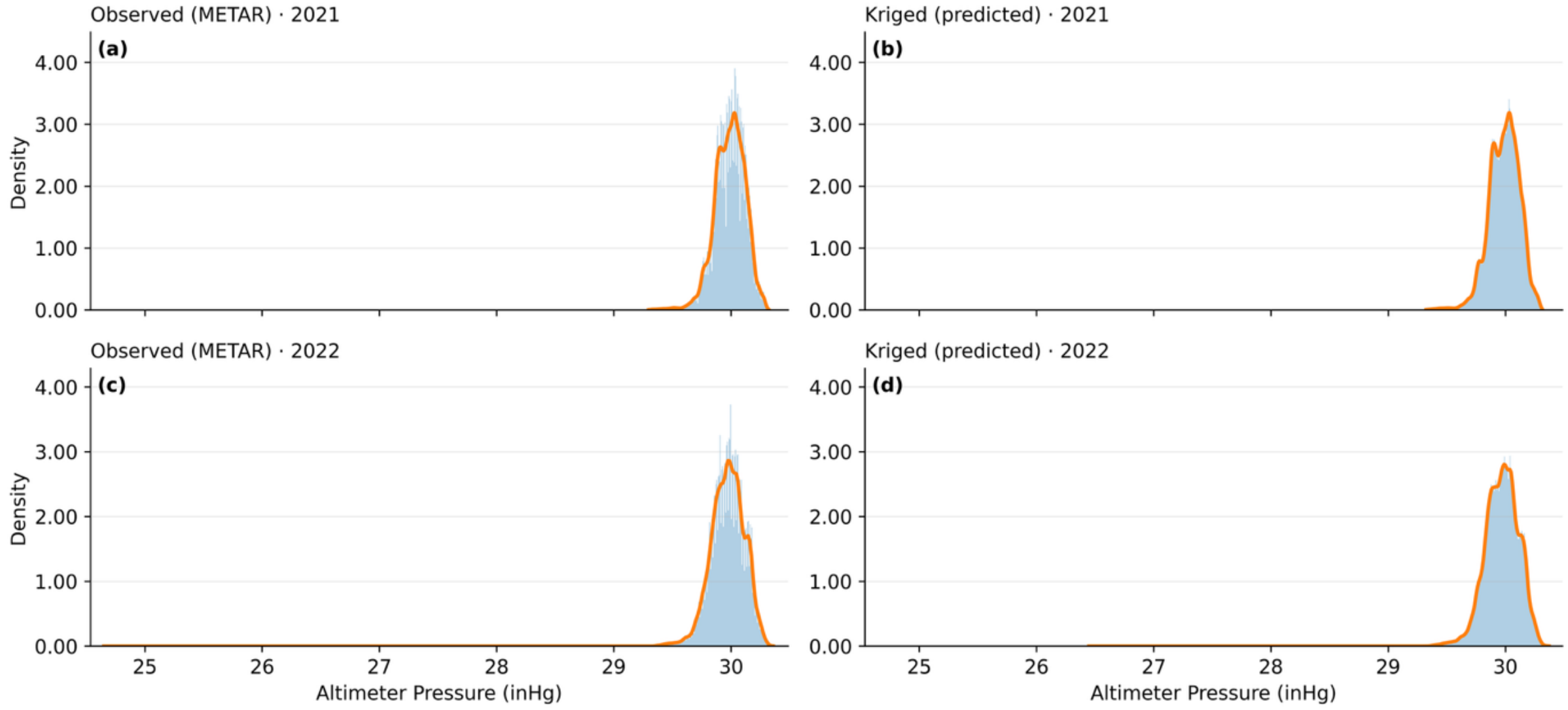

**Figure C.2.1.** Altimeter pressure (inHg) in 2021-2022. Each panel shows a density histogram (Friedman-Diconis binning) with a Gaussian KDE overlay. (a) METAR observations, 2021; (b) Kriged to county centroids, 2021; (c) METAR observations, 2022; (d) Kriged to county centroids, 2022. The y-axis is probability density (area = 1).

**Table C.2.1.** Descriptive statistics for Altimeter Pressure (2021), distribution before vs. after interpolation.

| Statistics | Initial Data | Interpolated Data |
|---|---|---|
| Mean | 29.99 | 29.99 |
| Standard Deviation | 0.13 | 0.13 |
| Minimum | 29.3 | 29.32 |
| 1st Quartile (25%) | 29.91 | 29.91 |
| Median (50%) | 30.00 | 30.00 |
| 3rd Quartile (75%) | 30.08 | 30.08 |
| Maximum | 30.32 | 30.32 |

**Table C.2.2.** Descriptive statistics for Altimeter Pressure (2022), distribution before vs. after interpolation.

| Statistics | Initial Data | Interpolated Data |
|---|---|---|
| Mean | 29.97 | 29.97 |
| Standard Deviation | 0.15 | 0.14 |
| Minimum | 24.65 | 24.45 |
| 1st Quartile (25%) | 29.88 | 29.89 |
| Median (50%) | 29.98 | 29.98 |
| 3rd Quartile (75%) | 30.07 | 30.07 |

| Maximum | 30.36 | 30.37 |
|---|---|---|

**Table C.2.3.** Descriptive Statistics of kriging variance (Altimeter Pressure, in ($Hg^2$)

| Statistics | Kriging Variance (2021) | Kriging Variance (2022) |
|---|---|---|
| Mean | 0.000093 | 0.000127 |
| Standard Deviation | 0.000080 | 0.001464 |
| Minimum | 0.000004 | 0.000003 |
| 1st Quartile (25%) | 0.000037 | 0.000040 |
| Median (50%) | 0.000074 | 0.00008 |
| 3rd Quartile (75%) | 0.000121 | 0.000133 |
| Maximum | 0.001319 | 0.138951 |

**Table C.2.4.** Altimeter Pressure (inHg): Holdout validation mean RMSE

| Station (type) | Mean ± SD | n dates |
|---|---|---|
| DTW (dense) | 0.0021 ± 0.0005 | 24 |
| JXN (median) | 0.0353 ± 0.6172 | 24 |
| IWD (isolated) | 0.0041 ± 0.260 | 24 |

Note: Station columns indicate which station was removed from the training set.

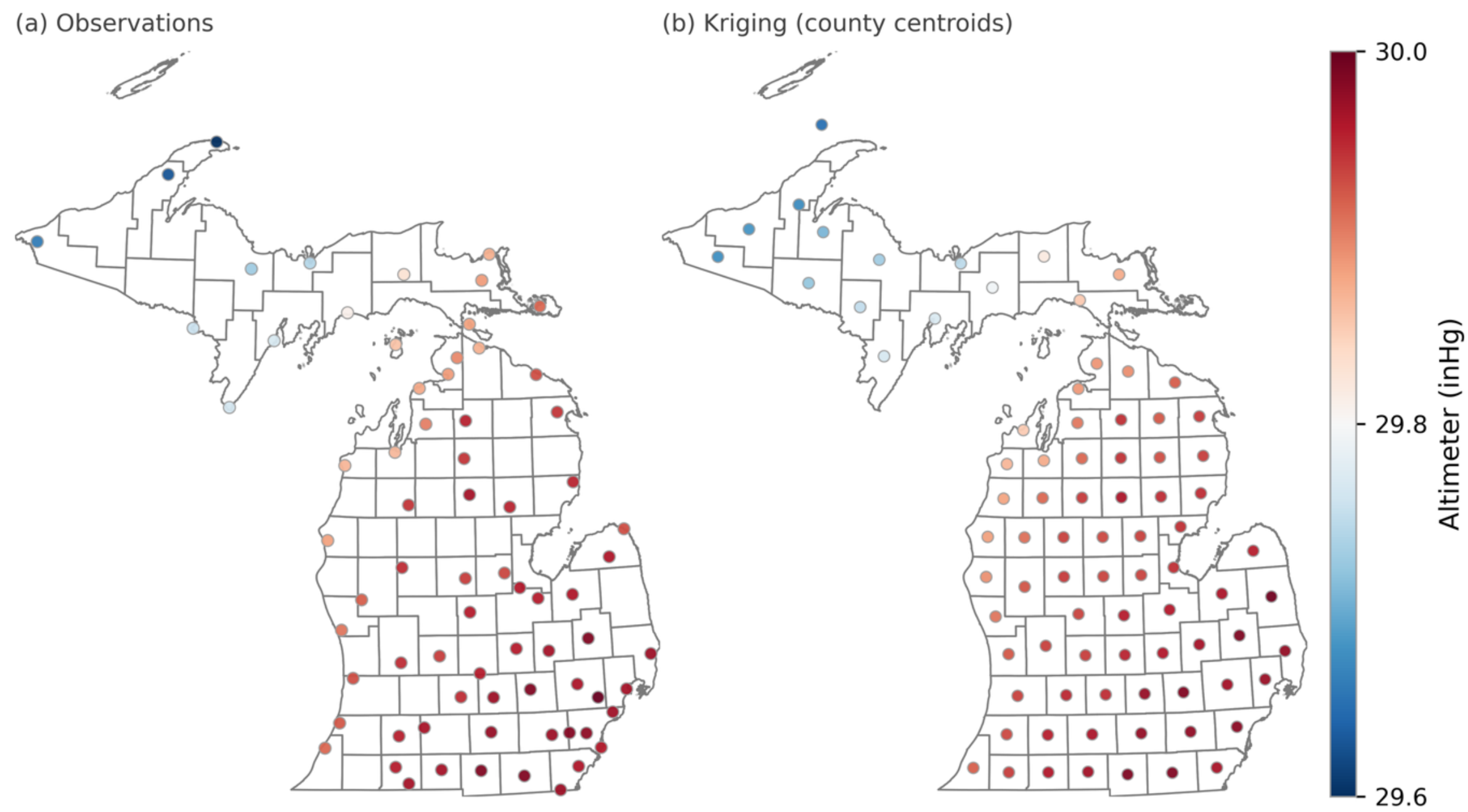

**Figure C.2.2**. altimeter pressure (inHg) over Michigan at 11.08.21 12:00 UTC. (a) METAR station observations; (b) kriging estimates at county centroids. County borders are shown in gray; the shared colorbar is on the right.

## C.3. Sea Level Pressure (mlsp)

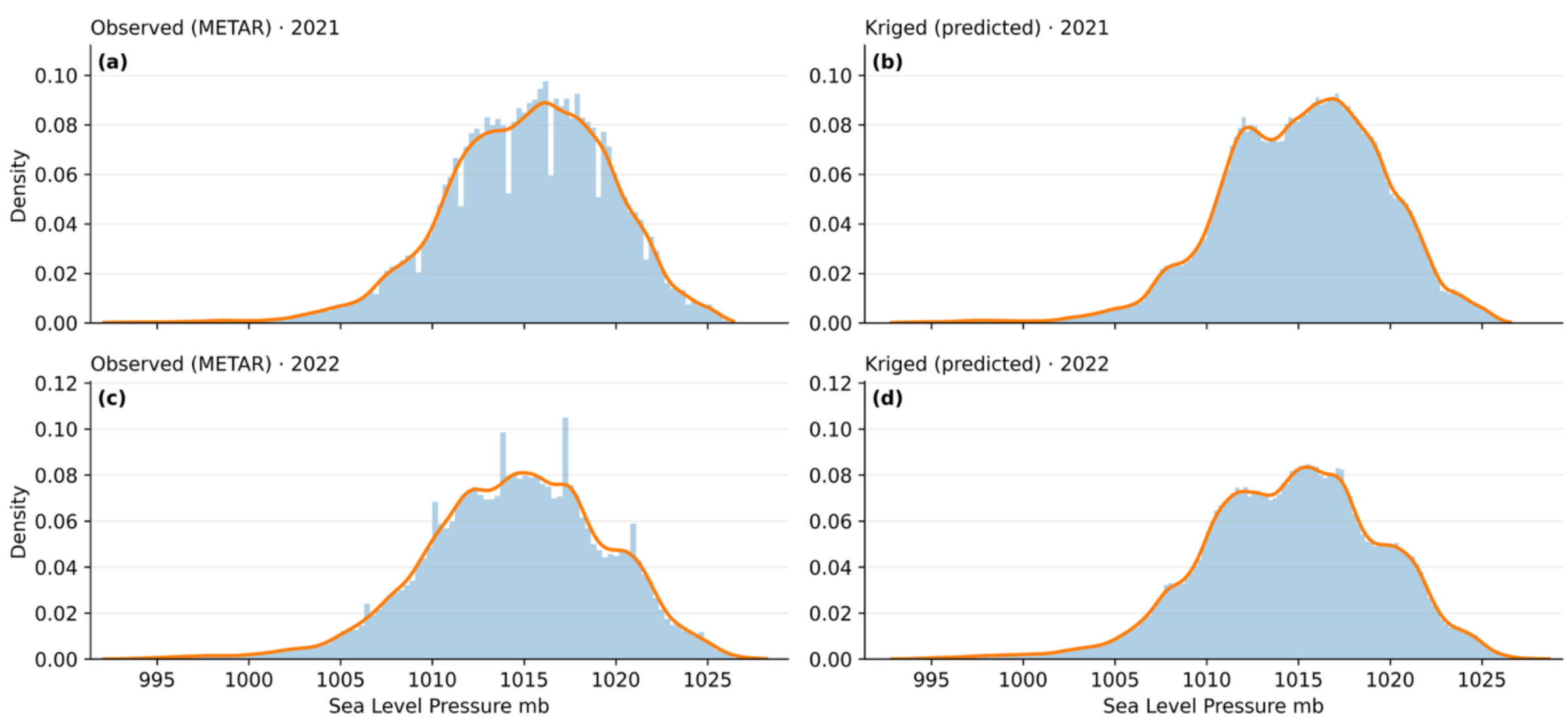

**Figure C.3.1.** Sea Level Pressure (mb) in 2021–2022. Each panel shows a density histogram (Friedman-Diconis binning) with a Gaussian KDE overlay. (a) METAR observations, 2021; (b) Kriged to county centroids, 2021; (c) METAR observations, 2022; (d) Kriged to county centroids, 2022. The y-axis is probability density (area = 1).

**Table C.3.1.** Descriptive Statistics for Sea Level Pressure (2021), distribution before vs. after interpolation.

| Statistics | Initial Data | Interpolated Data |
|---|---|---|
| Mean | 1015.17 | 1015.21 |
| Standard Deviation | 4.46 | 4.37 |
| Minimum | 992.10 | 992.84 |
| 1st Quartile (25%) | 1012.30 | 1012.28 |
| Median (50%) | 1015.50 | 1015.48 |
| 3rd Quartile (75%) | 1018.40 | 1018.31 |
| Maximum | 1026.40 | 1026.52 |

**Table C.3.2.** Descriptive statistics for Sea Level Pressure (2022), distribution before vs. after interpolation.

| Statistics | Initial Data | Interpolated Data |
|---|---|---|
| Mean | 1014.56 | 1014.61 |
| Standard Deviation | 4.910 | 4.82 |
| Minimum | 992.10 | 992.84 |
| 1st Quartile (25%) | 1011.40 | 1011.42 |
| Median (50%) | 1014.70 | 1014.82 |
| 3rd Quartile (75%) | 1017.90 | 1017.91 |
| Maximum | 1028.20 | 1028.65 |

**Table C.3.3.** Descriptive statistics of kriging variance (Sea Level Pressure, in $(mb)^2$)

| Statistics | Kriging Variance (2021) | Kriging Variance (2022) |
|---|---|---|
| Mean | 0.171 | 0.170 |
| Standard Deviation | 0.130 | 0.137 |
| Minimum | 0.004 | 0.002 |
| 1st Quartile (25%) | 0.091 | 0.085 |
| Median (50%) | 0.144 | 0.138 |
| 3rd Quartile (75%) | 0.215 | 0.218 |
| Maximum | 1.868 | 3.133 |

**Table C.3.4.** Sea Level Pressure (mb): Holdout validation mean RMSE

| Station (type) | Mean ± SD | n dates |
|---|---|---|
| DTW (dense) | 0.24 ± 0.12 | 24 |
| JXN (median) | 0.27 ± 0.14 | 24 |
| IWD (isolated) | 1.19 ± 0.83 | 24 |

Note: Station columns indicate which station was removed from the training set.

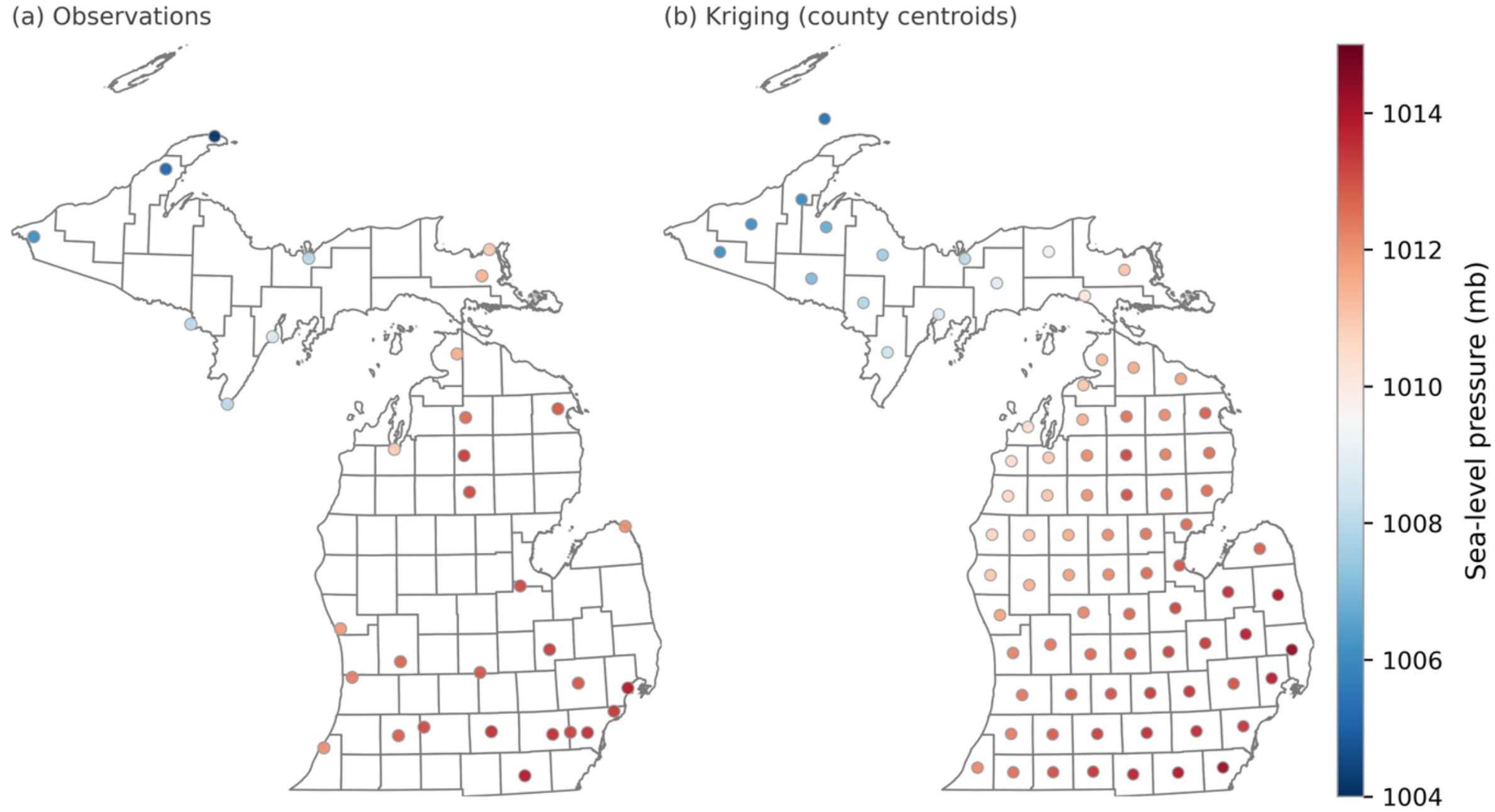

**Figure C.3.2**. Spatial map of Sea Level Pressure (mb) over Michigan at 11.08.21 12:00 UTC. (a) METAR station observations; (b) kriging estimates at county centroids. County borders are shown in gray; the shared colorbar is on the right.

## C.4. Relative Humidity (relh)

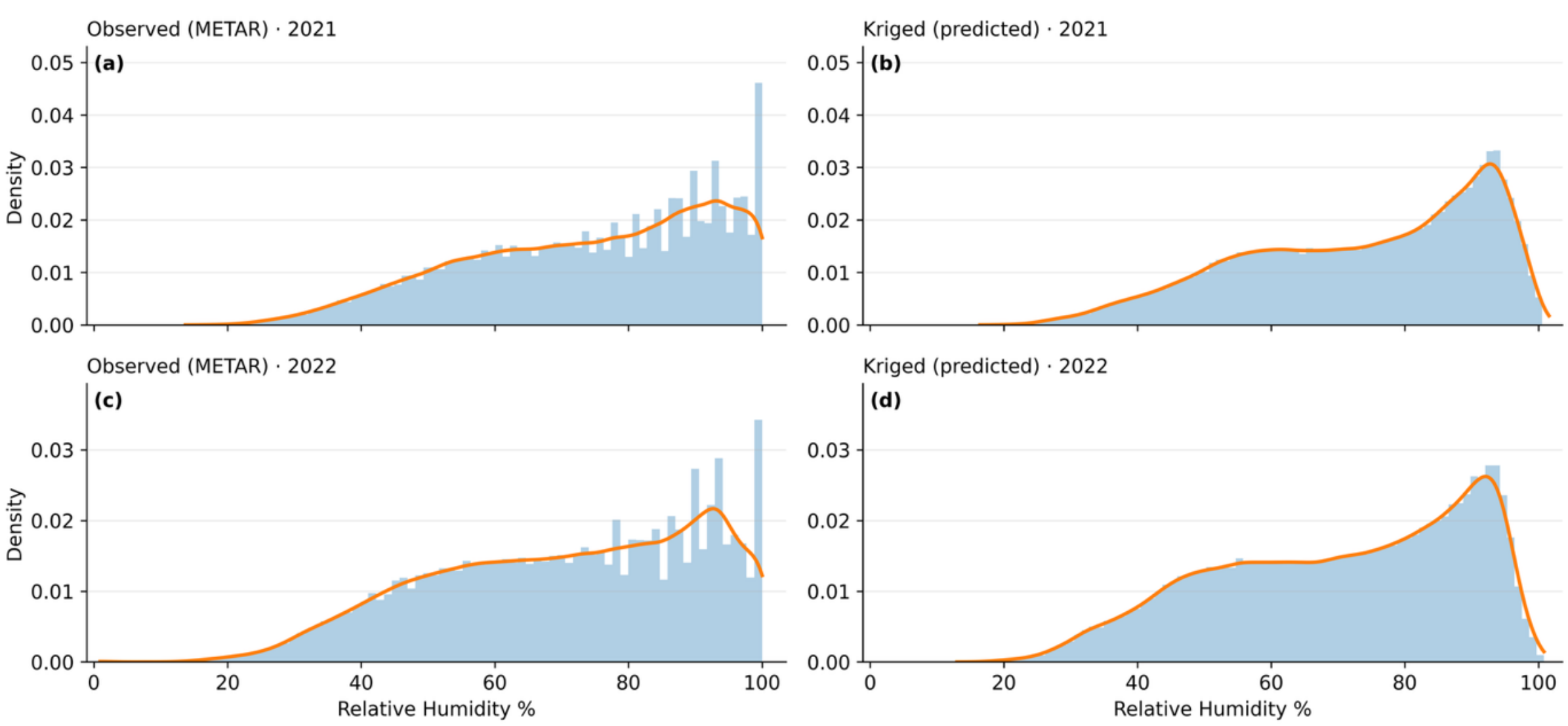

**Figure C.4.1.** Relative humidity (%) in 2021-2022. Each panel shows a density histogram (Friedman-Diconis binning) with a Gaussian KDE overlay. (a) METAR observations, 2021; (b) Kriged to county centroids, 2021; (c) METAR observations, 2022; (d) Kriged to county centroids, 2022. The y-axis is probability density (area = 1).

**Table C.4.1.** Descriptive Statistics (2021), Relative Humidity distribution before vs. after interpolation.

| Statistics | Initial Data | Interpolated Data |
|---|---|---|
| Mean | 74.09 | 74.19 |
| Standard Deviation | 18.44 | 16.13 |
| Minimum | 13.69 | 17.84 |
| 1st Quartile (25%) | 60.11 | 61.43 |
| Median (50%) | 76.79 | 77.47 |
| 3rd Quartile (75%) | 89.94 | 88.36 |
| Maximum | 100.00 | 99.65 |

**Table C.4.2.** Descriptive Statistics (2022), Relative Humidity before vs. after interpolation.

| Statistics | Initial Data | Interpolated Data |
|---|---|---|
| Mean | 70.51 | 70.48 |
| Standard Deviation | 19.76 | 17.07 |
| Minimum | 0.93 | 16.23 |
| 1st Quartile (25%) | 55.26 | 56.94 |
| Median (50%) | 72.65 | 72.70 |
| 3rd Quartile (75%) | 87.68 | 85.60 |
| Maximum | 100.00 | 99.60 |

**Table C.4.3.** Descriptive statistics of kriging variance (Relative Humidity, in $(\%RH)^2$)

| Statistics | Kriging Variance (2021) | Kriging Variance (2022) |
|---|---|---|
| Mean | 40.92 | 53.93 |
| Standard Deviation | 27.78 | 36.65 |
| Minimum | 1.15 | 1.23 |
| 1st Quartile (25%) | 21.54 | 27.43 |
| Median (50%) | 33.99 | 43.70 |
| 3rd Quartile (75%) | 52.43 | 68.06 |
| Maximum | 213.48 | 312.36 |

**Table C.4.4.** Relative Humidity (%): Holdout validation mean RMSE

| Station (type) | Mean ± SD | n dates |
|---|---|---|
| DTW (dense) | 7.01 ± 2.68 | 24 |
| JXN (median) | 7.06 ± 2.96 | 24 |
| IWD (isolated) | 10.56 ± 3.47 | 24 |

Note: Station columns indicate which station was removed from the training set.

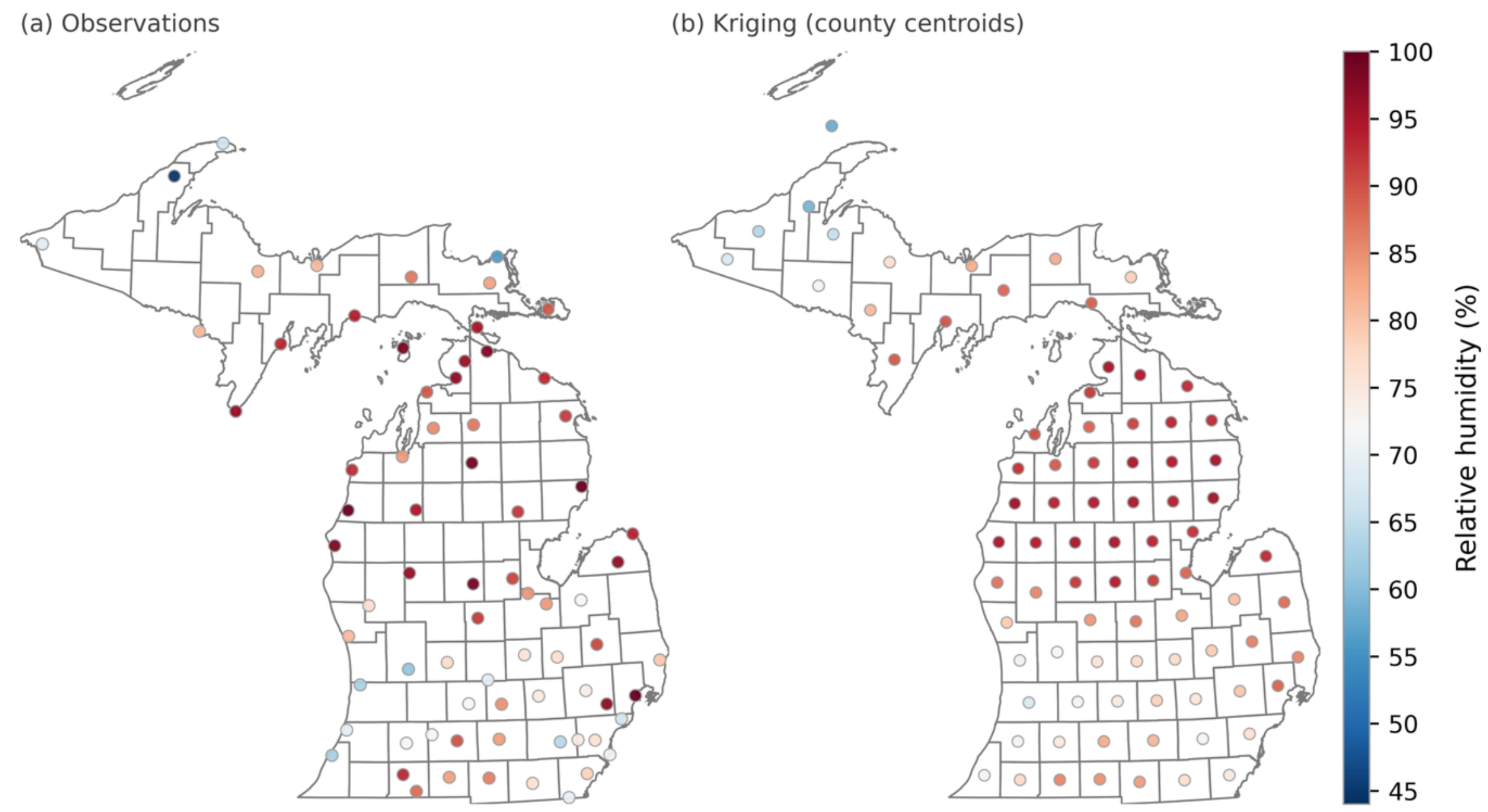

**Figure C.6.5**. Spatial map of relative humidity (%) over Michigan at 07.07.2021 20:00 UTC. (a) METAR station observations; (b) kriging estimates at county centroids. County borders are shown in gray; the shared colorbar is on the right.

**Table C.6.6.** Descriptive statistics relative humidity gradient (before interpolation)

| Statistics | Initial Data | Interpolated Data |
|---|---|---|
| Mean | 5.99 | 3.65 |
| Standard Deviation | 6.89 | 7.12 |
| Minimum | 0.00 | 0.00 |
| 1st Quartile (25%) | 0.34 | 0.00 |
| Median (50%) | 4.03 | 0.00 |
| 3rd Quartile (75%) | 8.88 | 4.70 |
| Maximum | 72.90 | 79.01 |

## C.5. Dew Point in Fahrenheit (dwpf)

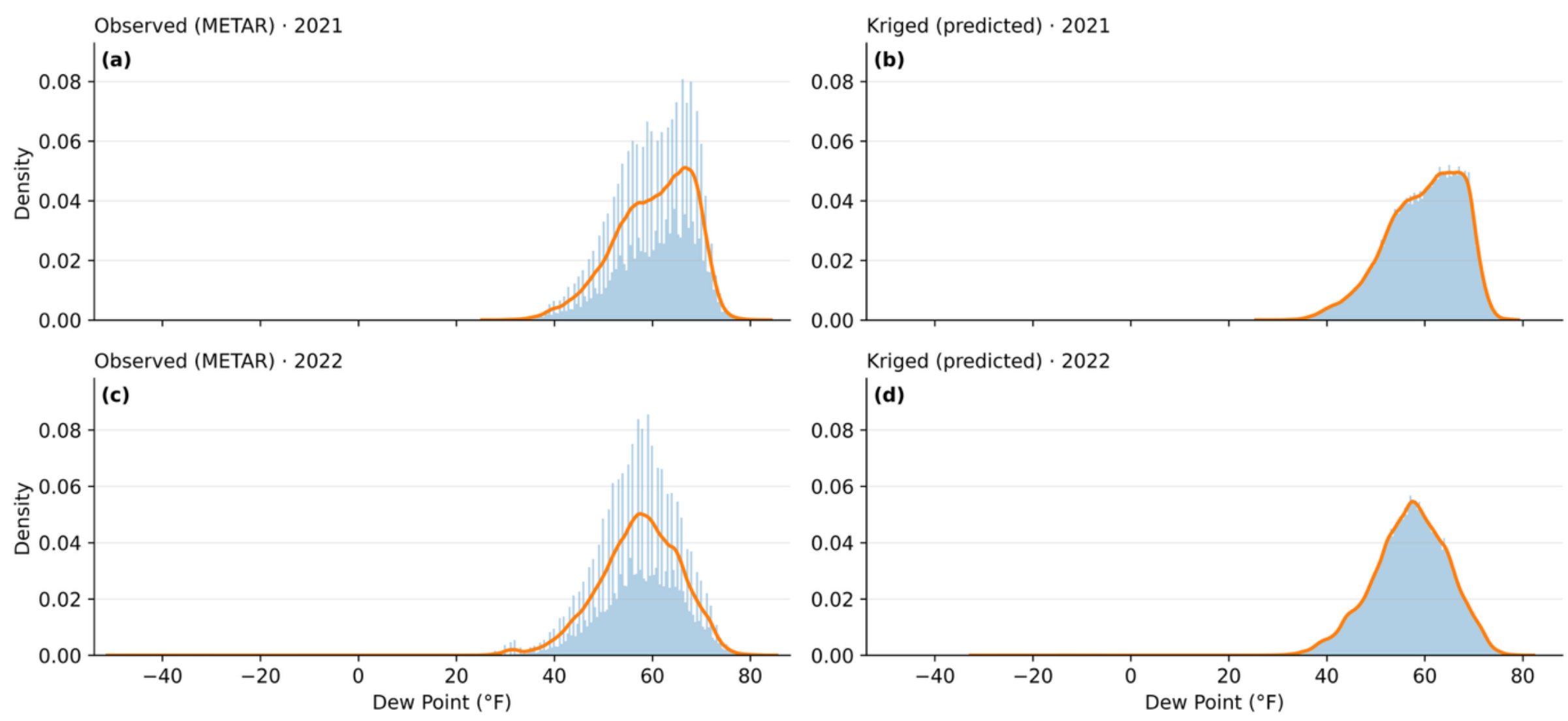

**Figure C.5.1.** Dew point (°F) in 2021-2022. Count histograms (Friedman-Diconis binning) with KDE scaled to counts: (a) METAR observations, 2021; (b) Kriged to county centroids, 2021; (c) METAR observations, 2022; (d) Kriged to county centroids, 2022. The y-axis is count; panels may have different sample sizes (N).

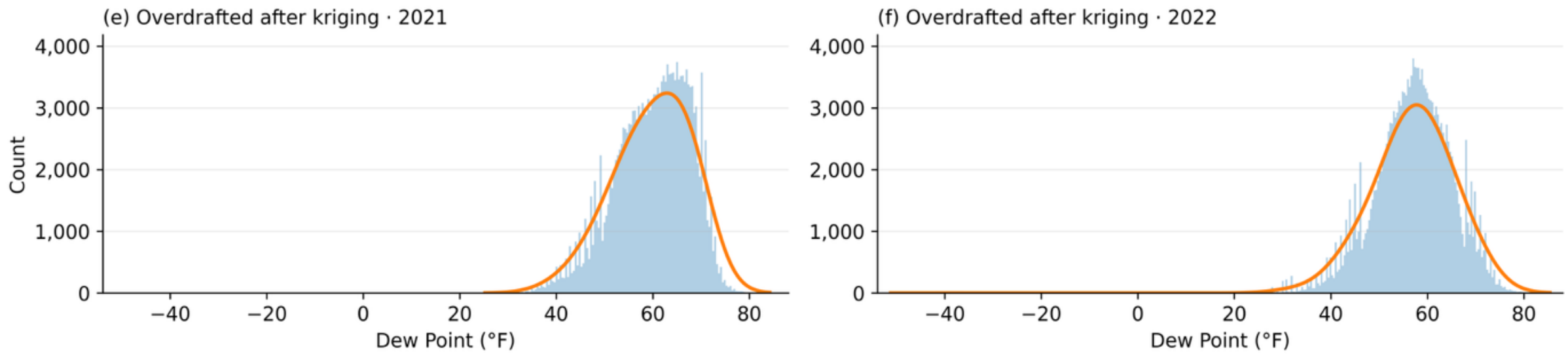

**Figure C.5.1.** Dew point (°F) overdrafted after kriging. Count histograms (Friedman-Diconis) with KDE scaled to counts. (e) 2021; (f) 2022. The y-axis is count; overdrafting preserves physically plausible extremes.

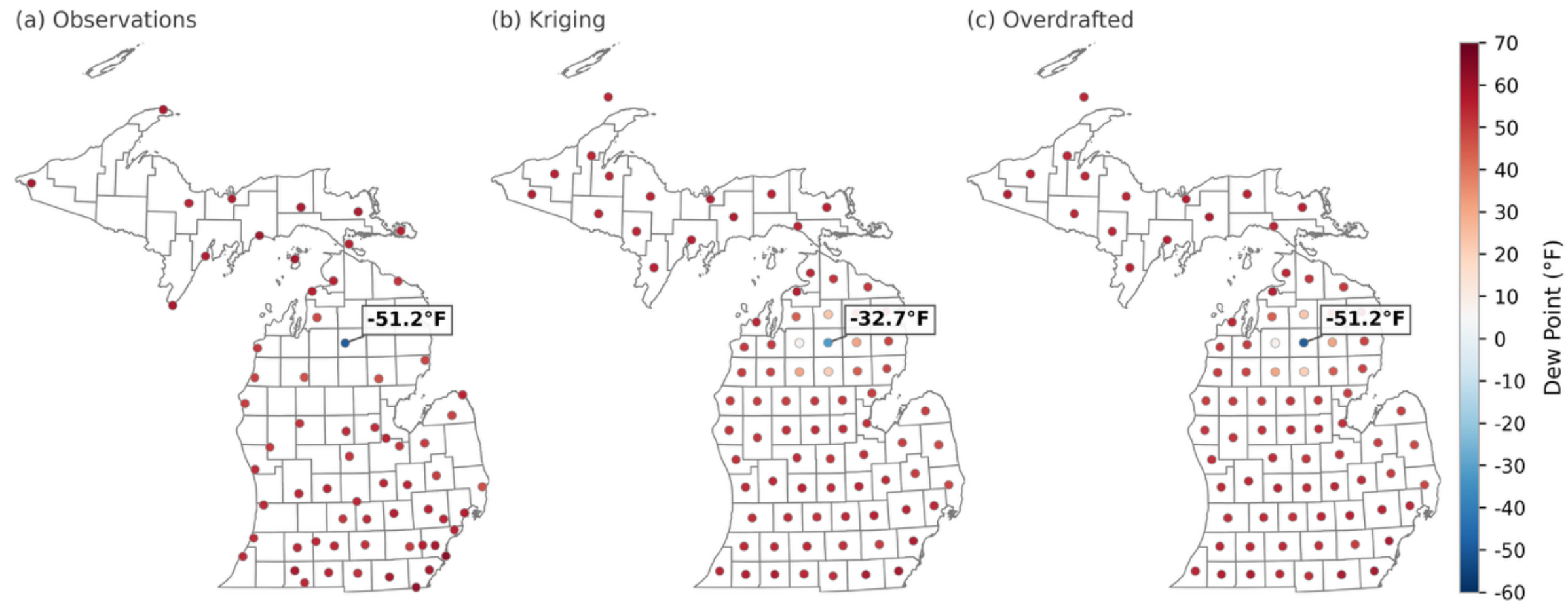

**Figure 1.** Dew Point (°F) at 10.08.2022 09:00. (a) METAR stations; (b) Kriging to county centroids; (c) Overdrafted after kriging. The most extreme value is annotated. Shared color scale.

**Table C.5.1.** Descriptive statistics for Dew Point in °F (2021).
Distribution before vs. after kriging, interpolation, and overdrafting.

| Statistics | Initial Data | Interpolated Data | Interpolated Data with Overdrafting |
|---|---|---|---|
| Mean | 60.28 | 59.88 | 59.91 |
| Standard Deviation | 7.94 | 7.62 | 7.80 |
| Minimum | 25.23 | 25.54 | 25.23 |
| 1st Quartile (25%) | 55.00 | 54.72 | 54.71 |
| Median (50%) | 61.10 | 60.82 | 60.82 |
| 3rd Quartile (75%) | 61.67 | 65.99 | 66.01 |
| Maximum | 84.20 | 79.08 | 84.20 |

**Table** C.5.2. Descriptive statistics for Dew Point in °F (2022).
Distribution before vs. after kriging, interpolation, and overdrafting.

| Statistics | Initial Data | Interpolated Data | Interpolated Data with Overdrafting |
|---|---|---|---|
| Mean | 57.26 | 57.06 | 59.88 |
| Standard Deviation | 8.50 | 7.70 | 8.22 |
| Minimum | -51.2 | -32.67 | -51.2 |
| 1st Quartile (25%) | 52.00 | 52.20 | 52.12 |
| Median (50%) | 57.87 | 57.41 | 57.39 |
| 3rd Quartile (75%) | 63.00 | 62.48 | 62.51 |
| Maximum | 85.35 | 82.16 | 85.35 |

**Table C.5.3.** Descriptive statistics of kriging variance, (Dew Point in $(°F)^2$)

| Statistics | Kriging Variance (2021) | Kriging Variance (2022) |
|---|---|---|
| Mean | 3.00 | 7.73 |
| Standard Deviation | 2.31 | 12.53 |
| Minimum | 0.07 | 0.08 |
| 1st Quartile (25%) | 1.32 | 2.83 |
| Median (50%) | 2.52 | 5.10 |
| 3rd Quartile (75%) | 3.10 | 8.69 |
| Maximum | 27.32 | 169.06 |

**Table C.5.4.** Dew point (°F): Holdout validation mean RMSE

| Station (type) | Mean ± SD | n dates |
|---|---|---|
| DTW (dense) | 1.32 ± 0.51 | 24 |
| JXN (median) | 1.69 ± 0.45 | 24 |
| IWD (isolated) | 3.36 ± 1.74 | 24 |

Note: Station columns indicate which station was removed from the training set.

## C.6. Wind Speed (sknt)

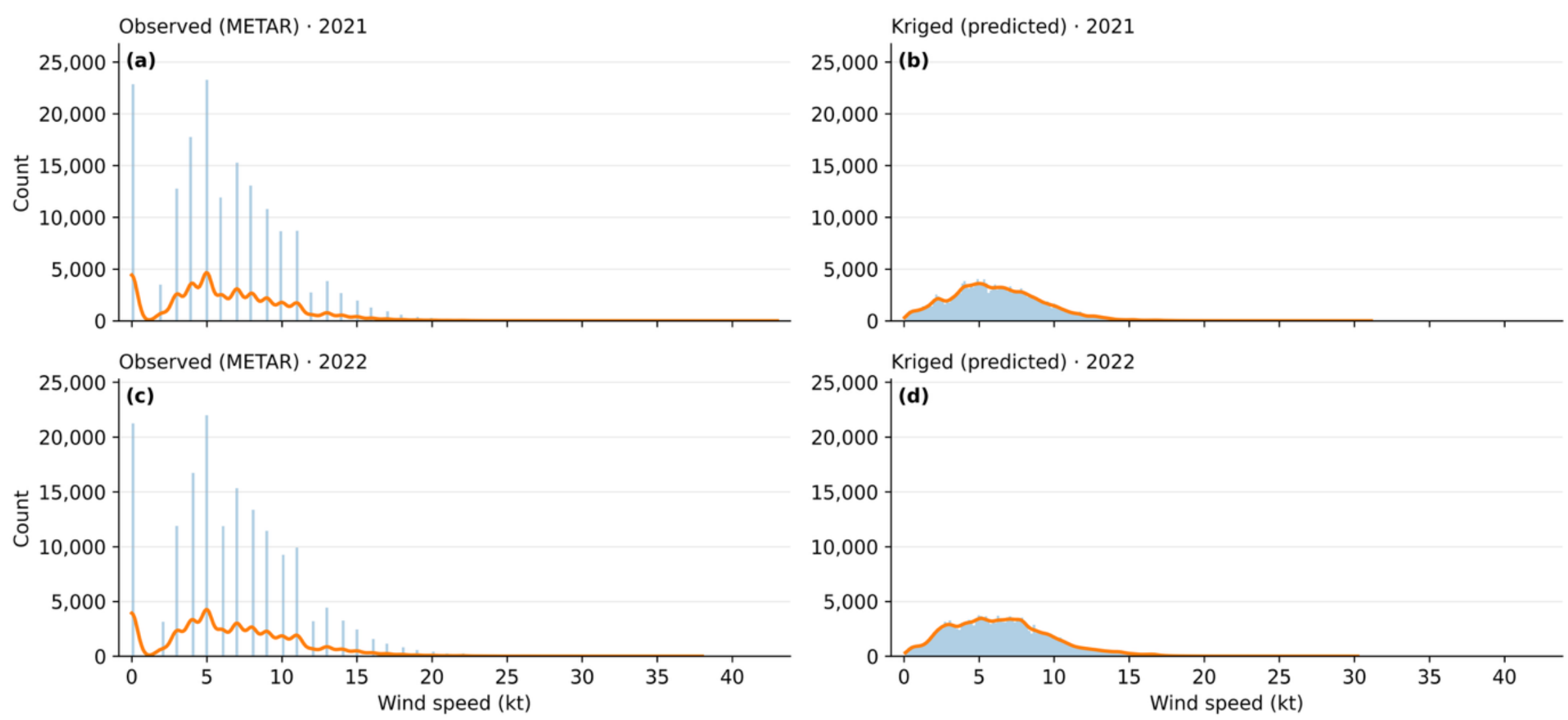

**Figure C.6.1.** Wind Speed (kt) in 2021-2022. Count histograms (Friedman-Diconis binning) with KDE scaled to counts: (a) METAR observations, 2021; (b) Kriged to county centroids, 2021; (c) METAR observations, 2022; (d) Kriged to county centroids, 2022. The y-axis is count; panels may have different sample sizes (N).

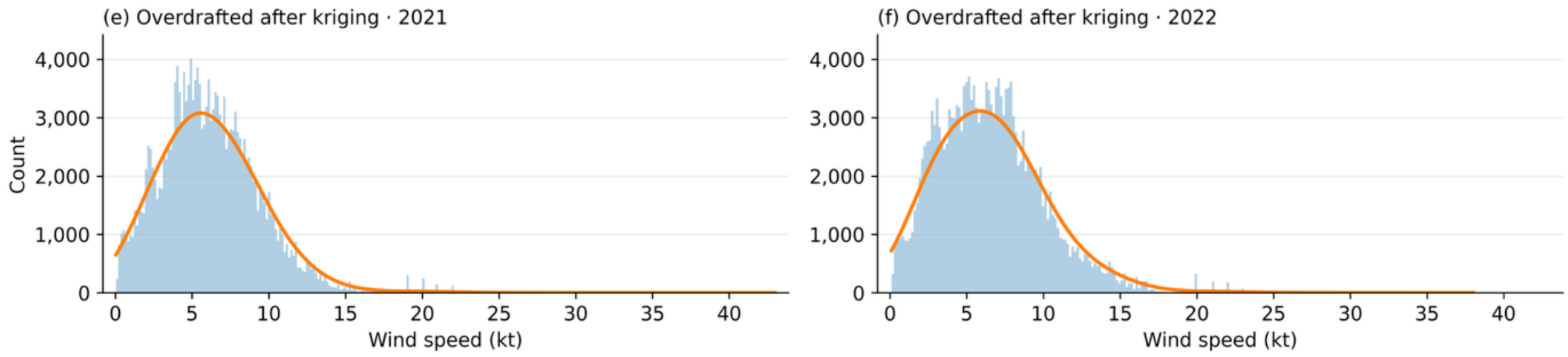

**Figure C.6.2.** Wind Speed (kt): overdrafted after kriging. Count histograms (Friedman-Diconis) with KDE scaled to counts. (e) 2021; (f) 2022. The y-axis is count; overdrafting preserves physically plausible extremes.

**Table C.6.1.** Descriptive Statistics for Wind Speed in knots (2021). Distribution before vs. after kriging, interpolation, and overdrafting.

| Statistics | Initial Data | Interpolated Data | Interpolated Data with Overdrafting |
|---|---|---|---|
| Mean | 6.21 | 6.12 | 6.16 |
| Standard Deviation | 4.15 | 3.06 | 3.19 |
| Minimum | 0.00 | 0.04 | 0.05 |
| 1st Quartile (25%) | 4.00 | 3.96 | 3.96 |
| Median (50%) | 6.00 | 5.90 | 5.90 |
| 3rd Quartile (75%) | 9.00 | 08.09 | 8.10 |
| Maximum | 43.00 | 31.12 | 43.00 |

**Table C.6.2.** Descriptive Statistics for Wind Speed in knots (2022), Distribution before vs. after kriging, interpolation, and overdrafting.

| Statistics | Initial Data | Interpolated Data | Interpolated Data with Overdrafting |
|---|---|---|---|
| Mean | 6.58 | 6.48 | 6.51 |
| Standard Deviation | 4.33 | 3.35 | 345 |
| Minimum | 0.00 | 0.95 | 0.95 |
| 1st Quartile (25%) | 4.00 | 3.96 | 3.96 |
| Median (50%) | 6.00 | 6.24 | 6.24 |
| 3rd Quartile (75%) | 9.00 | 8.53 | 8.53 |
| Maximum | 38.00 | 30.22 | 38.00 |

**Table C.6.3.** Descriptive statistics of kriging variance, (Wind Speed $(kt)^2$)

| Statistics | Kriging Variance (2021) | Kriging Variance (2022) |
|---|---|---|
| Mean | 4.83 | 4.93 |
| Standard Deviation | 3.54 | 3.30 |
| Minimum | 0.55 | 0.14 |
| 1st Quartile (25%) | 2.76 | 2.89 |
| Median (50%) | 4.19 | 4.32 |
| 3rd Quartile (75%) | 5.94 | 6.14 |
| Maximum | 61.14 | 43.57 |

**Table C.6.4.** Wind Speed (kt): Holdout validation mean RMSE

| Station (type) | Mean ± SD | n dates |
|---|---|---|
| DTW (dense) | 2.49 ± 0.61 | 24 |
| JXN (median) | 1.85 ± 0.71 | 24 |
| IWD (isolated) | 2.89 ± 0.97 | 24 |

Note: Station columns indicate which station was removed from the training set.

## C.7. Wind Direction (drct)

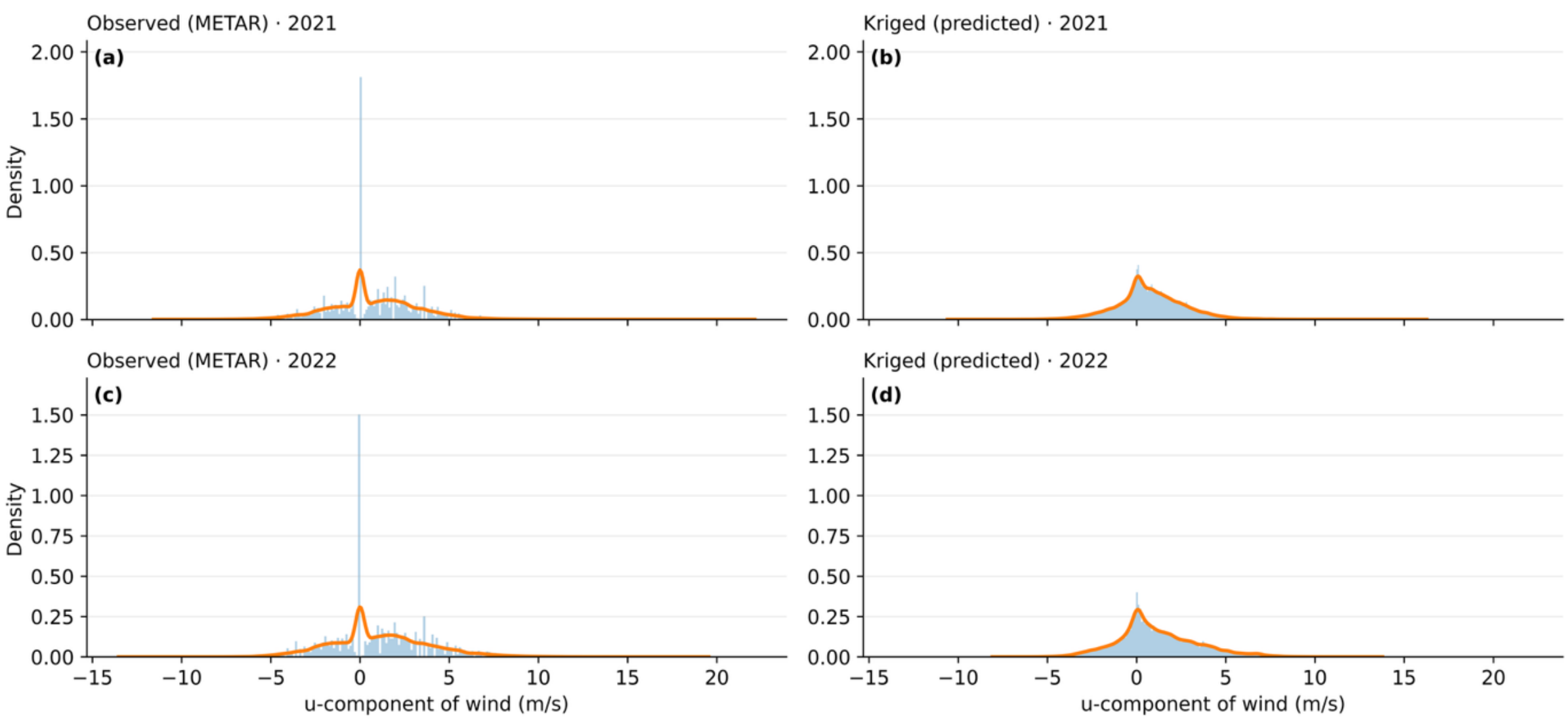

**Figure C.7.1.** u-component of wind (m/s) in 2021–2022. Each panel shows a density histogram (Friedman-Diconis binning) with a Gaussian KDE overlay. (a) METAR observations, 2021; (b) Kriged to county centroids, 2021; (c) METAR observations, 2022; (d) Kriged to county centroids, 2022. The y-axis is probability density (area = 1).

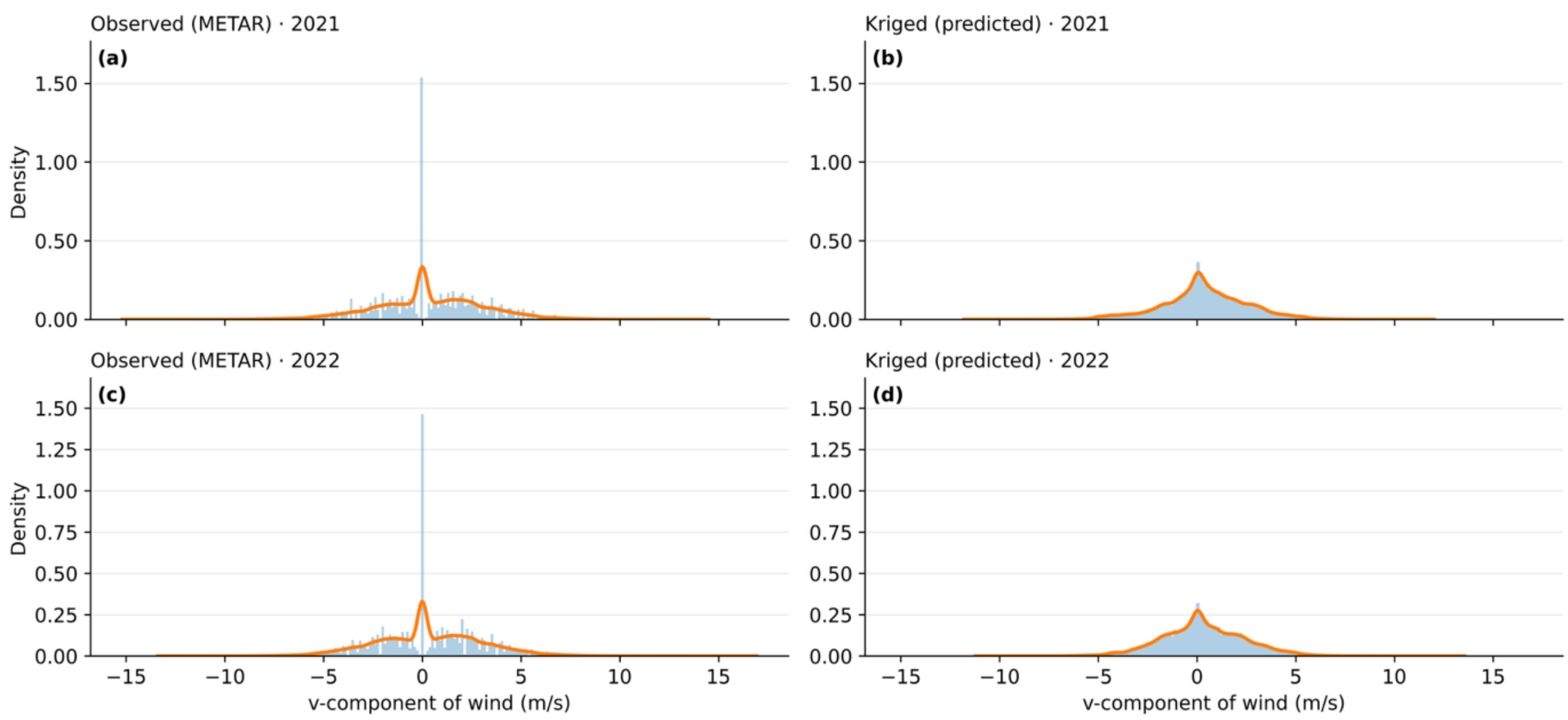

**Figure C.7.2.** v-component of wind (m/s) in 2021–2022. Each panel shows a density histogram

(Friedman-Diconis binning) with a Gaussian KDE overlay. (a) METAR observations, 2021; (b) Kriged to county centroids, 2021; (c) METAR observations, 2022; (d) Kriged to county centroids, 2022. The y-axis is probability density (area = 1).

**Table C.7.1.** Descriptive statistics for 'u' component of wind direction (2021), before vs. after interpolation.

| Statistics | Initial Data | Interpolated Data |
|---|---|---|
| Mean | 0.76 | 0.78 |
| Standard Deviation | 2.49 | 1.88 |
| Minimum | -11.57 | -10.60 |
| 1st Quartile (25%) | -0.63 | -0.25 |
| Median (50%) | 0.45 | 0.62 |
| 3rd Quartile (75%) | 2.31 | 1.88 |
| Maximum | 22.12 | 16.27 |

**Table C.7.2.** Descriptive statistics for 'u' component of wind direction (2022), before vs. after interpolation.

| Statistics | Initial Data | Interpolated Data |
|---|---|---|
| Mean | 2.75 | 1.12 |
| Standard Deviation | 2.75 | 2.22 |
| Minimum | -13.54 | -8.09 |
| 1st Quartile (25%) | -0.45 | -0.21 |
| Median (50%) | 0.79 | 0.75 |
| 3rd Quartile (75%) | 2.26 | 2.43 |
| Maximum | 19.55 | 13.78 |

**Table C.7.3.** Descriptive statistics for 'v' component of wind direction (2021), before vs. after interpolation.

| Statistics | Initial Data | Interpolated Data |
|---|---|---|
| Mean | 0.40 | 0.39 |
| Standard Deviation | 2.70 | 2.13 |

| Statistics | Initial Data | Interpolated Data |
|---|---|---|
| Minimum | -15.20 | -11.76 |
| 1st Quartile (25%) | -1.29 | -0.76 |
| Median (50%) | 0.00 | 0.29 |
| 3rd Quartile (75%) | 2.06 | 1.70 |
| Maximum | 14.50 | 12.00 |

**Table C.7.4.** Descriptive statistics for 'v' component of wind direction (2022), before vs. after interpolation.

| Statistics | Initial Data | Interpolated Data |
|---|---|---|
| Mean | 0.32 | 0.32 |
| Standard Deviation | 2.65 | 2.05 |
| Minimum | -13.39 | -11.21 |
| 1st Quartile (25%) | -1.40 | -1.00 |
| Median (50%) | 0.00 | 0.19 |
| 3rd Quartile (75%) | 2.03 | 1.67 |
| Maximum | 16.93 | 13.54 |

**Table C.7.5.** Descriptive statistics of kriging variance ('u' component of wind direction)

| Statistics | Kriging Variance (2021) | Kriging Variance (2022) |
|---|---|---|
| Mean | 1.46 | 1.46 |
| Standard Deviation | 1.33 | 1.19 |
| Minimum | 0.01 | 0.03 |
| 1st Quartile (25%) | 0.64 | 0.64 |
| Median (50%) | 1.11 | 1.13 |
| 3rd Quartile (75%) | 1.87 | 1.96 |
| Maximum | 18.27 | 12.47 |

**Table C.7.6.** Descriptive statistics of kriging variance

('v' component of wind direction)

| Statistics | Kriging Variance (2021) | Kriging Variance (2022) |
|---|---|---|
| Mean | 1.56 | 1.51 |
| Standard Deviation | 1.39 | 1.25 |
| Minimum | 0.02 | 0.03 |
| 1st Quartile (25%) | 0.66 | 0.69 |
| Median (50%) | 1.20 | 1.18 |
| 3rd Quartile (75%) | 2.04 | 2.00 |
| Maximum | 23.26 | 21.37 |

**Table C.7.7.** 'u' component of wind direction:
Holdout validation mean RMSE

| Station (type) | Mean ± SD | n dates |
|---|---|---|
| DTW (dense) | 1.16 ± 0.39 | 24 |
| JXN (median) | 1.24 ± 0.42 | 24 |
| IWD (isolated) | 2.07 ± 0.73 | 24 |

Note: Station columns indicate which station was removed from the training set.

**Table C.7.7.** 'v' component of wind direction:
Holdout validation mean RMSE

| Station (type) | Mean ± SD | n dates |
|---|---|---|
| DTW (dense) | 3.88 ± 1,75 | 24 |
| JXN (median) | 3.29 ± 1.25 | 24 |
| IWD (isolated) | 5.28 ± 1.68 | 24 |

Note: Station columns indicate which station was removed from the training set.

## Appendix D. Exploratory Data Analysis

**Table D.1**. Autocorrelation values (ACF) by county on the whole period (2021)

| No. | FIPS code | ACF lag1 | ACF lag6 | ACF lag12 | ACF lag18 | ACF lag24 | ACF lag36 | ACF lag48 |
|---|---|---|---|---|---|---|---|---|
| 1 | **26001** | 0.85 | 0.21 | 0.03 | -0.04 | -0.06 | -0.07 | 0.04 |
| 2 | **26003** | 0.72 | -0.02 | -0.06 | 0.03 | -0.07 | -0.07 | -0.06 |
| 3 | **26005** | 0.97 | 0.78 | 0.58 | 0.41 | 0.40 | 0.54 | 0.39 |
| 4 | **26007** | 0.87 | 0.26 | 0.20 | 0.10 | -0.12 | -0.19 | -0.16 |
| 5 | **26009** | 0.75 | 0.12 | 0.05 | 0.08 | 0.10 | -0.04 | -0.05 |
| 6 | **26011** | 0.98 | 0.89 | 0.75 | 0.60 | 0.46 | 0.17 | -0.01 |
| 7 | **26013** | 0.60 | -0.04 | -0.03 | -0.04 | -0.04 | -0.02 | -0.02 |
| 8 | **26015** | 0.94 | 0.65 | 0.40 | 0.18 | 0.25 | 0.41 | 0.18 |
| 9 | **26017** | 0.93 | 0.65 | 0.34 | 0.15 | 0.07 | -0.01 | -0.01 |
| 10 | **26019** | 0.74 | 0.28 | 0.10 | 0.02 | 0.02 | 0.09 | 0.02 |
| 11 | **26021** | 0.97 | 0.80 | 0.60 | 0.44 | 0.29 | 0.01 | -0.04 |
| 12 | **26023** | 0.99 | 0.95 | 0.88 | 0.79 | 0.71 | 0.53 | 0.40 |
| 13 | **26025** | 0.99 | 0.92 | 0.86 | 0.79 | 0.70 | 0.56 | 0.45 |
| 14 | **26027** | 0.93 | 0.56 | 0.34 | 0.21 | 0.10 | 0.01 | -0.04 |
| 15 | **26029** | 0.76 | 0.28 | 0.12 | -0.02 | -0.09 | -0.07 | -0.07 |
| 16 | **26031** | 0.68 | 0.13 | 0.07 | -0.01 | -0.05 | 0.01 | 0.05 |
| 17 | **26033** | 0.83 | 0.40 | 0.36 | 0.17 | 0.04 | 0.02 | -0.02 |
| 18 | **26035** | 0.96 | 0.75 | 0.54 | 0.36 | 0.19 | 0.04 | -0.09 |
| 19 | **26037** | 0.84 | 0.23 | -0.01 | -0.03 | 0.07 | -0.02 | -0.09 |
| 20 | **26039** | 0.76 | 0.53 | 0.41 | 0.26 | 0.21 | 0.03 | -0.06 |
| 21 | **26041** | 0.53 | -0.03 | -0.04 | 0.03 | 0.04 | 0.09 | -0.06 |
| 22 | **26043** | 0.92 | 0.63 | 0.27 | 0.03 | -0.05 | -0.05 | -0.10 |
| 23 | **26045** | 0.73 | 0.25 | 0.05 | 0.00 | -0.06 | 0.02 | 0.13 |
| 24 | **26047** | 0.87 | 0.40 | 0.27 | 0.07 | -0.02 | -0.05 | -0.07 |
| 25 | **26049** | 0.99 | 0.90 | 0.76 | 0.62 | 0.49 | 0.28 | 0.13 |
| 26 | **26051** | 0.98 | 0.86 | 0.70 | 0.57 | 0.44 | 0.23 | 0.05 |
| 27 | **26053** | 0.96 | 0.69 | 0.42 | 0.16 | 0.01 | -0.04 | -0.08 |
| 28 | **26055** | 0.94 | 0.64 | 0.40 | 0.17 | 0.12 | 0.18 | 0.04 |
| 29 | **26057** | 0.89 | 0.28 | 0.08 | -0.03 | -0.06 | -0.04 | 0.24 |
| 30 | **26059** | 0.99 | 0.95 | 0.89 | 0.82 | 0.75 | 0.61 | 0.46 |
| 31 | **26061** | 0.73 | 0.01 | -0.02 | -0.02 | -0.02 | -0.02 | 0.05 |
| 32 | **26063** | 0.96 | 0.78 | 0.55 | 0.34 | 0.23 | 0.08 | -0.04 |
| 33 | **26065** | 0.82 | 0.43 | 0.21 | 0.18 | 0.13 | 0.09 | 0.03 |
| 34 | **26067** | 0.93 | 0.63 | 0.37 | 0.16 | 0.16 | 0.30 | 0.25 |
| 35 | **26069** | 0.88 | 0.21 | 0.08 | -0.03 | -0.04 | 0.12 | 0.12 |
| 36 | **26071** | 0.70 | 0.25 | 0.15 | 0.11 | 0.01 | 0.08 | -0.15 |
| 37 | **26073** | 0.95 | 0.62 | 0.23 | 0.03 | 0.02 | 0.02 | -0.04 |
| 38 | **26075** | 0.99 | 0.92 | 0.85 | 0.83 | 0.78 | 0.63 | 0.46 |
| 39 | **26077** | 0.92 | 0.59 | 0.27 | 0.10 | 0.10 | 0.20 | 0.12 |
| 40 | **26079** | 0.93 | 0.52 | 0.16 | 0.04 | 0.04 | -0.02 | -0.04 |
| 41 | **26081** | 0.96 | 0.70 | 0.41 | 0.19 | 0.22 | 0.37 | 0.20 |
| 42 | **26083** | 0.87 | 0.21 | -0.20 | -0.16 | -0.05 | -0.09 | 0.00 |
| 43 | **26085** | 0.97 | 0.78 | 0.56 | 0.40 | 0.33 | 0.25 | 0.07 |
| 44 | **26087** | 0.96 | 0.80 | 0.62 | 0.49 | 0.34 | 0.19 | 0.05 |
| 45 | **26089** | 0.91 | 0.49 | 0.29 | 0.13 | 0.07 | 0.21 | 0.07 |
| 46 | **26091** | 0.97 | 0.85 | 0.74 | 0.70 | 0.62 | 0.43 | 0.25 |
| 47 | **26093** | 0.98 | 0.93 | 0.88 | 0.82 | 0.73 | 0.55 | 0.39 |
| 48 | **26095** | 0.41 | -0.02 | -0.01 | 0.04 | -0.02 | 0.04 | 0.02 |
| 49 | **26097** | 0.50 | 0.17 | 0.20 | 0.23 | 0.10 | 0.06 | 0.00 |
| 50 | **26099** | 0.99 | 0.89 | 0.78 | 0.68 | 0.58 | 0.46 | 0.40 |

| 51 | **26101** | 0.91 | 0.45 | 0.03 | -0.04 | -0.04 | -0.06 | -0.06 |
|---|---|---|---|---|---|---|---|---|
| 52 | **26103** | 0.83 | 0.25 | 0.17 | -0.01 | -0.02 | -0.07 | -0.01 |
| 53 | **26105** | 0.96 | 0.66 | 0.46 | 0.27 | 0.21 | 0.29 | 0.18 |
| 54 | **26107** | 0.95 | 0.71 | 0.43 | 0.21 | 0.22 | 0.34 | 0.16 |
| 55 | **26109** | 0.82 | 0.44 | 0.10 | 0.01 | 0.05 | -0.02 | -0.07 |
| 56 | **26111** | 0.97 | 0.77 | 0.54 | 0.32 | 0.12 | -0.02 | -0.07 |
| 57 | **26113** | 0.92 | 0.52 | 0.23 | 0.11 | 0.17 | 0.18 | 0.10 |
| 58 | **26115** | 0.99 | 0.95 | 0.90 | 0.85 | 0.78 | 0.62 | 0.48 |
| 59 | **26117** | 0.95 | 0.67 | 0.43 | 0.20 | 0.18 | 0.25 | 0.18 |
| 60 | **26119** | 0.85 | 0.41 | 0.32 | 0.21 | 0.03 | -0.07 | -0.10 |
| 61 | **26121** | 0.94 | 0.66 | 0.38 | 0.16 | 0.11 | 0.27 | 0.18 |
| 62 | **26123** | 0.97 | 0.76 | 0.49 | 0.27 | 0.23 | 0.27 | 0.15 |
| 63 | **26125** | 0.99 | 0.93 | 0.84 | 0.75 | 0.65 | 0.48 | 0.31 |
| 64 | **26127** | 0.93 | 0.65 | 0.43 | 0.24 | 0.16 | 0.25 | 0.15 |
| 65 | **26129** | 0.97 | 0.79 | 0.60 | 0.41 | 0.22 | -0.01 | 0.00 |
| 66 | **26131** | 0.93 | 0.67 | 0.31 | 0.00 | -0.11 | -0.15 | -0.19 |
| 67 | **26133** | 0.96 | 0.75 | 0.50 | 0.31 | 0.27 | 0.27 | 0.10 |
| 68 | **26135** | 0.88 | 0.46 | 0.37 | 0.36 | 0.33 | 0.10 | -0.02 |
| 69 | **26137** | 0.78 | 0.10 | 0.04 | 0.06 | 0.04 | -0.06 | -0.06 |
| 70 | **26139** | 0.94 | 0.69 | 0.42 | 0.21 | 0.23 | 0.33 | 0.15 |
| 71 | **26141** | 0.91 | 0.64 | 0.38 | 0.08 | -0.02 | -0.04 | -0.05 |
| 72 | **26143** | 0.88 | 0.52 | 0.28 | 0.16 | -0.02 | -0.01 | -0.02 |
| 73 | **26145** | 0.90 | 0.34 | 0.13 | 0.04 | 0.05 | -0.04 | -0.08 |
| 74 | **26147** | 0.97 | 0.81 | 0.64 | 0.49 | 0.37 | 0.17 | 0.05 |
| 75 | **26149** | 0.99 | 0.92 | 0.89 | 0.86 | 0.81 | 0.69 | 0.57 |
| 76 | **26151** | 0.92 | 0.66 | 0.53 | 0.35 | 0.23 | 0.02 | -0.05 |
| 77 | **26153** | 0.56 | -0.05 | -0.05 | -0.05 | -0.05 | -0.01 | -0.04 |
| 78 | **26155** | 0.57 | 0.18 | 0.06 | 0.05 | 0.12 | -0.05 | -0.05 |
| 79 | **26157** | 0.80 | 0.28 | 0.19 | 0.06 | -0.01 | -0.04 | -0.03 |
| 80 | **26159** | 0.96 | 0.66 | 0.27 | 0.06 | 0.01 | 0.03 | 0.01 |
| 81 | **26161** | 0.99 | 0.96 | 0.91 | 0.85 | 0.77 | 0.60 | 0.46 |
| 82 | **26163** | 1.00 | 0.97 | 0.92 | 0.86 | 0.79 | 0.62 | 0.47 |
| 83 | **26165** | 0.95 | 0.68 | 0.33 | 0.05 | 0.07 | 0.18 | 0.02 |

**Table D.2**. Autocorrelation Values (ACF) by County on the Whole Period (2022)

| No. | FIPS code | ACF lag1 | ACF lag6 | ACF lag12 | ACF lag18 | ACF lag24 | ACF lag36 | ACF lag48 |
|---|---|---|---|---|---|---|---|---|
| 1 | **26001** | 0.96 | 0.71 | 0.58 | 0.47 | 0.33 | 0.09 | -0.08 |
| 2 | **26003** | 0.31 | 0.03 | -0.08 | -0.06 | 0.03 | -0.02 | -0.05 |
| 3 | **26005** | 0.97 | 0.79 | 0.60 | 0.44 | 0.29 | 0.10 | 0.00 |
| 4 | **26007** | 0.85 | 0.51 | 0.32 | 0.14 | 0.10 | 0.09 | -0.08 |
| 5 | **26009** | 0.80 | 0.23 | 0.20 | 0.11 | 0.08 | -0.01 | -0.07 |
| 6 | **26011** | 0.73 | 0.00 | -0.02 | 0.04 | 0.15 | -0.03 | 0.06 |
| 7 | **26013** | 0.52 | 0.00 | 0.10 | -0.06 | -0.01 | -0.09 | -0.04 |
| 8 | **26015** | 0.95 | 0.79 | 0.62 | 0.44 | 0.27 | 0.02 | -0.06 |
| 9 | **26017** | 0.95 | 0.60 | 0.29 | 0.13 | 0.05 | -0.04 | -0.06 |
| 10 | **26019** | 0.51 | 0.07 | 0.00 | -0.02 | -0.04 | 0.01 | -0.02 |
| 11 | **26021** | 0.93 | 0.64 | 0.45 | 0.33 | 0.25 | 0.16 | 0.04 |
| 12 | **26023** | 0.97 | 0.81 | 0.63 | 0.44 | 0.26 | 0.07 | -0.03 |
| 13 | **26025** | 0.98 | 0.85 | 0.71 | 0.56 | 0.40 | 0.16 | 0.02 |
| 14 | **26027** | 0.92 | 0.57 | 0.34 | 0.16 | 0.13 | 0.11 | 0.05 |
| 15 | **26029** | 0.61 | 0.06 | 0.05 | -0.02 | 0.04 | -0.03 | -0.03 |
| 16 | **26031** | 0.91 | 0.35 | 0.20 | 0.24 | 0.14 | -0.03 | -0.05 |
| 17 | **26033** | 0.53 | 0.18 | 0.04 | 0.01 | 0.12 | 0.02 | -0.01 |
| 18 | **26035** | 0.82 | 0.22 | 0.15 | 0.04 | -0.01 | -0.09 | -0.07 |
| 19 | **26037** | 0.90 | 0.37 | 0.12 | 0.06 | 0.08 | -0.08 | 0.00 |
| 20 | **26039** | 0.93 | 0.50 | 0.31 | 0.24 | 0.11 | -0.03 | -0.09 |
| 21 | **26041** | 0.95 | 0.71 | 0.60 | 0.46 | 0.34 | 0.13 | -0.10 |
| 22 | **26043** | 0.68 | 0.03 | 0.02 | -0.02 | -0.02 | 0.04 | 0.15 |
| 23 | **26045** | 0.96 | 0.73 | 0.57 | 0.41 | 0.25 | 0.03 | -0.01 |
| 24 | **26047** | 0.80 | 0.25 | 0.09 | 0.03 | -0.03 | -0.03 | -0.05 |
| 25 | **26049** | 0.95 | 0.70 | 0.44 | 0.23 | 0.08 | -0.02 | -0.02 |
| 26 | **26051** | 0.79 | 0.23 | 0.18 | 0.05 | 0.04 | -0.08 | 0.15 |
| 27 | **26053** | 0.71 | 0.13 | 0.03 | -0.03 | -0.04 | -0.06 | -0.06 |
| 28 | **26055** | 0.82 | 0.13 | 0.09 | 0.05 | 0.01 | -0.02 | 0.15 |
| 29 | **26057** | 0.84 | 0.52 | 0.34 | 0.27 | 0.18 | 0.09 | -0.01 |
| 30 | **26059** | 0.97 | 0.82 | 0.68 | 0.53 | 0.39 | 0.16 | 0.03 |
| 31 | **26061** | 0.73 | 0.11 | -0.01 | 0.00 | -0.05 | -0.06 | -0.03 |
| 32 | **26063** | 0.96 | 0.80 | 0.60 | 0.40 | 0.23 | -0.01 | -0.05 |
| 33 | **26065** | 0.95 | 0.69 | 0.48 | 0.33 | 0.17 | 0.04 | -0.01 |
| 34 | **26067** | 0.91 | 0.63 | 0.43 | 0.28 | 0.15 | -0.02 | -0.03 |
| 35 | **26069** | 0.88 | 0.45 | 0.12 | 0.07 | 0.02 | -0.02 | -0.09 |
| 36 | **26071** | 0.87 | 0.44 | 0.36 | 0.23 | 0.15 | 0.03 | -0.04 |
| 37 | **26073** | 0.62 | 0.20 | 0.08 | 0.19 | 0.10 | -0.03 | -0.01 |
| 38 | **26075** | 0.98 | 0.87 | 0.72 | 0.58 | 0.43 | 0.18 | 0.01 |
| 39 | **26077** | 0.96 | 0.75 | 0.53 | 0.36 | 0.20 | 0.05 | -0.02 |
| 40 | **26079** | 0.78 | 0.08 | 0.03 | -0.02 | -0.04 | 0.01 | -0.04 |
| 41 | **26081** | 0.97 | 0.77 | 0.54 | 0.37 | 0.20 | 0.03 | -0.01 |
| 42 | **26083** | 0.78 | 0.23 | 0.06 | -0.16 | 0.00 | -0.02 | -0.08 |
| 43 | **26085** | 0.83 | 0.16 | 0.05 | 0.14 | 0.22 | -0.04 | -0.05 |
| 44 | **26087** | 0.90 | 0.51 | 0.24 | 0.05 | -0.01 | -0.04 | 0.01 |
| 45 | **26089** | 0.68 | 0.07 | 0.05 | 0.09 | 0.05 | -0.02 | 0.06 |
| 46 | **26091** | 0.98 | 0.85 | 0.71 | 0.56 | 0.40 | 0.19 | 0.02 |
| 47 | **26093** | 0.79 | 0.19 | 0.06 | 0.10 | 0.32 | 0.05 | 0.07 |
| 48 | **26095** | 0.75 | -0.09 | -0.09 | -0.15 | -0.08 | -0.06 | 0.00 |
| 49 | **26097** | 0.48 | 0.12 | 0.04 | -0.03 | 0.10 | 0.05 | -0.02 |
| 50 | **26099** | 0.98 | 0.79 | 0.44 | 0.14 | 0.02 | 0.00 | 0.00 |
| 51 | **26101** | 0.62 | 0.00 | 0.02 | 0.03 | 0.00 | -0.04 | 0.08 |
| 52 | **26103** | 0.86 | 0.29 | 0.22 | -0.01 | -0.10 | -0.04 | 0.14 |
| 53 | **26105** | 0.75 | 0.00 | -0.05 | -0.08 | 0.13 | 0.02 | -0.08 |
| 54 | **26107** | 0.79 | 0.06 | 0.04 | 0.06 | 0.01 | -0.02 | -0.04 |
| 55 | **26109** | 0.94 | 0.63 | 0.48 | 0.39 | 0.21 | 0.04 | -0.04 |
| 56 | **26111** | 0.93 | 0.56 | 0.18 | 0.11 | 0.04 | 0.00 | -0.02 |
| 57 | **26113** | 0.87 | 0.26 | 0.14 | 0.05 | 0.06 | -0.08 | -0.07 |
| 58 | **26115** | 0.83 | 0.23 | 0.07 | 0.27 | 0.31 | 0.07 | 0.02 |
| 59 | **26117** | 0.94 | 0.69 | 0.47 | 0.25 | 0.06 | -0.06 | -0.06 |
| 60 | **26119** | 0.95 | 0.69 | 0.55 | 0.39 | 0.23 | 0.05 | -0.08 |
| 61 | **26121** | 0.92 | 0.53 | 0.21 | 0.06 | 0.01 | -0.03 | -0.02 |
| 62 | **26123** | 0.77 | 0.20 | 0.04 | -0.06 | -0.08 | -0.09 | 0.00 |
| 63 | **26125** | 0.84 | 0.16 | 0.00 | 0.07 | 0.34 | 0.00 | 0.00 |
| 64 | **26127** | 0.73 | -0.03 | -0.04 | -0.07 | -0.06 | -0.05 | -0.08 |
| 65 | **26129** | 0.90 | 0.53 | 0.38 | 0.25 | 0.04 | -0.07 | -0.11 |
| 66 | **26131** | 0.66 | 0.11 | -0.04 | -0.03 | 0.03 | 0.05 | -0.02 |
| 67 | **26133** | 0.83 | 0.32 | 0.15 | 0.04 | 0.10 | -0.08 | -0.05 |
| 68 | **26135** | 0.94 | 0.68 | 0.47 | 0.30 | 0.17 | -0.03 | -0.10 |
| 69 | **26137** | 0.89 | 0.43 | 0.24 | 0.08 | 0.00 | -0.06 | -0.08 |
| 70 | **26139** | 0.93 | 0.63 | 0.35 | 0.16 | 0.10 | 0.05 | 0.05 |
| 71 | **26141** | 0.80 | 0.48 | 0.30 | 0.22 | 0.15 | -0.01 | -0.11 |
| 72 | **26143** | 0.87 | 0.15 | 0.07 | 0.06 | 0.23 | 0.13 | 0.03 |
| 73 | **26145** | 0.67 | 0.09 | 0.03 | 0.00 | 0.01 | -0.02 | -0.02 |
| 74 | **26147** | 0.90 | 0.57 | 0.26 | 0.09 | 0.00 | -0.01 | 0.00 |
| 75 | **26149** | 0.97 | 0.81 | 0.60 | 0.46 | 0.34 | 0.13 | -0.01 |
| 76 | **26151** | 0.91 | 0.53 | 0.20 | 0.02 | -0.03 | -0.02 | -0.03 |
| 77 | **26153** | 0.42 | 0.09 | -0.07 | -0.07 | 0.00 | -0.01 | -0.02 |
| 78 | **26155** | 0.81 | 0.26 | 0.06 | 0.06 | 0.02 | -0.07 | -0.03 |
| 79 | **26157** | 0.95 | 0.70 | 0.33 | 0.14 | 0.08 | 0.11 | 0.05 |
| 80 | **26159** | 0.96 | 0.71 | 0.54 | 0.36 | 0.23 | 0.04 | -0.05 |
| 81 | **26161** | 0.85 | 0.27 | 0.15 | 0.20 | 0.36 | 0.05 | 0.02 |
| 82 | **26163** | 0.98 | 0.80 | 0.48 | 0.18 | 0.03 | 0.00 | 0.00 |
| 83 | **26165** | 0.91 | 0.52 | 0.35 | 0.22 | 0.11 | -0.06 | 0.09 |

**Table D.3**. Autocorrelation Values (ACF) by County during the Large-scale Outages (2021)

| **No.** | **Fips code** | **ACF lag1** | **ACF lag6** | **ACF lag12** | **ACF lag18** | **ACF lag24** | **ACF lag36** | **ACF lag48** |
|---|---|---|---|---|---|---|---|---|
| **1** | 26001 | 0,53 | 0,22 | -0,12 | -0,31 | -0,15 | 0,06 | |
| **2** | 26003 | 0,29 | | | | | | |
| **3** | 26005 | 0,66 | 0,10 | 0,02 | 0,04 | 0,12 | 0,02 | -0,07 |
| **4** | 26007 | 0,61 | 0,06 | -0,01 | -0,13 | | | |
| **5** | 26009 | 0,41 | 0,30 | 0,08 | 0,15 | -0,06 | -0,04 | -0,07 |
| **6** | 26011 | 0,51 | 0,03 | -0,19 | -0,17 | 0,10 | -0,01 | -0,09 |
| **7** | 26013 | 0,37 | -0,20 | | | | | |
| **8** | 26015 | 0,59 | -0,01 | 0,01 | -0,03 | 0,01 | 0,00 | -0,05 |
| **9** | 26017 | 0,80 | -0,01 | -0,03 | -0,04 | -0,03 | 0,00 | -0,02 |
| **10** | 26019 | 0,07 | -0,07 | -0,01 | 0,01 | -0,09 | | |
| **11** | 26021 | 0,92 | 0,44 | 0,02 | -0,10 | -0,08 | 0,06 | -0,07 |
| **12** | 26023 | 0,92 | 0,59 | 0,25 | 0,03 | -0,21 | -0,14 | -0,19 |
| **13** | 26025 | 0,88 | 0,53 | 0,26 | 0,08 | -0,11 | -0,11 | -0,05 |
| **14** | 26027 | 0,91 | 0,51 | 0,08 | 0,00 | 0,03 | 0,00 | -0,06 |
| **15** | 26029 | 0,44 | 0,15 | 0,10 | -0,15 | -0,10 | 0,01 | |
| **16** | 26031 | 0,45 | -0,01 | -0,03 | -0,04 | 0,03 | -0,04 | -0,02 |
| **17** | 26033 | 0,45 | -0,06 | 0,01 | -0,06 | -0,10 | -0,13 | -0,08 |
| **18** | 26035 | 0,83 | -0,01 | -0,02 | -0,02 | -0,03 | -0,05 | -0,06 |
| **19** | 26037 | 0,20 | 0,00 | -0,06 | -0,04 | -0,02 | -0,02 | 0,23 |
| **20** | 26039 | -0,12 | 0,07 | -0,08 | | | | |
| **21** | 26041 | 0,32 | 0,13 | -0,03 | 0,02 | 0,01 | | |
| **22** | 26043 | 0,37 | -0,13 | -0,15 | -0,04 | | | |
| **23** | 26045 | 0,75 | -0,01 | 0,07 | -0,03 | -0,15 | -0,16 | 0,08 |
| **24** | 26047 | 0,54 | -0,09 | -0,08 | 0,03 | | | |
| **25** | 26049 | 0,79 | 0,06 | -0,02 | -0,14 | -0,17 | -0,10 | 0,01 |
| **26** | 26051 | 0,87 | -0,02 | -0,02 | 0,01 | -0,02 | -0,02 | -0,05 |
| **27** | 26053 | 0,41 | -0,18 | 0,04 | 0,01 | 0,02 | | |
| **28** | 26055 | 0,47 | -0,06 | -0,06 | -0,03 | -0,05 | 0,03 | -0,03 |
| **29** | 26057 | 0,83 | 0,00 | -0,01 | -0,04 | -0,05 | -0,04 | -0,08 |
| **30** | 26059 | 0,93 | 0,58 | 0,23 | 0,01 | -0,20 | -0,17 | -0,13 |
| **31** | 26061 | 0,11 | 0,20 | 0,00 | -0,14 | -0,02 | 0,00 | |
| **32** | 26063 | 0,83 | 0,03 | -0,07 | -0,02 | -0,02 | -0,01 | -0,03 |
| **33** | 26065 | 0,88 | 0,38 | 0,01 | -0,06 | -0,11 | -0,11 | -0,03 |
| **34** | 26067 | 0,83 | -0,08 | -0,07 | -0,03 | -0,15 | -0,10 | 0,27 |
| **35** | 26069 | 0,81 | 0,00 | 0,00 | -0,02 | -0,03 | 0,00 | -0,06 |
| **36** | 26071 | 0,77 | -0,07 | -0,14 | 0,00 | | | |
| **37** | 26073 | 0,85 | -0,02 | -0,03 | -0,02 | 0,00 | 0,01 | -0,05 |
| **38** | 26075 | 0,94 | 0,65 | 0,36 | 0,11 | -0,15 | -0,11 | -0,06 |
| **39** | 26077 | 0,83 | 0,10 | -0,01 | -0,09 | -0,02 | -0,01 | 0,03 |
| **40** | 26079 | 0,76 | -0,08 | -0,05 | 0,02 | -0,05 | -0,11 | |
| **41** | 26081 | 0,83 | 0,13 | -0,03 | -0,02 | -0,03 | -0,01 | -0,02 |
| **42** | 26083 | 0,67 | -0,29 | 0,03 | | | | |
| **43** | 26085 | 0,58 | 0,11 | 0,04 | -0,11 | -0,13 | -0,21 | -0,02 |

| **44** | 26087 | 0,71 | 0,06 | 0,01 | -0,01 | -0,08 | -0,07 | -0,07 |
|---|---|---|---|---|---|---|---|---|
| **45** | 26089 | 0,79 | -0,09 | -0,24 | 0,19 | -0,10 | 0,05 | |
| **46** | 26091 | 0,91 | 0,53 | 0,11 | -0,05 | -0,08 | -0,10 | -0,04 |
| **47** | 26093 | 0,86 | 0,42 | 0,47 | 0,27 | -0,01 | -0,10 | -0,11 |
| **48** | 26095 | 0,80 | -0,21 | -0,20 | | | | |
| **49** | 26097 | 0,46 | 0,06 | -0,03 | -0,04 | -0,01 | -0,04 | 0,01 |
| **50** | 26099 | 0,88 | 0,29 | 0,13 | 0,15 | -0,12 | -0,06 | -0,03 |
| **51** | 26101 | 0,86 | 0,39 | -0,18 | -0,21 | -0,12 | -0,08 | -0,10 |
| **52** | 26103 | -0,04 | -0,05 | -0,06 | 0,00 | | | |
| **53** | 26105 | 0,61 | 0,01 | -0,04 | -0,05 | -0,04 | -0,07 | -0,08 |
| **54** | 26107 | 0,82 | -0,01 | -0,08 | -0,08 | -0,02 | -0,05 | 0,05 |
| **55** | 26109 | 0,17 | -0,27 | | | | | |
| **56** | 26111 | 0,85 | 0,04 | -0,02 | -0,07 | -0,09 | 0,10 | 0,14 |
| **57** | 26113 | 0,50 | -0,02 | -0,04 | -0,07 | -0,09 | -0,07 | |
| **58** | 26115 | 0,94 | 0,55 | 0,21 | 0,02 | -0,19 | -0,15 | -0,09 |
| **59** | 26117 | 0,84 | 0,05 | -0,01 | -0,05 | -0,06 | -0,05 | -0,03 |
| **60** | 26119 | 0,67 | -0,02 | -0,03 | -0,05 | -0,06 | | |
| **61** | 26121 | 0,54 | 0,03 | -0,04 | -0,06 | -0,06 | -0,03 | -0,04 |
| **62** | 26123 | 0,85 | 0,09 | 0,01 | 0,04 | 0,01 | -0,04 | -0,05 |
| **63** | 26125 | 0,95 | 0,61 | 0,42 | 0,24 | 0,15 | -0,17 | -0,26 |
| **64** | 26127 | 0,67 | -0,04 | -0,05 | -0,05 | -0,03 | -0,05 | 0,29 |
| **65** | 26129 | 0,03 | -0,04 | -0,03 | -0,03 | -0,04 | 0,00 | -0,02 |
| **66** | 26133 | 0,88 | 0,12 | 0,00 | -0,05 | -0,08 | -0,11 | -0,11 |
| **67** | 26135 | 0,45 | 0,06 | -0,06 | -0,10 | -0,01 | 0,01 | |
| **68** | 26137 | 0,19 | -0,07 | -0,01 | 0,04 | 0,14 | -0,01 | |
| **69** | 26139 | 0,67 | 0,06 | -0,12 | -0,04 | -0,08 | -0,07 | -0,08 |
| **70** | 26141 | 0,43 | -0,13 | -0,07 | -0,16 | | | |
| **71** | 26143 | 0,38 | -0,07 | -0,01 | 0,00 | -0,02 | 0,01 | 0,00 |
| **72** | 26145 | 0,84 | 0,04 | -0,09 | -0,10 | -0,08 | -0,08 | 0,15 |
| **73** | 26147 | 0,78 | 0,33 | 0,17 | 0,04 | -0,12 | -0,15 | -0,16 |
| **74** | 26149 | 0,93 | 0,64 | 0,33 | 0,06 | -0,24 | -0,20 | -0,07 |
| **75** | 26151 | 0,48 | 0,12 | -0,06 | -0,07 | -0,06 | -0,07 | -0,03 |
| **76** | 26153 | 0,52 | -0,20 | | | | | |
| **77** | 26155 | 0,73 | 0,07 | -0,29 | -0,12 | 0,40 | -0,22 | 0,22 |
| **78** | 26157 | 0,83 | 0,11 | 0,09 | -0,04 | -0,03 | -0,04 | -0,01 |
| **79** | 26159 | 0,94 | 0,51 | 0,05 | -0,04 | 0,02 | -0,02 | -0,07 |
| **80** | 26161 | 0,93 | 0,55 | 0,26 | 0,13 | -0,13 | -0,14 | -0,08 |
| **81** | 26163 | 0,94 | 0,62 | 0,38 | 0,15 | -0,22 | -0,28 | -0,18 |
| **82** | 26165 | 0,88 | 0,35 | -0,07 | -0,02 | -0,12 | -0,15 | -0,11 |

**Table D.4**. Autocorrelation Values (ACF) by County during the Large-scale Outages (2022)

| **No.** | **FIPS code** | **ACF lag1** | **ACF lag6** | **ACF lag12** | **ACF lag18** | **ACF lag24** | **ACF lag36** | **ACF lag48** |
|---|---|---|---|---|---|---|---|---|
| 1 | **26001** | 0.77 | -0.09 | -0.15 | -0.05 | 0.05 | -0.05 | 0.00 |
| 2 | **26003** | 0.02 | -0.05 | 0.01 | | | | |
| 3 | **26005** | 0.91 | 0.29 | -0.14 | 0.08 | 0.25 | -0.09 | -0.12 |
| 4 | **26007** | 0.69 | -0.12 | -0.18 | 0.00 | 0.02 | | |
| 5 | **26009** | 0.76 | -0.30 | 0.25 | -0.23 | -0.13 | -0.08 | 0.03 |
| 6 | **26011** | 0.74 | -0.27 | -0.14 | -0.01 | 0.01 | 0.01 | |
| 7 | **26013** | 0.64 | -0.36 | 0.17 | 0.02 | | | |
| 8 | **26015** | 0.81 | 0.01 | -0.02 | 0.24 | 0.10 | -0.10 | -0.18 |
| 9 | **26017** | 0.93 | 0.55 | 0.09 | -0.09 | -0.08 | -0.18 | -0.20 |
| 10 | **26019** | 0.33 | -0.04 | -0.04 | -0.05 | -0.09 | | |
| 11 | **26021** | 0.81 | 0.02 | -0.09 | -0.02 | -0.19 | 0.25 | -0.11 |
| 12 | **26023** | 0.92 | 0.38 | -0.02 | -0.02 | -0.09 | -0.11 | -0.17 |
| 13 | **26025** | 0.91 | 0.35 | -0.08 | 0.13 | 0.15 | -0.07 | -0.11 |
| 14 | **26027** | 0.82 | -0.04 | -0.15 | 0.00 | -0.17 | -0.13 | 0.29 |
| 15 | **26029** | 0.80 | -0.11 | -0.10 | -0.12 | -0.09 | 0.03 | 0.01 |
| 16 | **26031** | 0.89 | -0.10 | -0.22 | -0.10 | -0.10 | 0.03 | |
| 17 | **26033** | 0.82 | 0.26 | 0.02 | -0.09 | -0.12 | -0.08 | -0.11 |
| 18 | **26035** | 0.79 | 0.00 | 0.16 | -0.13 | -0.18 | 0.01 | 0.04 |
| 19 | **26037** | 0.82 | -0.08 | -0.20 | 0.11 | -0.07 | 0.18 | -0.19 |
| 20 | **26039** | 0.83 | -0.19 | -0.26 | -0.07 | 0.03 | 0.01 | |
| 21 | **26041** | 0.79 | 0.21 | 0.11 | -0.29 | -0.24 | -0.07 | -0.04 |
| 22 | **26043** | 0.58 | 0.10 | -0.08 | -0.13 | -0.01 | 0.02 | |
| 23 | **26045** | 0.92 | 0.39 | -0.03 | -0.02 | -0.10 | -0.11 | -0.22 |
| 24 | **26047** | 0.84 | -0.17 | -0.19 | -0.19 | 0.06 | 0.02 | |
| 25 | **26049** | 0.88 | 0.13 | -0.15 | -0.11 | 0.43 | -0.09 | -0.03 |
| 26 | **26051** | 0.69 | -0.08 | -0.04 | -0.04 | 0.00 | -0.12 | 0.13 |
| 27 | **26053** | 0.84 | 0.11 | -0.46 | -0.25 | 0.10 | | |
| 28 | **26055** | 0.80 | -0.23 | 0.18 | -0.10 | -0.22 | -0.02 | 0.04 |
| 29 | **26057** | 0.85 | 0.19 | -0.15 | -0.19 | -0.17 | -0.07 | 0.17 |
| 30 | **26059** | 0.89 | 0.42 | 0.03 | -0.07 | -0.12 | -0.11 | -0.19 |
| 31 | **26061** | 0.55 | 0.05 | -0.11 | -0.01 | | | |
| 32 | **26063** | 0.87 | 0.44 | -0.06 | -0.40 | -0.29 | | |
| 33 | **26065** | 0.90 | 0.31 | -0.08 | -0.11 | -0.15 | -0.17 | 0.29 |
| 34 | **26067** | 0.89 | 0.39 | -0.11 | -0.12 | -0.11 | 0.07 | -0.17 |
| 35 | **26069** | 0.79 | 0.33 | -0.02 | -0.12 | -0.05 | 0.01 | -0.21 |
| 36 | **26071** | 0.76 | 0.38 | 0.15 | 0.08 | -0.32 | -0.14 | -0.12 |
| 37 | **26073** | 0.77 | 0.19 | 0.01 | -0.05 | -0.04 | -0.10 | -0.09 |
| 38 | **26075** | 0.89 | 0.29 | -0.07 | 0.19 | 0.16 | -0.10 | -0.14 |
| 39 | **26077** | 0.88 | 0.20 | 0.09 | 0.33 | 0.00 | 0.00 | -0.14 |
| 40 | **26079** | 0.82 | -0.09 | -0.13 | -0.14 | -0.04 | -0.06 | 0.02 |
| 41 | **26081** | 0.93 | 0.36 | -0.19 | -0.18 | -0.21 | -0.23 | 0.09 |
| 42 | **26083** | 0.75 | -0.41 | | | | | |
| 43 | **26085** | 0.82 | -0.14 | -0.11 | -0.15 | -0.06 | -0.04 | 0.01 |
| 44 | **26087** | 0.72 | 0.16 | -0.19 | 0.07 | -0.05 | -0.14 | -0.01 |
| 45 | **26089** | 0.55 | 0.08 | -0.08 | -0.07 | -0.07 | 0.01 | |
| 46 | **26091** | 0.92 | 0.37 | -0.02 | 0.02 | -0.02 | -0.07 | -0.09 |
| 47 | **26093** | 0.16 | 0.05 | -0.03 | 0.00 | 0.02 | -0.01 | -0.04 |
| 48 | **26095** | 0.52 | -0.31 | -0.09 | -0.06 | | | |
| 49 | **26097** | 0.81 | -0.34 | -0.15 | -0.04 | 0.05 | | |
| 50 | **26099** | 0.86 | 0.12 | -0.02 | -0.02 | -0.02 | -0.03 | -0.05 |
| 51 | **26101** | 0.52 | -0.18 | -0.15 | 0.04 | -0.05 | 0.16 | -0.05 |
| 52 | **26103** | 0.79 | 0.36 | 0.01 | -0.26 | -0.39 | -0.14 | 0.15 |
| 53 | **26105** | 0.76 | 0.02 | -0.10 | -0.14 | -0.20 | 0.39 | |
| 54 | **26107** | 0.84 | -0.06 | -0.06 | -0.04 | -0.10 | 0.03 | 0.04 |
| 55 | **26109** | 0.80 | -0.05 | -0.26 | -0.05 | -0.20 | 0.07 | |
| 56 | **26111** | 0.92 | 0.39 | -0.07 | -0.09 | -0.08 | -0.13 | -0.16 |
| 57 | **26113** | 0.79 | 0.04 | 0.06 | -0.33 | -0.16 | 0.08 | 0.03 |
| 58 | **26115** | 0.85 | 0.51 | 0.38 | 0.18 | -0.01 | -0.19 | -0.28 |
| 59 | **26117** | 0.92 | 0.36 | -0.21 | -0.22 | -0.22 | -0.14 | 0.09 |
| 60 | **26119** | 0.75 | -0.11 | -0.11 | -0.15 | -0.06 | 0.03 | 0.00 |
| 61 | **26121** | 0.93 | 0.39 | -0.12 | -0.16 | -0.19 | -0.20 | -0.06 |
| 62 | **26123** | 0.77 | 0.09 | 0.01 | -0.11 | -0.01 | -0.11 | -0.11 |
| 63 | **26125** | 0.72 | -0.04 | -0.07 | -0.08 | -0.06 | 0.40 | 0.00 |
| 64 | **26127** | 0.71 | 0.40 | -0.17 | -0.37 | -0.29 | -0.01 | 0.06 |
| 65 | **26129** | 0.83 | 0.23 | -0.11 | 0.01 | -0.14 | -0.18 | -0.02 |
| 66 | **26131** | 0.54 | -0.14 | 0.02 | | | | |
| 67 | **26133** | 0.79 | -0.10 | -0.08 | 0.13 | -0.06 | -0.08 | 0.01 |
| 68 | **26135** | 0.77 | -0.12 | -0.05 | -0.02 | -0.06 | -0.03 | |
| 69 | **26137** | 0.79 | -0.32 | 0.02 | -0.01 | -0.18 | 0.05 | |
| 70 | **26139** | 0.90 | 0.32 | -0.21 | -0.26 | -0.27 | -0.01 | 0.13 |
| 71 | **26141** | 0.58 | -0.14 | 0.23 | -0.07 | -0.11 | | |
| 72 | **26143** | 0.48 | -0.07 | -0.12 | -0.07 | 0.16 | -0.08 | -0.01 |
| 73 | **26145** | 0.83 | -0.12 | -0.09 | -0.04 | 0.04 | 0.08 | -0.04 |
| 74 | **26147** | 0.86 | 0.12 | -0.01 | -0.04 | -0.05 | -0.07 | -0.08 |
| 75 | **26149** | 0.70 | -0.04 | -0.23 | 0.11 | 0.11 | -0.16 | 0.01 |
| 76 | **26151** | 0.95 | 0.55 | 0.01 | -0.15 | -0.21 | -0.25 | -0.03 |
| 77 | **26153** | 0.77 | -0.07 | -0.10 | -0.15 | 0.15 | -0.03 | |
| 78 | **26155** | 0.81 | -0.04 | -0.06 | -0.13 | 0.25 | -0.03 | -0.10 |
| 79 | **26157** | 0.93 | 0.48 | -0.13 | -0.25 | -0.15 | -0.16 | -0.05 |
| 80 | **26159** | 0.86 | -0.10 | -0.31 | 0.19 | 0.20 | -0.21 | 0.07 |
| 81 | **26161** | 0.63 | 0.18 | -0.07 | -0.03 | 0.03 | -0.10 | -0.06 |
| 82 | **26163** | 0.86 | 0.13 | -0.02 | -0.03 | -0.03 | -0.04 | -0.06 |
| 83 | **26165** | 0.80 | -0.28 | 0.12 | -0.06 | -0.17 | -0.09 | 0.01 |

**Table D.5.** Results of Moran's I Analysis on major event windows (≥ 50,000), 2021 - 2022

| Event No | Start time | End time | Number of counties | Moran's I Index | p-value (permutation test) |
|---|---|---|---|---|---|
| 1 | 2021-06-21 17:00:00 | 2021-06-21 19:00:00 | 65 | 0.29963 | 0.005 |
| 2 | 2021-06-27 00:00:00 | 2021-06-27 02:00:00 | 56 | 0.04028 | 0.049 |
| 3 | 2021-06-29 19:00:00 | 2021-06-29 21:00:00 | 49 | 0.208727 | 0.007 |
| 4 | 2021-07-07 22:00:00 | 2021-07-08 00:00:00 | 47 | 0.291255 | 0.01 |
| 5 | 2021-07-25 15:00:00 | 2021-07-25 17:00:00 | 46 | 0.129834 | 0.015 |
| 6 | 2021-08-16 15:00:00 | 2021-08-16 15:00:00 | 59 | 0.278866 | 0.002 |
| 7 | 2021-08-25 01:00:00 | 2021-08-25 03:00:00 | 57 | 0.355745 | 0.001 |
| 8 | 2022-06-16 22:00:00 | 2022-06-17 00:00:00 | 79 | 0.324211 | 0.001 |
| 9 | 2022-07-24 11:00:00 | 2022-07-24 13:00:00 | 64 | 0.2031506 | 0.003 |
| 10 | 2022-08-04 00:00:00 | 2022-08-04 02:00:00 | 58 | 0.230081 | 0.013 |
| 11 | 2022-08-30 04:00:00 | 2022-08-30 06:00:00 | 62 | 0.135088 | 0.018 |

# Appendix E. Feature Engineering

**Table E.1.** Final list of engineered features by group

| No. | Feature (code) | Group | Window | Units |
|---|---|---|---|---|
| 1 | dwpf | raw | - | °F |
| 2 | tmpf | raw | - | °F |
| 3 | alti | raw | - | inHg |
| 4 | mslp | raw | - | mb (hPa) |
| 5 | gust | raw | - | kt |
| 6 | p01i | raw | - | in |
| 7 | sknt | raw | - | kt |
| 8 | drct_u | raw | - | kt |
| 9 | drct_v | raw | - | kt |
| 10 | relh | raw | - | % |
| 11 | relh_grad | gradient | - | %/km |
| 12 | sq_flag | raw | - | - |
| 13 | ts_flag | raw | - | - |
| 14 | hr_flag | raw | - | - |
| 15 | dwpf lag 6h | lag | 6 h | °F |
| 16 | dwpf_lag_12h | lag | 12 h | °F |
| 17 | dwpf_lag_24h | lag | 24 h | °F |
| 18 | dwpf_lag_48h | lag | 48 h | °F |
| 19 | tmpf_lag_6h | lag | 6 h | °F |
| 20 | tmpf_lag_12h | lag | 12 h | °F |
| 21 | tmpf_lag_24h | lag | 24 h | °F |
| 22 | tmpf_lag_48h | lag | 48 h | °F |
| 23 | drct_u_lag_6h | lag | 6 h | kt |
| 24 | drct_u_lag_12h | lag | 12 h | kt |
| 25 | drct_u_lag_24h | lag | 24 h | kt |
| 26 | drct_u_lag_48h | lag | 48 h | kt |
| 27 | drct_v_lag_6h | lag | 6 h | kt |
| 28 | drct_v_lag_12h | lag | 12 h | kt |
| 29 | drct_v_lag_24h | lag | 24 h | kt |

| 30 | drct_v_lag_48h | lag | 48 h | kt |
| --- | --- | --- | --- | --- |
| 31 | p01i_rolling_sum_6h | rolling | 6 h | in |
| 32 | p01i_rolling_sum_12h | rolling | 12 h | in |
| 33 | p01i_rolling_sum_24h | rolling | 24 h | in |
| 34 | p01i_rolling_sum_48h | rolling | 48 h | in |
| 35 | alti_rolling_mean_6h | rolling | 6 h | inHg |
| 36 | alti_rolling_mean_12h | rolling | 12 h | inHg |
| 37 | alti_rolling_mean_24h | rolling | 24 h | inHg |
| 38 | alti_rolling_mean_48h | rolling | 48 h | inHg |
| 39 | mslp_rolling_mean_6h | rolling | 6 h | mb (hPa) |
| 40 | mslp_rolling_mean_12h | rolling | 12 h | mb (hPa) |
| 41 | mslp_rolling_mean_24h | rolling | 24 h | mb (hPa) |
| 42 | mslp_rolling_mean_48h | rolling | 48 h | mb (hPa) |
| 43 | relh rolling mean 6h | rolling | 6 h | % |
| 44 | relh_rolling_mean_12h | rolling | 12 h | % |
| 45 | relh_rolling_mean_24h | rolling | 24 h | % |
| 46 | relh_rolling_mean_48h | rolling | 48 h | % |
| 47 | gust_rolling_max_6h | rolling | 6 h | kt |
| 48 | gust rolling max 12h | rolling | 12 h | kt |
| 49 | gust_rolling_max_24h | rolling | 24 h | kt |
| 50 | gust_rolling_max_48h | rolling | 48 h | kt |
| 51 | sknt_rolling_max_6h | rolling | 6 h | kt |
| 52 | sknt_rolling_max_12h | rolling | 12 h | kt |
| 53 | sknt_rolling_max_24h | rolling | 24 h | kt |
| 54 | sknt_rolling_max_48h | rolling | 48 h | kt |
| 55 | relh_grad_rolling_max_6h | rolling | 6 h | %/km |
| 56 | relh_grad_rolling_max_12h | rolling | 12 h | %/km |
| 57 | relh_grad_rolling_max_24h | rolling | 24 h | %/km |
| 58 | relh_grad_rolling_max_48h | rolling | 48 h | %/km |
| 59 | ts_flag_rolling_sum_6h | rolling | 6 h | - |
| 60 | ts_flag_rolling_sum_12h | rolling | 12 h | - |
| 61 | ts_flag_rolling_sum_24h | rolling | 24 h | - |
| 62 | ts_flag_rolling_sum_48h | rolling | 48 h | - |

| 63 | hr_flag_rolling_sum_6h | rolling | 6 h | - |
|---|---|---|---|---|
| 64 | hr_flag_rolling_sum_12h | rolling | 12 h | - |
| 65 | hr_flag_rolling_sum_24h | rolling | 24 h | - |
| 66 | hr_flag_rolling_sum_48h | rolling | 48 h | - |
| 67 | sq_flag_rolling_sum_6h | rolling | 6 h | - |
| 68 | sq_flag_rolling_sum_12h | rolling | 12 h | - |
| 69 | sq_flag_rolling_sum_24h | rolling | 24 h | - |
| 70 | sq_flag_rolling_sum_48h | rolling | 48 h | - |
| 71 | IDW_alti | IDW | - | inHg |
| 72 | IDW_dwpf | IDW | - | °F |
| 73 | IDW_drct_u | IDW | - | kt |
| 74 | IDW_drct_v | IDW | - | kt |
| 75 | IDW_tmpf_lag_6h | IDW | 6 h | °F |
| 76 | IDW drct v lag 6h | IDW | 6 h | kt |
| 77 | IDW_drct_u_lag_12h | IDW | 12 h | kt |
| 78 | IDW_dwpf_lag_12h | IDW | 12 h | °F |
| 79 | IDW_relh_rolling_mean_48h | IDW | 48 h | % |
| 80 | IDW_gust_rolling_max_24h | IDW | 24 h | kt |
| 81 | IDW sknt rolling max 48h | IDW | 48 h | kt |
| 82 | IDW_ts_flag_rolling_sum_12h | IDW | 12 h | - |
| 83 | IDW_p0li_rolling_sum_24h | IDW | 24 h | in |
| 84 | day_of_week_num | calendar | - | - |
| 85 | population_density | static | - | people/$km^2$ |
| 86 | x (lon) | static | - | deg |
| 87 | y (lat) | static | - | deg |

**Table E.2.** Pearson correlation analysis with targets on the train split.

| Corr (flag, t+48) | | | Corr (log1p(sum_48) | | |
|---|---|---|---|---|---|
| **Rank** | **Feature** | **Corr** | **Rank** | **Feature** | **Corr** |
| 1 | IDW_drct_v_lag_6h | 0.171 | 1 | population_density | 0.274 |
| 2 | IDW_dwpf_lag_12h | 0.164 | 2 | sq_flag_rolling_sum_48h | 0.231 |
| 3 | dwpf_lag_6h | 0.163 | 3 | sq_flag_rolling_sum_24h | 0.193 |
| 4 | dwpf_lag_12h | 0.160 | 4 | ts_flag_rolling_sum_48h | 0.168 |
| 5 | IDW_drct_v | 0.160 | 5 | sknt_rolling_max_48h | 0.162 |
| 6 | dwpf | 0.158 | 6 | p0li_rolling_sum_48h | 0.161 |
| 7 | drct_v_lag_12h | 0.157 | 7 | gust_rolling_max_48h | 0.149 |
| 8 | drct_v_lag_24h | 0.156 | 8 | ts_flag_rolling_sum_24h | 0.148 |
| 9 | IDW_gust_rolling_max_24h | 0.154 | 9 | sknt_rolling_max_24h | 0.144 |
| 10 | IDW_dwpf | 0.153 | 10 | sq_flag_rolling_sum_12h | 0.141 |
| 11 | drct_v | 0.150 | 11 | p0li_rolling_sum_24h | 0.139 |
| 12 | drct_v_lag_6h | 0.148 | 12 | gust_rolling_max_24h | 0.134 |
| 13 | sq_flag_rolling_sum_48h | 0.142 | 13 | IDW_gust_rolling_max_24h | 0.130 |
| 14 | IDW_sknt_rolling_max_48h | 0.133 | 14 | sknt_rolling_max_12h | 0.125 |
| 15 | sknt_rolling_max_24h | 0.131 | 15 | gust_rolling_max_12h | 0.123 |
| 16 | sknt_rolling_max_12h | 0.130 | 16 | ts_flag_rolling_sum_12h | 0.122 |
| 17 | dwpf_lag_24h | 0.127 | 17 | hr_flag_rolling_sum_48h | 0.118 |
| 18 | sknt_rolling_max_48h | 0.122 | 18 | gust_rolling_max_6h | 0.117 |
| 19 | IDW_ts_flag_rolling_sum_12h | 0.118 | 19 | x | 0.113 |
| 20 | gust_rolling_max_24h | 0.118 | 20 | p0li_rolling_sum_12h | 0.110 |
| 21 | gust_rolling_max_48h | 0.118 | 21 | sknt_rolling_max_6h | 0.106 |
| 22 | sknt_rolling_max_6h | 0.116 | 22 | drct_v_lag_24h | 0.106 |
| 23 | ts_flag_rolling_sum_48h | 0.113 | 23 | county_encoded | 0.103 |
| 24 | gust_rolling_max_12h | 0.112 | 24 | hr_flag_rolling_sum_24h | 0.102 |
| 25 | sq_flag_rolling_sum_24h | 0.110 | 25 | IDW_drct_v_lag_6h | 0.102 |
| 26 | IDW_relh_rolling_mean_48h | 0.102 | 26 | sq_flag_rolling_sum_6h | 0.101 |
| 27 | gust_rolling_max_6h | 0.101 | 27 | dwpf_lag_6h | 0.100 |
| 28 | IDW_tmpf_lag_6h | 0.099 | 28 | dwpf_lag_12h | 0.100 |
| 29 | relh_rolling_mean_24h | 0.098 | 29 | dwpf | 0.099 |
| 30 | ts_flag_rolling_sum_24h | 0.097 | 30 | gust | 0.099 |
| 31 | sknt | 0.091 | 31 | IDW_sknt_rolling_max_48h | 0.098 |
| 32 | p0li_rolling_sum_48h | 0.088 | 32 | drct_v_lag_12h | 0.097 |
| 33 | relh_rolling_mean_48h | 0.088 | 33 | ts_flag_rolling_sum_6h | 0.097 |
| 34 | tmpf_lag_12h | 0.087 | 34 | IDW_dwpf_lag_12h | 0.096 |
| 35 | tmpf_lag_6h | 0.087 | 35 | dwpf_lag_24h | 0.095 |
| 36 | IDW_p0li_rolling_sum_24h | 0.086 | 36 | drct_v_lag_6h | 0.091 |
| 37 | drct_v_lag_48h | 0.085 | 37 | IDW_dwpf | 0.090 |
| 38 | relh_rolling_mean_12h | 0.083 | 38 | IDW_ts_flag_rolling_sum_12h | 0.089 |
| 39 | hr_flag_rolling_sum_48h | 0.081 | 39 | IDW_drct_v | 0.086 |
| 40 | ts_flag_rolling_sum_12h | 0.080 | 40 | hr_flag_rolling_sum_12h | 0.085 |
| 41 | sq_flag_rolling_sum_12h | 0.078 | 41 | drct_v | 0.085 |
| 42 | tmpf_lag_24h | 0.078 | 42 | p0li_rolling_sum_6h | 0.084 |
| 43 | tmpf | 0.075 | 43 | IDW_p0li_rolling_sum_24h | 0.081 |
| 44 | p0li_rolling_sum_24h | 0.074 | 44 | drct_v_lag_48h | 0.080 |
| 45 | relh_rolling_mean_6h | 0.073 | 45 | sknt | 0.079 |
| 46 | hr_flag_rolling_sum_24h | 0.070 | 46 | dwpf_lag_48h | 0.078 |
| 47 | ts_flag_rolling_sum_6h | 0.068 | 47 | hr_flag_rolling_sum_6h | 0.068 |
| 48 | dwpf_lag_48h | 0.065 | 48 | tmpf | 0.064 |
| 49 | population_density | 0.065 | 49 | tmpf_lag_6h | 0.062 |
| 50 | p0li_rolling_sum_12h | 0.063 | 50 | drct_u | 0.061 |
| 51 | hr_flag_rolling_sum_12h | 0.063 | 51 | tmpf_lag_24h | 0.061 |
| 52 | relh | 0.063 | 52 | tmpf_lag_12h | 0.061 |
| 53 | hr_flag_rolling_sum_6h | 0.057 | 53 | IDW_tmpf_lag_6h | 0.061 |
| 54 | sq_flag_rolling_sum_6h | 0.056 | 54 | ts_flag | 0.060 |
| 55 | gust | 0.056 | 55 | drct_u_lag_6h | 0.055 |
| 56 | p0li_rolling_sum_6h | 0.054 | 56 | relh_grad_rolling_max_48h | 0.053 |
| 57 | x | 0.052 | 57 | IDW_relh_rolling_mean_48h | 0.053 |
| 58 | IDW_drct_u_lag_12h | 0.048 | 58 | relh_grad_rolling_max_24h | 0.049 |
| 59 | IDW_drct_u | 0.046 | 59 | IDW_drct_u | 0.046 |
| 60 | drct_u_lag_6h | 0.045 | 60 | sq_flag | 0.046 |
| 61 | ts_flag | 0.042 | 61 | tmpf_lag_48h | 0.045 |
| 62 | drct_u | 0.040 | 62 | p0li | 0.044 |
| 63 | county_encoded | 0.039 | 63 | drct_u_lag_12h | 0.044 |
| 64 | drct_u_lag_12h | 0.039 | 64 | relh_grad_rolling_max_12h | 0.042 |
| 65 | relh_grad | 0.036 | 65 | relh_grad_rolling_max_6h | 0.039 |
| 66 | relh_grad_rolling_max_6h | 0.035 | 66 | hr_flag | 0.037 |
| 67 | drct_u_lag_24h | 0.032 | 67 | IDW_drct_u_lag_12h | 0.033 |
| 68 | hr_flag | 0.032 | 68 | drct_u_lag_24h | 0.033 |
| 69 | p01i | 0.030 | 69 | relh_rolling_mean_24h | 0.031 |
| 70 | relh_grad_rolling_max_12h | 0.029 | 70 | relh_rolling_mean_48h | 0.030 |
| 71 | relh_grad_rolling_max_48h | 0.028 | 71 | relh_grad | 0.029 |
| 72 | relh_grad_rolling_max_24h | 0.027 | 72 | relh_rolling_mean_12h | 0.025 |
| 73 | sq_flag | 0.024 | 73 | relh_rolling_mean_6h | 0.019 |
| 74 | tmpf_lag_48h | 0.021 | 74 | relh | 0.016 |
| 75 | drct_u_lag_48h | 0.002 | 75 | drct_u_lag_48h | 0.000 |
| 76 | day_of_week_num | -0.062 | 76 | y | -0.086 |
| 77 | alti_rolling_mean_48h | -0.069 | 77 | mslp_rolling_mean_24h | -0.040 |
| 78 | alti | -0.070 | 78 | mslp_rolling_mean_48h | -0.039 |
| 79 | alti_rolling_mean_6h | -0.073 | 79 | mslp_rolling_mean_12h | -0.038 |
| 80 | mslp | -0.076 | 80 | IDW_alti | -0.038 |
| 81 | mslp_rolling_mean_48h | -0.077 | 81 | mslp_rolling_mean_6h | -0.036 |
| 82 | alti_rolling_mean_12h | -0.078 | 82 | mslp | -0.035 |
| 83 | IDW_alti | -0.079 | 83 | alti_rolling_mean_24h | -0.033 |
| 84 | mslp_rolling_mean_6h | -0.079 | 84 | alti_rolling_mean_48h | -0.031 |
| 85 | alti_rolling_mean_24h | -0.080 | 85 | alti_rolling_mean_12h | -0.031 |
| 86 | mslp_rolling_mean_12h | -0.084 | 86 | day_of_week_num | -0.030 |
| 87 | y | -0.085 | 87 | alti_rolling_mean_6h | -0.030 |
| 88 | mslp_rolling_mean_24h | -0.088 | 88 | alti | -0.029 |

# Appendix I. Modeling

**Table I.1** Selected features coefficients (Logistic Regression Stage-1) (features min-max scaled on train; coefficients are for the final L2 model)

| No. | Feature | Coefficient |
|---|---|---|
| 1 | IDW_dwpf_lag_12h | 0.096 |
| 2 | dwpf_lag_12h | 0.078 |
| 3 | tmpf_lag_6h | 0.057 |
| 4 | idw_tmpf_lag_6h | 0.038 |
| 5 | idw_drct_u_lag_12h | 0.033 |
| 6 | drct_u_lag_12h | 0.032 |
| 7 | rolling_max_12h | -0.007 |
| 8 | rolling_mean_12h | -0.005 |

**Table I.2. Hyperparameters & CV protocol (Logistic Regression Stage-1)**

- **Selection (L1)**: logistic (*penalty=L1*, *solver = 'saga'),* time-series CV (3 folds), select 8 predictors by non-zero coefficients.
- **Final fit (L2)**: logistic (*penalty* = L2, *solver* = *lbfgs*), grid over $C \in \{1e-3,\ 1e-2,\ 1e-1{,}1\}$; best C = 0.001.
- **Class weights:** {0: 1, 1: 5}(chosen on recall).
- **CV protocol:** time-series split (non-shuffled); train only in 2021.
- Threshold: 0.70 fixed on 2021 validation (PR-optimized).

**Table I.3**. Threshold sweep on the original (unsampled) distribution (Metrics on Train-2021 and Test-2022, predictions applied out-of-sample: Pass-through = share of county x hour forwarded to Stage-2).

| Split | Threshold | Recall | Precision | F1 | Pass-through (%) |
|---|---|---|---|---|---|
| Train-2021 | 0.65 | 0.985 | 0.069 | 0.130 | 95.58 |
| Test-2022 | 0.65 | 0.940 | 0.051 | 0.097 | 92.55 |
| Train-2021 | 0.70 | 0.873 | 0.088 | 0.160 | 66.73 |
| Test-2022 | 0.70 | 0.568 | 0.06 | 0.109 | 47.25 |

**Table I.4.** Confusion Matrix, Train (2021)

| | Actually Positive | Actually Negative |
|---|---|---|
| Predicted Positive | TP = 4,038 | FP = 41,845 |
| Predicted Negative | FN = 588 | TN = 22,284 |

**Table I.5.** Confusion Matrix, Test (2022)

| | Actually Positive | Actually Negative |
|---|---|---|
| Predicted Positive | TP = 1,790 | FP = 27,950 |
| Predicted Negative | FN = 1,362 | TN = 31,834 |

## **Methods Note I.6.** Stage-2 (Regression): LSTM hyperparameters and setup

**Inputs:** single-step tensors (*batch, 1, n_features*) without temporal windowing. After the Stage-1 filtering, the series becomes irregular and contains long gaps, complicating the formation of regular windows. The single-step input thus avoids padding and keeps all valid rows.
**Architecture**: LSTM (16 cells), Dense (1), with MSE loss function computed on *log_sum48;* optimizer =*Adam*; early stopping,
**Training**: *batch=32*, max=30 epochs, EarlyStopping (*patience = 2*)
**Environment (for reproducibility)**: Python 3.10; TensorFlow 2.18.0.
**Hyperparameter exploration:** hidden sizes {16, 32, 64, 128}; epoch counts {20, 50, 100}. Larger networks overfit on this small sample; LSTM-16 + Dense-1, trained for 30 epochs with early-stopping gave the most stable CV results. Light dropout (0.2/0.1) slightly improved mean MASE but worsened peak magnitude.
**Baseline (One-Step LSTM):** Same architecture/target/features/preprocessing as T-wo-Stage, but trained without Stage-1 filtering (full 2021). Windowed LSTMs (12/24/24 h) were also tried and failed to reproduce the magnitude of peaks.
**CV metrics (avg across folds, original scale).** Two-Stage LSTM : RMSE 7,997, MAE 1,360, $R^2$ 0.238. One-Step LSTM (avg across folds, original scale): RMSE 8,193, MAE 1,397, $R^2$ 0.216.

**Table I.7**.

| Fold | Train N | Val N | log-RMSE | log-MAE | $R^2$ log | log RMSE (Pruned) | log MAE (Pruned) | R2 log (Pruned) | Δ log RMSE |
|---|---|---|---|---|---|---|---|---|---|
| **1** | 3545 | 1493 | 4.023 | 3.306 | -0.739 | 4.512 | 3.862 | -1.187 | 0.489 |
| **2** | 10424 | 2231 | 2.232 | 1.723 | 0.458 | 2.247 | 1.726 | 0.45 | 0.014 |
| **3** | 16584 | 2480 | 1.743 | 1.264 | 0.389 | 1.738 | 1.278 | 0.392 | -0.005 |
| **4** | 24186 | 816 | 1.895 | 1.408 | 0.099 | 1.889 | 1.444 | 0.105 | -0.006 |
| **5** | 31576 | 2144 | 3.614 | 2.744 | -0.129 | 3.021 | 2.235 | 0.211 | -0.593 |

Forward-changing CV on train-2021, Stage-1 positives only.

**Table I.8**. 35-Feature Set (LSTM)

| No. | Feature (LSTM) |
|---|---|
| 1 | dwpf |
| 2 | tmpf |
| 3 | relh |
| 4 | alti |
| 5 | mslp |
| 6 | gust |
| 7 | p01i |
| 8 | sknt |
| 9 | drct_u |
| 10 | drct_v |
| 11 | relh_grad |
| 12 | ts_flag |
| 13 | hr_flag |
| 14 | dwpf_lag_6h |
| 15 | dwpf_lag_12h |
| 16 | dwpf_lag_24h |
| 17 | dwpf_lag_48h |
| 18 | gust_rolling_max_6h |
| 19 | idw_rolling_max_24h |
| 20 | idw_dwpf |
| 21 | idw_alti |
| 22 | Iidw_drct_u |
| 23 | idw_drct_v |
| 24 | idw_tmpf_lag_6h |
| 25 | idw_drct_v_lag_6h |
| 26 | idw_dwpf_lag_12h |
| 27 | idw_drct_u_lag_12h |
| 28 | Iidw_p01i_rolling_sum_24h |
| 29 | idw_gust_rolling_max_24h |
| 30 | idw_sknt_rolling_max_48h |
| 31 | idw_relh_rolling_mean_48h |
| 32 | day_of_week_num |
| 33 | population_density |
| 34 | county_encoded |
| 35 | y |

**Table I.9.** PCA Results with Possible Interpretation for the 35-Feature Set before Stage-1

| PC | Variance % | Key Features (> 0.2) | Possible Interpretation |
|---|---|---|---|
| PC 1 | **21%** | IDW_dwpf; dwpf; dwpf_lag_6h; IDW_dwpf_lag_12h; dwpf_lag_12h; IDW_drct_v_lag_6h; mslp; IDW_alti | Moisture gradient under falling pressure |
| PC 2 | **13%** | sknt; IDW_drct_u; alti; IDW_alti; drct_u; mslp; IDW_sknt_rolling_max_48h; IDW_gust_rolling_max_24h; gust | Background gusts and wind shear |
| PC 3 | **8%** | IDW_drct_v, drct_v, IDW_drct_v_lag_6h; dwpf_lag_24h; dwpf_lag_48h; IDW_drct_u_lag_12h; dwpf_lag_12h; IDW_sknt_rolling_max_48h | Indicate coming of moist southern air masses over 1-2 days |
| PC 4 | **6%** | tmpf; relh; sknt; alti; IDW_alti; gust; mslp; drct_u | Current conditions potentially indicating mesoscale instability peak |

**Table I.10.** PCA Results with Possible Interpretation for the 35-Feature Set after Stage-1

| PC | Variance % | Key Features (> 0.2) | Possible Interpretation |
|---|---|---|---|
| PC 1 | **19%** | IDW_drct_v_lag_6h; IDW_drct_v; IDW_dwpf; dwpf; drct_v; IDW_gust_rolling_max_24h; dwpf_lag_6h; sknt; IDW_dwpf_lag_12h; IDW_sknt_rolling_max_48h; mslp; IDW_alti | Low-level moisture advection with gusty winds indicating southern air masses. |
| PC 2 | **12%** | IDW_alti, alti; mslp; dwpf; IDW_dwpf; dwpf_lag_6h; IDW_dwpf_lag_12h; dwpf_lag_12h; sknt; IDW_drct_u | Represents the background pressure field and the slowly varying air masses |
| PC 3 | **9%** | dwpf_lag_48h; relh; dwpf_lag_24h; IDW_drct_v; drct_u; IDW_drct_u; drct_v; alti; mslp; IDW_drct_u_lag_12h | Lagged moisture and directional shear. Might indicate accumulation of low-level moisture (might indirectly indicate soil and vegetation saturation) |
| PC 4 | **7%** | tmpf; relh; IDW_sknt_rolling_max_48h; IDW_p0li_rolling_sum_24h; IDW_gust_rolling_max_24h; IDW_relh_rolling_mean_48h | Combines temperature-humidity contrast. High PC 4 denote hours when the air is both warm and moist, with strong gusts and precipitation indicators. |

**Table I.11**. Key metrics of the LSTM-Two-Stage (original scale)

| Split | MSE | RMSE | MAE | $R^2$ |
|---|---|---|---|---|
| Train | 16,890,813 | 4,110 | 372 | -0.05 |
| Test | 21,239,688 | 4,608 | 606 | 0.13 |

**Table I.12.** Key metrics of the LSTM-Two-Stage (log-scale)

| Split | MSE | RMSE | MAE | $R^2$ |
|---|---|---|---|---|
| Train | 3.57 | 1.89 | 1.43 | 0.59 |
| Test | 7.29 | 2.70 | 2.03 | -0.27 |

**Table I.13**. Key metrics of One-Step LSTM (original scale)

| Split | MSE | RMSE | MAE | $R^2$ |
|---|---|---|---|---|
| Train | 87,742 | 9,367 | 1,089 | 0.03 |
| Test | 16,891 | 4,110 | 372 | 0.11 |

**Table I.14**. Key metrics of One-Step LSTM (log-scale)

| Sample | MSE | RMSE | MAE | $R^2$ |
|---|---|---|---|---|
| Train (log-scale) | 3.44 | 1.86 | 1.40 | 0.56 |
| Test (log) | 5.40 | 2.32 | 1.73 | -0.05 |

**Table I.15.** cMASE on peak-hour (thr=50,000), All-hours. Lower is better.
Δ% = 100*(1 - Two-Stage/Baseline)

| Window (±h) | Baseline cMASE (median [2.5; 97.5]) | Two-Stage cMASE (median [2.5; 97.5]) | Δ% (Two-Stage vs. Baseline) |
|---|---|---|---|
| 6 | 4.310 [2.665; 11.315] | 3.767 [2.023; 11.166] | 12.6 |
| 12 | 3.385 [1.915; 9.078] | 3.118 [1.816; 8.790] | 7.9 |
| 24 | 2.394 [1.388; 6.196] | 2.432 [1.536; 6.291] | -1.6 |
| 36 | 2.071 [1.181; 4.851] | 2.180 [1.326; 4.877] | -5.3 |
| 48 | 1.867 [1.139; 4.035] | 2.001 [1.177; 4.095] | -7.2 |

**Table I.16.** cMASE on peak-hour (thr=50,000), Available-hours (coverage = 65.7%). Lower is better. Δ% = 100*(1 - Two-Stage/Baseline)

| Window (±h) | Baseline cMASE (median [2.5; 97.5]) | Two-Stage cMASE (median [2.5; 97.5]) | Δ% (Two-Stage vs. Baseline) |
|---|---|---|---|
| 6 | 3.194 [2.282; 8.079] | 2.902 [1.360; 8.062] | 9.2 |
| 12 | 2.557 [1.528; 6.211] | 2.324 [1.152; 6.273] | 9.1 |
| 24 | 1.791 [1.046; 4.301] | 1.821 [1.052; 4.381] | -1.7 |
| 36 | 1.569 [0.968; 3.381] | 1.680 [0.917; 3.464] | -7.1 |
| 48 | 1.408 [0.851; 2.838] | 1.545 [0.813; 2.965] | -9.8 |

**Table I.17.** Event detection (All-hours). One-to-one matching within ±w hours. Values are medians with 95% percentile intervals 2.5; 9.5.

| Window (±h) | Recall Baseline (%) | Recall Two-Stage (%) | F1 Baseline (%) | F1 Two-Stage (%) |
|---|---|---|---|---|
| 6 | 0.00 [0.00; 0.00] | 0.00 [0.00; 33.33] | 0.00 [0.00; 0.00] | 25.00 14.29; 40.00] |
| 12 | 0.00 [0.00; 0.00] | 0.00 [0.00; 50.00] | 22.50 [20.12; 24.88] | 25.00 [15.38; 66.67] |
| 24 | 0.00 [0.00; 25.00] | 0.00 [0.00; 50.00] | 25.00 [18.18; 50.00] | 25.00 [15.38; 66.67] |
| 36 | 0.00 [0.00; 33.33] | 0.00 [0.00; 62.50] | 8.57 [20.00; 50.00] | 25.00 [15.38; 66.67] |
| 48 | 25 [0.0; 100.0] | 33.3 [0.0; 100.0] | 40.00 [22.22; 80.00] | 40.00 [18.18; 82.00] |

**Table I.18.** Event detection (Available-hours). Same as I.17; scope restricted to timestamps where Stage-2 predictions exist.

| Window (±h) | Recall Baseline (%) | Recall Two-Stage (%) | F1 Baseline (%) | F1 Two-Stage (%) |
|---|---|---|---|---|
| 6 | 0.00 [0.00; 0.00] | 0.00 [0.00; 33.33] | 0.00 [0.00; 0.00] | 25.00 [17.05; 62.50] |
| 12 | 0.00 [0.00; 0.00] | 0.00 [0.00; 74.38] | 47.62 [29.52; 65.71] | 28.57 [16.67; 66.67] |
| 24 | 0.00 [0.00; 48.75] | 0.00 [0.00; 100.00] | 33.33 [20.00; 66.67] | 28.57 [16.67; 66.67] |
| 36 | 0.00 [0.00; 50.00] | 0.00 [0.00; 100.00] | 33.33 [19.45; 66.67] | 28.57 [16.67; 66.67] |
| 48 | 33.3 [0.0; 100.0] | 50.0 [0.0; 100.0] | 50.0 [25.7; 85.7] | 50.0 [20.0; 88.9] |

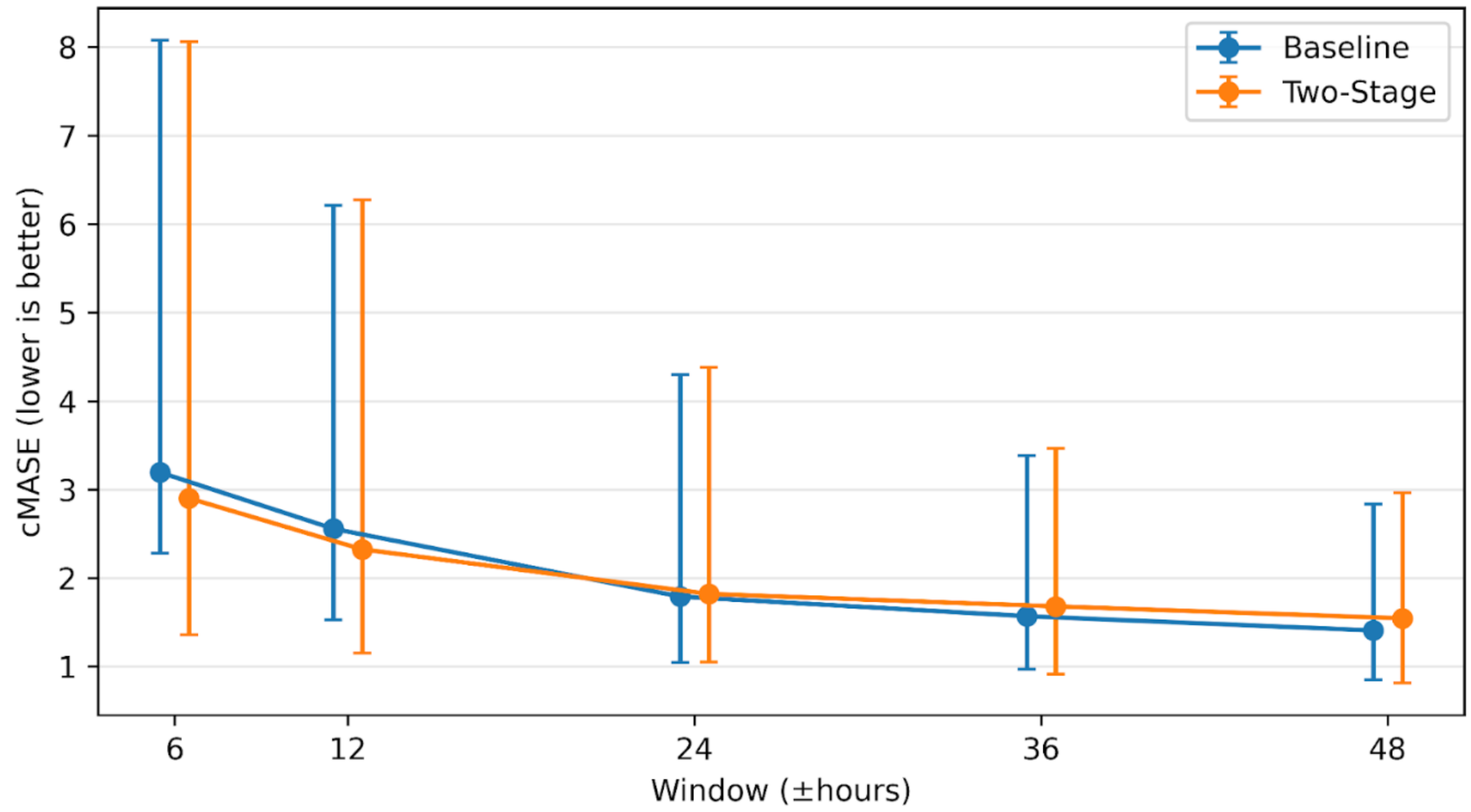

**Fig. I.1.** Peak-conditional cMASE on Available Hours (Test 2022). Points bootstrap medians (moving-block, 168 h, B=500). 95% percentile intervals [2.5, 97.5]. Lower is better.

## Appendix J. SHAP

**Table J.1.** SHAP for Two-Stage LSTM Model based on 3000

| Feature | Mean Positive SHAP | Mean Negative SHAP | Mean Absolute SHAP | Share % |
|---|---|---|---|---|
| IDW_dwpf_lag_12h | 2.257 | 0.000 | 2.255 | 36.15% |
| IDW_tmpf_lag_6h | 0.246 | -0.917 | 0.803 | 12.86% |
| dwpf_lag_12h | 0.649 | -0.118 | 0.529 | 8.47% |
| population_density | 1.374 | -0.486 | 0.515 | 8.26% |
| y | 0.516 | -0.622 | 0.341 | 5.46% |
| IDW_drct_v | 0.136 | -0.552 | 0.212 | 3.40% |
| IDW_sknt_rolling_max_48h | 0.522 | -0.338 | 0.198 | 3.18% |
| alti | 0.396 | -0.248 | 0.157 | 2.52% |
| day_of_week_num | 0.455 | -0.426 | 0.134 | 2.14% |
| IDW_alti | 0.230 | -0.380 | 0.106 | 1.70% |
| mslp | 0.275 | -0.354 | 0.088 | 1.41% |
| IDW_dwpf | 0.176 | -0.322 | 0.084 | 1.34% |
| relh | 0.260 | -0.259 | 0.066 | 1.05% |
| IDW_drct_v_lag_6h | 0.196 | -0.290 | 0.056 | 0.90% |
| IDW_relh_rolling_mean_48h | 0.266 | -0.269 | 0.056 | 0.89% |
| IDW_drct_u_lag_12h | 0.246 | -0.250 | 0.051 | 0.81% |
| IDW_gust_rolling_max_24h | 0.292 | -0.211 | 0.049 | 0.79% |
| gust_rolling_max_6h | 0.235 | -0.262 | 0.044 | 0.71% |
| sknt | 0.179 | -0.280 | 0.043 | 0.69% |
| dwpf_lag_48h | 0.213 | -0.281 | 0.039 | 0.62% |
| IDW_drct_u | 0.227 | -0.220 | 0.039 | 0.62% |
| tmpf | 0.236 | -0.186 | 0.039 | 0.62% |
| IDW_p0li_rolling_sum_24h | 0.242 | -0.233 | 0.034 | 0.55% |
| drct_v | 0.218 | -0.176 | 0.032 | 0.51% |
| gust | 0.214 | -0.203 | 0.031 | 0.50% |
| dwpf_lag_6h | 0.219 | -0.215 | 0.030 | 0.48% |
| dwpf | 0.183 | -0.227 | 0.029 | 0.47% |
| relh_grad | 0.195 | -0.219 | 0.028 | 0.46% |
| county_encoded | 0.195 | -0.198 | 0.027 | 0.43% |
| drct_u | 0.198 | -0.206 | 0.026 | 0.41% |
| dwpf_lag_24h | 0.183 | -0.224 | 0.026 | 0.41% |
| p0li | 0.191 | -0.208 | 0.026 | 0.41% |
| hr_flag | 0.210 | -0.187 | 0.025 | 0.39% |
| ts_flag | 0.176 | -0.210 | 0.024 | 0.38% |

**Table J.2.** SHAP for One-Step LSTM Model based on 3000 Sample

| Feature | Mean Positive SHAP | Mean Negative SHAP | Mean Absolute SHAP | Share % |
|---|---|---|---|---|
| population_density | 1.312 | -0.359 | 0.544 | 28.13% |
| y | 0.509 | -0.339 | 0.371 | 19.22% |
| IDW_sknt_rolling_max_48h | 0.439 | -0.226 | 0.237 | 12.26% |
| day_of_week_num | 0.169 | -0.271 | 0.138 | 7.15% |
| gust_rolling_max_6h | 0.198 | -0.151 | 0.136 | 7.02% |
| relh | 0.146 | -0.105 | 0.064 | 3.31% |
| IDW_drct_u | 0.126 | -0.166 | 0.050 | 2.56% |
| gust | 0.170 | -0.049 | 0.042 | 2.17% |
| county_encoded | 0.121 | -0.091 | 0.042 | 2.17% |
| alti | 0.126 | -0.097 | 0.041 | 2.14% |
| IDW_drct_v_lag_6h | 0.103 | -0.088 | 0.030 | 1.53% |
| IDW_relh_rolling_mean_48h | 0.085 | -0.102 | 0.026 | 1.34% |
| sknt | 0.079 | -0.080 | 0.019 | 1.00% |
| dwpf_lag_48h | 0.088 | -0.078 | 0.019 | 0.98% |
| IDW_dwpf | 0.092 | -0.060 | 0.016 | 0.83% |
| drct_u | 0.082 | -0.074 | 0.015 | 0.79% |
| IDW_alti | 0.107 | -0.044 | 0.014 | 0.71% |
| IDW_p0li_rolling_sum_24h | 0.101 | -0.071 | 0.013 | 0.69% |
| mslp | 0.095 | -0.049 | 0.013 | 0.69% |
| IDW_gust_rolling_max_24h | 0.090 | -0.047 | 0.013 | 0.69% |
| relh_grad | 0.109 | -0.049 | 0.012 | 0.64% |
| IDW_drct_v | 0.078 | -0.054 | 0.011 | 0.57% |
| dwpf_lag_24h | 0.065 | -0.050 | 0.007 | 0.36% |
| dwpf_lag_6h | 0.065 | -0.045 | 0.007 | 0.36% |
| hr_flag | 0.100 | -0.037 | 0.007 | 0.36% |
| dwpf | 0.059 | -0.039 | 0.006 | 0.33% |
| drct_v | 0.060 | -0.041 | 0.006 | 0.32% |
| ts_flag | 0.049 | -0.061 | 0.005 | 0.27% |
| tmpf | 0.054 | -0.042 | 0.005 | 0.25% |
| p0li | 0.065 | -0.036 | 0.005 | 0.24% |
| IDW_tmpf_lag_6h | 0.055 | -0.037 | 0.005 | 0.24% |
| dwpf_lag_12h | 0.051 | -0.037 | 0.004 | 0.22% |
| IDW_drct_u_lag_12h | 0.053 | -0.033 | 0.004 | 0.22% |
| IDW_dwpf_lag_12h | 0.048 | -0.036 | 0.004 | 0.22% |